\documentclass[conference]{IEEEtran}
\IEEEoverridecommandlockouts
\usepackage{cite}
\usepackage{amsmath,amssymb,amsfonts}
\usepackage{graphicx}
\usepackage{textcomp}
\usepackage{xcolor}
\def\BibTeX{{\rm B\kern-.05em{\sc i\kern-.025em b}\kern-.08em
    T\kern-.1667em\lower.7ex\hbox{E}\kern-.125emX}}
\usepackage{booktabs}
\usepackage{algorithm}
\usepackage{algpseudocode}
\usepackage{comment}
\usepackage{stfloats}
\usepackage{caption}
\usepackage{subcaption}
\newcommand{\ra}[1]{\renewcommand{\arraystretch}{#1}}
\usepackage[breaklinks,colorlinks]{hyperref}

\begin{document}

\title{A Skeleton-based Approach For Rock Crack Detection Towards A Climbing Robot Application}

\author{\IEEEauthorblockN{Josselin Somerville Roberts}
\IEEEauthorblockA{\textit{Department of}\\
\textit{Computer Science} \\
\textit{Stanford University}\\
josselin@stanford.edu}
\and
\IEEEauthorblockN{Paul-Emile Giacomelli}
\IEEEauthorblockA{\textit{Department of}\\
\textit{Aeronautics and Astronautics} \\
\textit{Stanford University}\\
pegiaco@stanford.edu}
\and
\IEEEauthorblockN{Yoni Gozlan}
\IEEEauthorblockA{\textit{Institute for Computational}\\
\textit{and Mathematical Engineering}\\
\textit{Stanford University}\\
yonigoz@stanford.edu}
\and
\IEEEauthorblockN{Julia Di}
\IEEEauthorblockA{\textit{Department of}\\
\textit{Mechanical Engineering} \\
\textit{Stanford University}\\
juliadi@stanford.edu}
}

\maketitle

\begin{abstract}
Conventional wheeled robots are unable to traverse scientifically interesting, but dangerous, cave environments. Multi-limbed climbing robot designs, such as ReachBot, are able to grasp irregular surface features and execute climbing motions to overcome obstacles, given suitable grasp locations. To support grasp site identification, we present a method for detecting rock cracks and edges, the SKeleton Intersection Loss (SKIL). SKIL is a loss designed for thin object segmentation that leverages the skeleton of the label. A dataset of rock face images was collected, manually annotated, and augmented with generated data. A new group of metrics, LineAcc, has been proposed for thin object segmentation such that the impact of the object width on the score is minimized. In addition, the metric is less sensitive to translation which can often lead to a score of zero when computing classical metrics such as Dice on thin objects. Our fine-tuned models outperform previous methods on similar thin object segmentation tasks such as blood vessel segmentation and show promise for integration onto a robotic system.
\end{abstract}



\section{Introduction}

\begin{figure*}[b]
  \centering
   \includegraphics[width=1\linewidth]{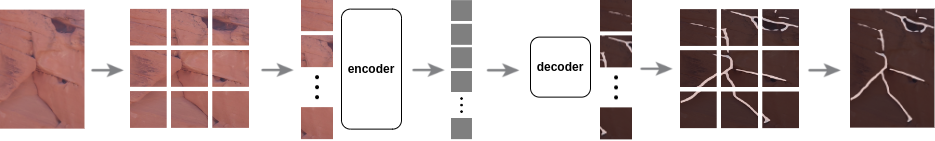}
   \caption{Workflow of the ViT-B based model discussed in this work (Section \ref{sec:model}) with crop size of $512$ and patch size of $16$.}
   \label{fig:reachbot-worklow}
\end{figure*}

In exploration tasks, robots offer versatility and robustness for navigating novel environments. However, challenging terrain, like caverns and steep slopes which are of high astrobiological and geological interest, have often hampered traditional approaches such as wheeled rovers. Aerial vehicles meanwhile may be suited for rough terrain exploration, but have limited flight time and ability to traverse caves. To address this mobility gap, ReachBot is a novel robot concept that uses microspine grippers at the end of extendable booms to achieve a large workspace and wrench capability. This results in higher navigation capability across the rocky scrambles that one may find in lava tubes and caves on the Moon and Mars \cite{SchneiderBylardEtAl2022, NewdickChenEtAl2023, NewdickOngoleEtAl2023}. 

Of crucial importance is a perception system that identifies a good grasp site on a realistic rock surface. While one may identify grasp sites through basic primitives, such as trial and error, we note that in climbing robot applications, misplaced holds may lead to mission failure and robot damage. 

Furthermore, ReachBot's large workspace encourages a grasp site identification method capable from a distance. As an example, we consider cracks and edges as promising rock features for microspine grippers, because microspines may engage well with the asperities of a crack or edge leading to a strong grasp \cite{asbeck2006scaling}. Yet identifying rock cracks from a distance can prove complex because cracks are typically long and thin features that are susceptible to lighting conditions and the angle of the camera with respect to the rock wall.

\textit{Statement of Contributions:} This work focuses on improving rock crack detection towards a climbing robot application, with the following contributions:
\begin{itemize}
    \item We present a new segmentation loss, SKIL (SKeleton Intersection Loss), to identify thin features such as cracks on rock walls, blood vessels, or other thin objects.
    \item We propose novel metrics for evaluating the detection results of fine and longitudinal objects, LineAcc. These metrics are based on the position (LineAcc-pos) and sizes (LineAcc-width/length) of the cracks, and are more robust to ground truth labeling errors than classical metrics.
    \item We manually collected and generated a new dataset of rock wall images, which is open-sourced to support the development of other rock climbing-related applications. All code, commands, and datasets are included in this Github \href{https://github.com/JosselinSomervilleRoberts/ReachBot-Predictor/tree/main}{repo} as well as instructions to recreate results.
\end{itemize}

To assess the added value of SKIL and our new metric, we evaluated and compared performance on blood vessel segmentation benchmark datasets, an existing and related task in the medical imaging community.

\textit{Paper Organization:} The rest of the paper is organized as follows.  Section \ref{sec:related_works} provides an overview of related works for the task. Section \ref{sec:mechanisms} discusses the robot design and important considerations for crack detection. Section \ref{sec:proposed approach} explains
the proposed SKIL loss and new metrics. Section \ref{sec:results} describes the experiments and results. Finally, Section \ref{sec:conclusion} summarizes our contributions and explores future work.


\vspace{0.4em}
\section{Related Works}
\vspace{-0.4em}
\label{sec:related_works}

\subsection{Existing Datasets}
We examined the possibility of leveraging existing datasets for this work. Rock segmentation for indoor climbing or climbing robots generally feature artificially colored rocks with indoor lighting, which are not representative of the outdoor volcanic rock walls that ReachBot will encounter on a potential mission to a Martian lava tube \cite{indoor_climbing, climb-o-vision, pandurevic2022analysis}. Rock classification datasets are collected outdoors and can be specified by sediment type (including volcanic), but rock type alone does not necessarily correspond to whether a rock is graspable, and many datasets are only of standalone rocks and not natural rock faces as one would encounter in a cave or lava tube \cite{rock_classification_1, rock_classification_2, ran2019rock, lacassie2006visualization}. Datasets on infrastructural cracks are also potential candidates for crack detection, but most man-made structures like concrete walls and bridges are macroscopically smooth with very fine cracks, and do not represent realistic surfaces for this robot mission \cite{ali2022structural, pavement_concrete_cracks_review}.

\subsection{Existing Methods}
Because ReachBot does not have an anthropomorphic gripper, existing grasp site identification approaches developed for human climbing do not fit the kinematics of the gripper \cite{pandurevic2022analysis}. Alternatively, there are models for the microspine grippers on a variety of rock-climbing robots \cite{parness2017lemur, asbeck2006scaling, ruotolo2019load}, and past analyses have developed a stochastic wrench limit surface, which depends on both the gripper design and surface topography \cite{wang2019spinyhand, jiang2018stochastic,  iacoponi2020simulation}. While these model-based approaches work for analyzing microspine grasps, they often require detailed ground truth topographical knowledge which limits them from deployment on a real robot. 

Therefore, we are interested in learning to detect rock \textit{features} for grasping, such as cracks and edges. As a first step, crack detection methods exist in other related applications such as infrastructural monitoring, but these tend to be extremely fine cracks \cite{ZHOU2023104678, MINHDANG2022129438}. A similar task can be found in the medical imaging space with blood vessel segmentation \cite{SAMUEL2021105769, Kar2023}. Several losses have been proposed for blood vessels segmentation task (e.g. imbalanced\cite{loss_patch_based}, connected\cite{cl-dice} and thin\cite{loss_thin}) and some state-of-the-art methods for vessel segmentation, such as U-Net \cite{unet, unet++, unet3+}, appear promising for rock crack detection. We further discuss in Section \ref{sec:results} their performance for this use case, and investigate a modified skeleton-based approach for detecting crack and edge features for a robotics application, and benchmark performance with related blood vessel datasets.

\subsection{Existing Metrics}

For image segmentation, IoU and Dice coefficient are commonly used metrics \cite{hesamian2019deep}. These metrics measure the pixel-wise agreement between prediction and ground truth, which penalizes any small misalignments between the predicted crack pixels and the true crack (which may often occur with manual annotation). Further, adjustments to the Dice score to make the prediction less sensitive to small position misalignments, such as a diffused prediction, would still not help us understand model performance on graspable cracks.

Many alternative metrics have been proposed for image segmentation \cite{metric-review}. In the context of crack detection, spatial distance-based metrics may be relevant such as the Hausdorff distance \cite{hausdorff}. However, they are sensitive to the thickness of the thin objects, which is uncertain in our case. One proposed metric is to only compute the Dice score around thin zones of the ground truth \cite{iou-thin}. This metric does not reflect performance on larger zones of ground truth and does not mitigate the translation sensitivity.


\vspace{0.3em}
\section{Gripper Discussion}
\vspace{-0.3em}
\label{sec:mechanisms}

\begin{figure}[h]
  \centering
   \includegraphics[width=1\linewidth]{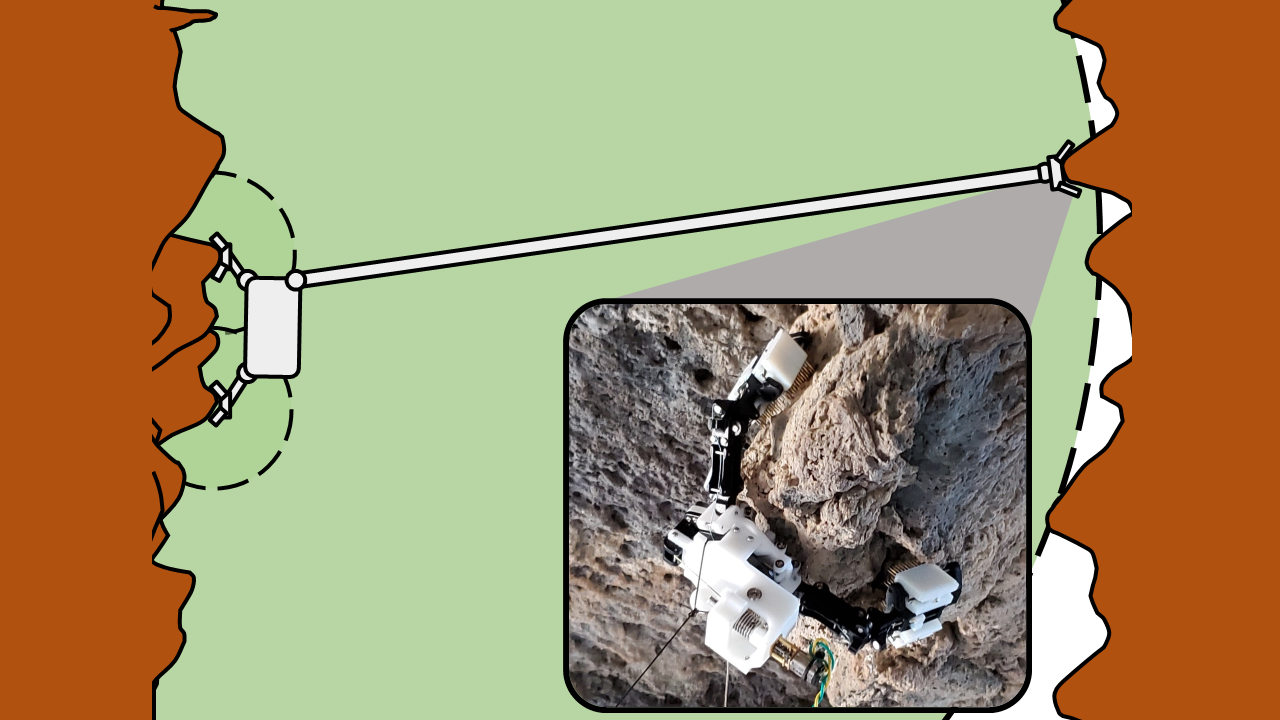}
   \caption{The ReachBot concept uses microspine grippers at the end of extendable booms to reach a large workspace inside a Martian lava tube. Inset is a microspine gripper prototype grasping real lava rock in a Mojave Lava Tube.}
   \label{fig:reachbot-diagram}
\end{figure}

The ReachBot design concept allows for navigation through cave-like environments with the use of microspine grippers on the end of extendable booms, as shown in Fig. \ref{fig:reachbot-diagram}. 

\setlength{\floatsep}{1pt}            
\setlength{\abovedisplayskip}{5pt}
\setlength{\belowdisplayskip}{5pt}
\setlength{\abovedisplayshortskip}{2pt}
\setlength{\belowdisplayshortskip}{2pt}

ReachBot's microspine gripper, inset in Fig. \ref{fig:reachbot-diagram}, grasps onto rocks successfully when a sufficient number of microspines are engaged on the surface asperities (such as those around the rock cracks). Because microspine grasping is inherently stochastic, there is no formal guarantee for grasp success. However, several parameters influence the envelope of graspable features, including the distribution of millimeter-scale asperities of the surface, the geometry of the surface based on the gripper design, and the sharpness of the microspines (which dull after too much continuous use) \cite{chen2023reachbot}. 

Because obtaining micrometer-level surface characterization of the asperities is not practical on the robot, we instead investigate macroscale rock \textit{features} for grasping. Akin to how a trained rock climber may look for certain rock shapes to fit their hand, previous work has already investigated rock geometry analysis for a ReachBot-specific gripper \cite{chen2023reachbot}. Classical shape-fitting techniques were used, which work best at identifying symmetrically convex shapes but miss potential irregular grasp sites such as cracks, ledges, and edges. It is important to identify as many grasp sites as possible because ReachBot's unique design enables navigation even when grasp sites are sparse, but only if the robot knows they exist. 

\vspace{0.8em}
\section{Proposed Approach}
\vspace{-0.3em}
\label{sec:proposed approach}

\subsection{Model Architecture}
\label{sec:model}
All the experiments were run with a ViT-B\cite{ViT_original} backbone with a crop size of $512 \times 512$ and a patch size of $16 \times 16$ using the UPerNet\cite{upernet} method. This model was chosen qualitatively as it offered good performances after a few hundred epochs. Other models considered included SAM\cite{SAM}, ResNet\cite{Resnet}, and DeiT\cite{deit}, but were deemed too complex for a resource-constrained hardware context. As the goal is to compare losses and eventually run on a field robot in a cave, we wanted to keep the model as simple as possible.

\subsection{LineAcc Segmentation Metrics}
\label{sec:metric}
We propose several metrics, grouped under the name of \textbf{LineAcc}, to measure different aspects of the segmentation that classic metrics did not capture as discussed earlier. 

\subsubsection{Center line position}
\label{sec:metric-center}
The first metric is \textbf{the center line position}, which is referred to as $\text{LineAcc}_{\text{pos}}$. The goal is to measure if the cracks predicted are well positioned regardless of the width or length of the crack.

Given a prediction $\mathcal{P}$ and label $\mathcal{L}$, we introduce their skeletons $\mathcal{S_P}$ and $\mathcal{S_L}$. The metric $\text{LineAcc}_{\text{pos}}$ is defined as:
\begin{equation}
\begin{split}
  \text{LineAcc}_{\text{pos}}(\mathcal{P},\mathcal{L}) &= \frac{\sum \mathcal{S_P} \times e^{- d(\mathcal{S_L})^2 / 2\sigma^2}}{\sum \mathcal{S_P}} \\
  &\times \frac{\sum \mathcal{S_L} \times e^{- d(\mathcal{S_P})^2 / 2\sigma^2}}{\sum \mathcal{S_L}}
  \label{eq:LineAccPos}
\end{split}
\end{equation}

With $d$ a distance function such that for $I \in \mathbb{R}^{H \times W}$, \begin{equation}
    d(I)_{x,y} = \min_{u,v, I(u,v) = 1} |x-u| + |y-v|
    \label{eq:d}
\end{equation}
This metric returns a score illustrating how close the two skeletons are (with a Gaussian decrease). 
\\

\subsubsection{Width and length ratios}
\label{sec:metric-width-and-length}
We further introduce two ratio-based metrics: the \textbf{width} and \textbf{length} ratios. These two metrics illustrate how close the width and length of a predicted crack is to the ground truth, independently of the position of the crack.
\begin{equation}
    \text{LineAcc}_{\text{length}}(\mathcal{P},\mathcal{L}) = \exp \left( - \left| \frac{\sum \mathcal{S_P} + \epsilon}{\sum \mathcal{S_L} + \epsilon} - 1\right| \right)
\end{equation}
\begin{equation}
    \text{LineAcc}_{\text{width}}(\mathcal{P},\mathcal{L}) = \exp \left( - \left| \frac{\left(\sum \mathcal{L}\right) \left(\sum \mathcal{S_P}\right) + \epsilon}{\left(\sum \mathcal{P}\right) \left(\sum \mathcal{S_L}\right) + \epsilon} - 1\right| \right)
\end{equation}
These two metrics are also not correlated as the length ratio $\text{LineAcc}_{\text{length}}$ relies on the skeletons (independent to width) and the width ratio $\text{LineAcc}_{\text{width}}$ uses the inverse of the length ratio to balance its effect. A constant $\epsilon = 10^{-3}$ is added to avoid dividing by zero.

\subsection{SKIL: a new loss}
\label{sec:loss}
We investigated new losses that optimize for the new metrics. Although CL-Dice \cite{cl-dice} loss appeared promising (computes the Dice score between the skeleton of the prediction and the ground truth label), the skeleton intersection is dependent on the width of the labels and the Dice score does not optimize the center line position well. As we plan to use a combination of Dice and a new loss, this was not desirable.

\subsubsection{SKIL-Dice, a loss adapted to the new metrics}
\label{sec:loss-dice}
The core idea of SKIL-Dice is a skeleton-based approach that abstracts the width of the crack. Another improvement that was necessary was to integrate the distance from the two center lines. As an example, with CL-Dice, if the ground truth is a thin vertical rectangle of width $w$, then whether the prediction is shifted by a distance $D$ of 1 pixel or $w/2 -1$, the score would be the same. However, the $\text{LineAcc}_{\text{pos}}$ score would be proportional to $e^{-D^2/2\sigma^2}$. 

In order to solve this issue, we propose to first compute the soft skeleton (first introduced \cite{cl-dice}), then apply a smooth diffusion and finally compute the Dice score between the two diffused skeletons. 

Given a binary image $I$, we compute with our smooth diffusion the diffused image $\text{Dif}(I)$ such that the pixels in $I$ equal to 1 are still equal to 1 in $\text{Dif}(I)$, and the pixels equal to 0 in $I$ are now equal in $\text{Dif}(I)$ to a value that depends on the distance to the closest pixel equal to 1 in $I$. One could think of using the distance function $d$ defined in Eq. \ref{eq:d}, but this is not differentiable. We propose the following algorithm:

\begin{algorithm}
  \caption{Smooth binary mask diffusion}\label{alg:diffusion}
  \begin{algorithmic}[h]
    \Procedure{smooth\_dilate}{$I, s_{\text{border}}, n_{\text{iter}}^{\text{max}}, f$}
      \State $s_{\text{dilate}} \gets \max\left(1, s_{\text{border}} // n_{\text{iter}}^{\text{max}}\right)$
      \State $I_{\text{enlarged}} \gets \text{copy}(I)$
      \For{$d \in \{1, ..., \min\left(s_{\text{border}}, s_{\text{dilate}}\right)\}$}
        \State $I_{\text{enlarged}}^{\text{new}} \gets \textbf{soft\_dilate}(I_{\text{enlarged}},  s_{\text{dilate}})$
        \State $I_{\text{enlarged}} \gets f * I_{\text{enlarged}} + (1-f) * I_{\text{enlarged}}^{\text{new}}$
      \EndFor
      \State \textbf{return} $I_{\text{enlarged}}$
    \EndProcedure
  \end{algorithmic}
\end{algorithm}

where $\textbf{soft\_dilate}(I, d)$ is a maxpool with a kernel of size $(1 + 2d) \times (1 + 2d)$ with the corners removed.

This introduces 3 parameters of the loss: $s_{\text{border}}$ the size of the diffused skeleton, $n_{\text{iter}}^{\text{max}}$ the maximum number of iterations for the diffusion (more iterations take more memory), and $f$ the factor that controls the decrease. Finally, SKIL-Dice is defined as (for the sake of simplicity parameters are omitted): 
\begin{equation}
    \text{SKIL}^{\text{Dice}}(P,L) = \text{Dice}(\text{Dif}(S_{\Tilde{P}}), \text{Dif}(S_L))
\end{equation}
with $\Tilde{P} = T(P,s)$ the prediction threshold. We use a smooth threshold $T(I,s)$ with $s$ the sharpness:
\begin{equation}
    T(P,s) = \text{sigmoid}\left(s(P-0.5)\right)
\end{equation}

\subsubsection{SKIL-Product}
\label{sec:loss-variant}
We also propose another variant, SKIL-Product. This version is as similar as possible to the CL-Dice loss. The predictions and labels are skeletonized, then diffused and finally the product of the predicted skeleton and the diffused skeleton of the label is computed. This is done instead of using the Dice score in order to have a normalization factor of 1.
\begin{equation}
    g(A,B) = \frac{\sum \left( S_A*\text{Dif}(S_B)\right) + \epsilon}{\sum S_A + \epsilon}
\end{equation}
\begin{equation}
    \text{SKIL}^{\text{Prod}}(P,L) = 1 - \sqrt{g(\Tilde{P},L) * g(L,\Tilde{P})}
\end{equation}


\vspace{0.7em}
\section{Results}
\vspace{-0.7em}
\label{sec:results}
We discuss the following questions: 
\begin{itemize}
    \item Are our new metrics visually compelling? For example, given two predictions of identical Dice scores, does LineAcc capture if one is visually better than the other?
    \item Does SKIL improve performance and if yes which variant is better: SKIL-Dice or SKIL-Product?
    \item Is SKIL useful in similar segmentation tasks?
    \item Does SKIL help on poorly annotated images?
\end{itemize}

\subsection{Dataset collection}
\label{sec:result-dataset}

The \textbf{Cracks Reachbot} dataset has been open-sourced and is available \href{https://drive.google.com/drive/u/1/folders/1-3A6pQJ-ASxK9UKKm5T2XSwLYLYB8qZV}{here}.

\subsubsection{Real data}
\label{sec:result-dataset-real}

\begin{figure}[h]
  \captionsetup[subfigure]{justification=centering, labelformat=empty} 
  \centering
  
  \begin{subfigure}{0.245\linewidth}
    \includegraphics[width=\linewidth]{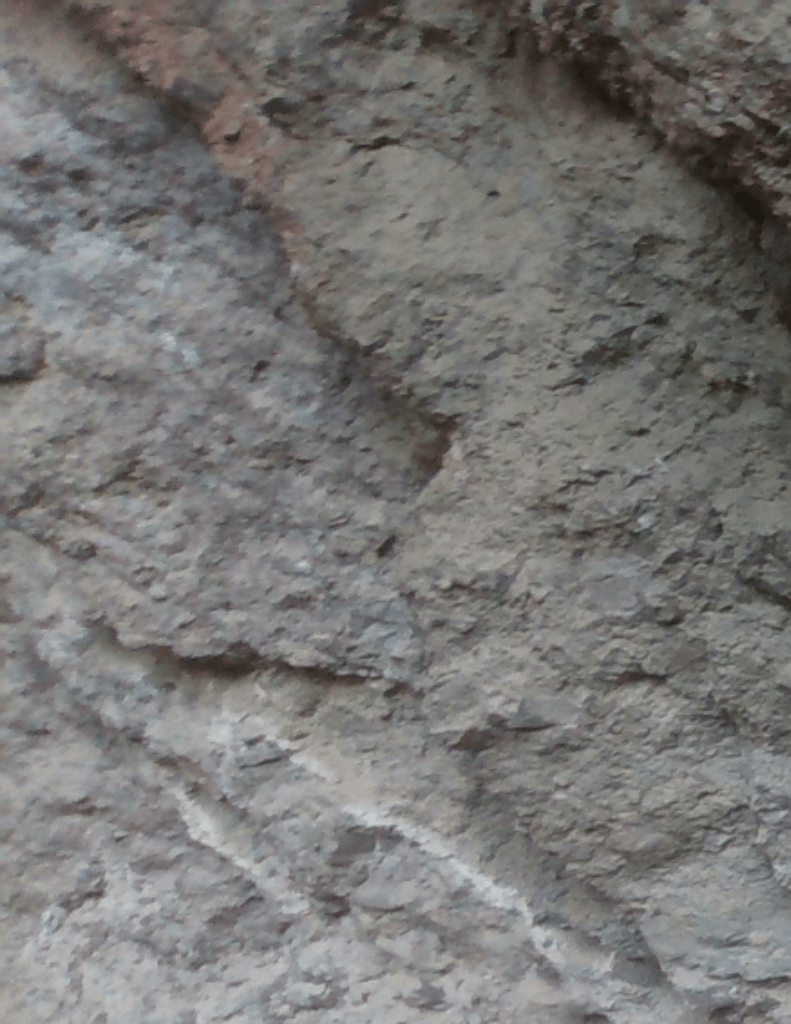}
  \end{subfigure}
  \begin{subfigure}{0.426\linewidth}
    \includegraphics[width=\linewidth]{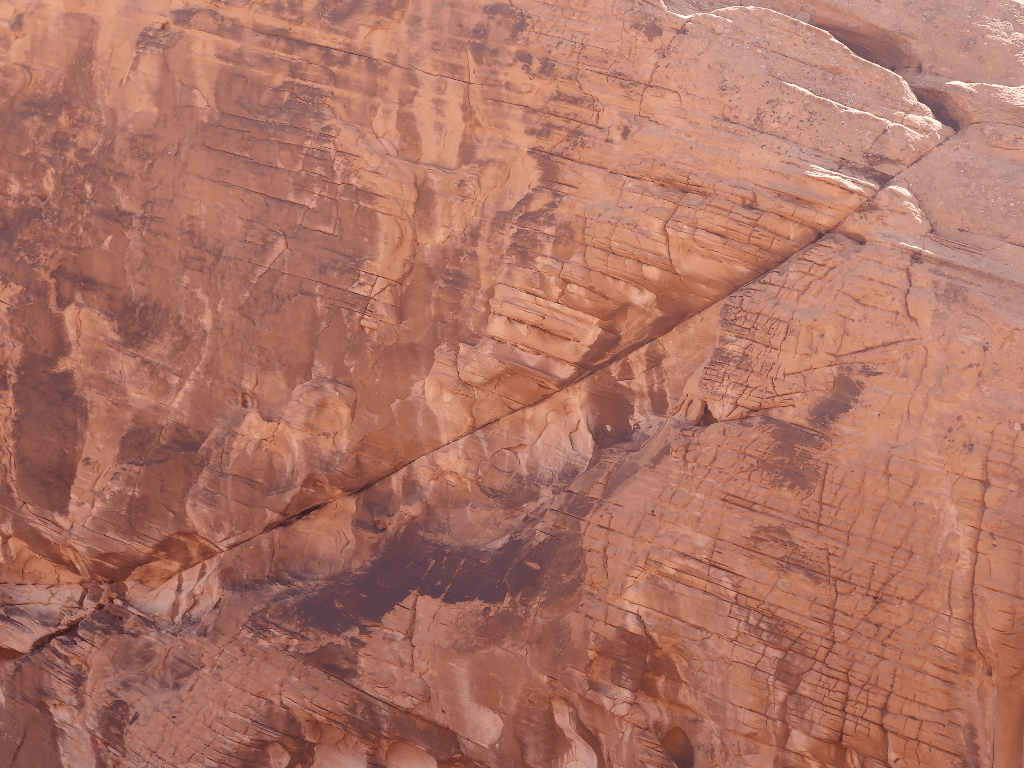}
  \end{subfigure}
  \begin{subfigure}{0.3013\linewidth}
    \includegraphics[width=\linewidth]{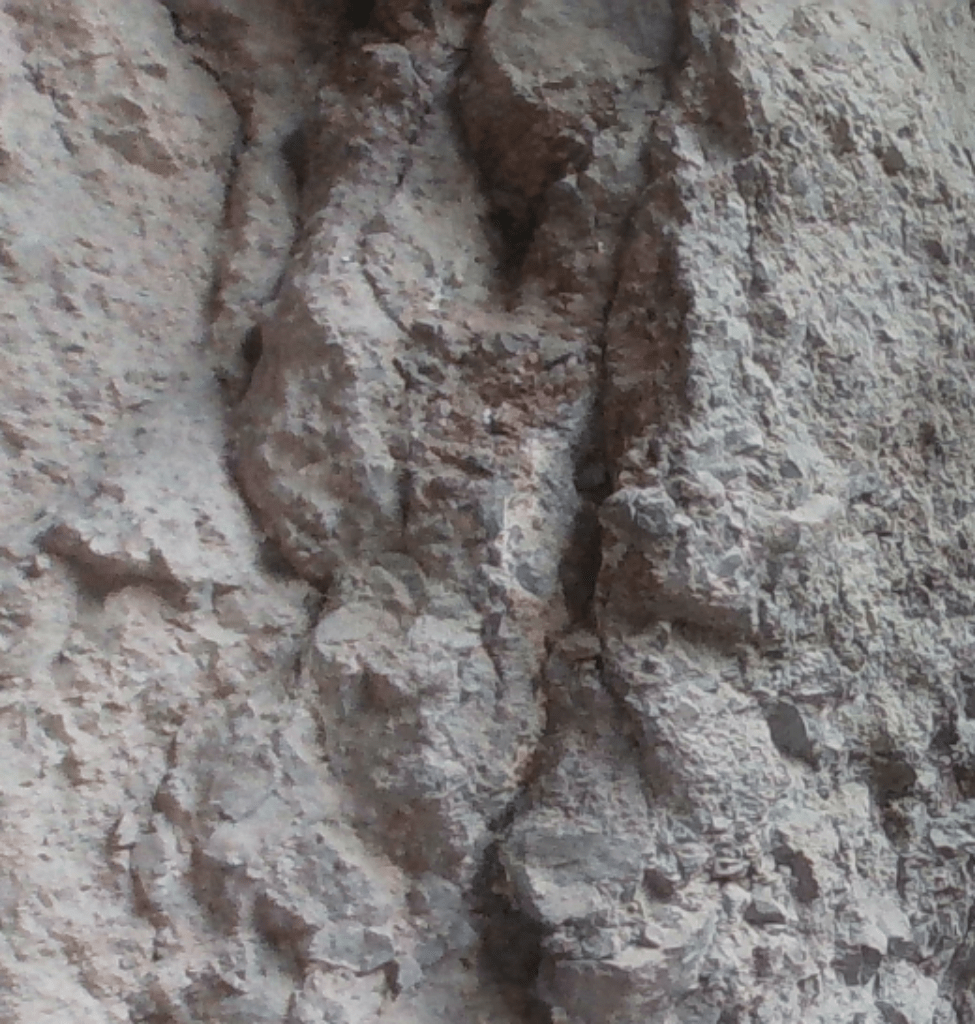}
  \end{subfigure}
  
  \vspace{0.01\linewidth}

  \begin{subfigure}{0.245\linewidth}
    \includegraphics[width=\linewidth]{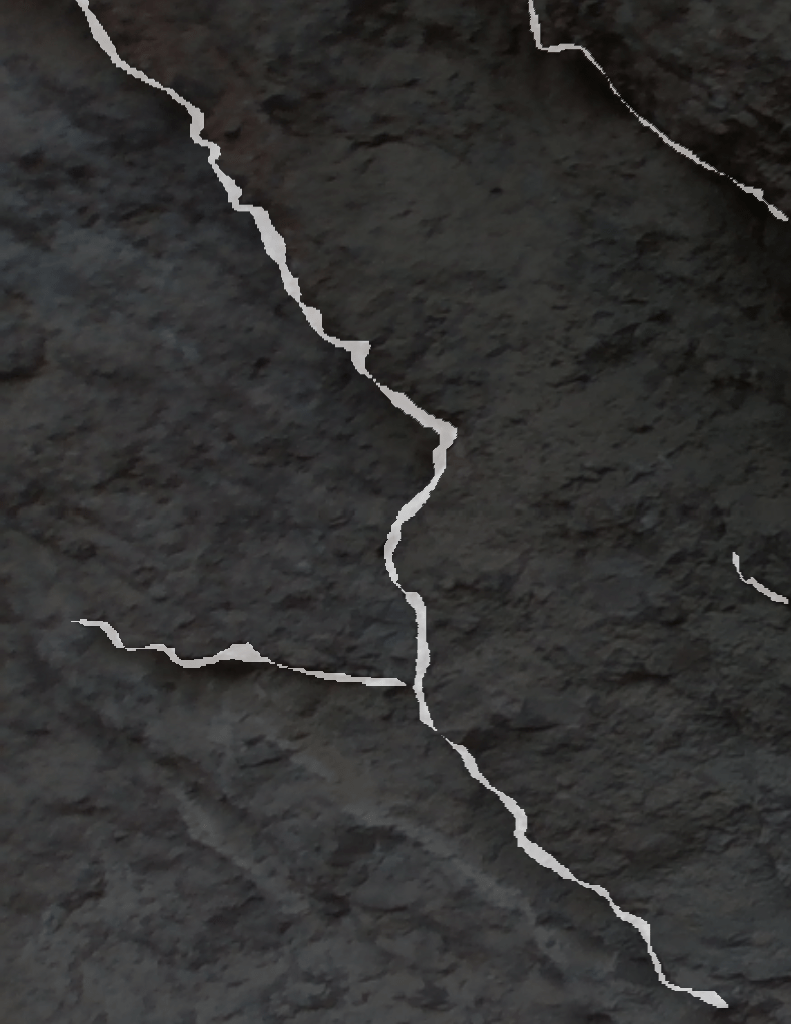}
  \end{subfigure}
  \begin{subfigure}{0.426\linewidth}
    \includegraphics[width=\linewidth]{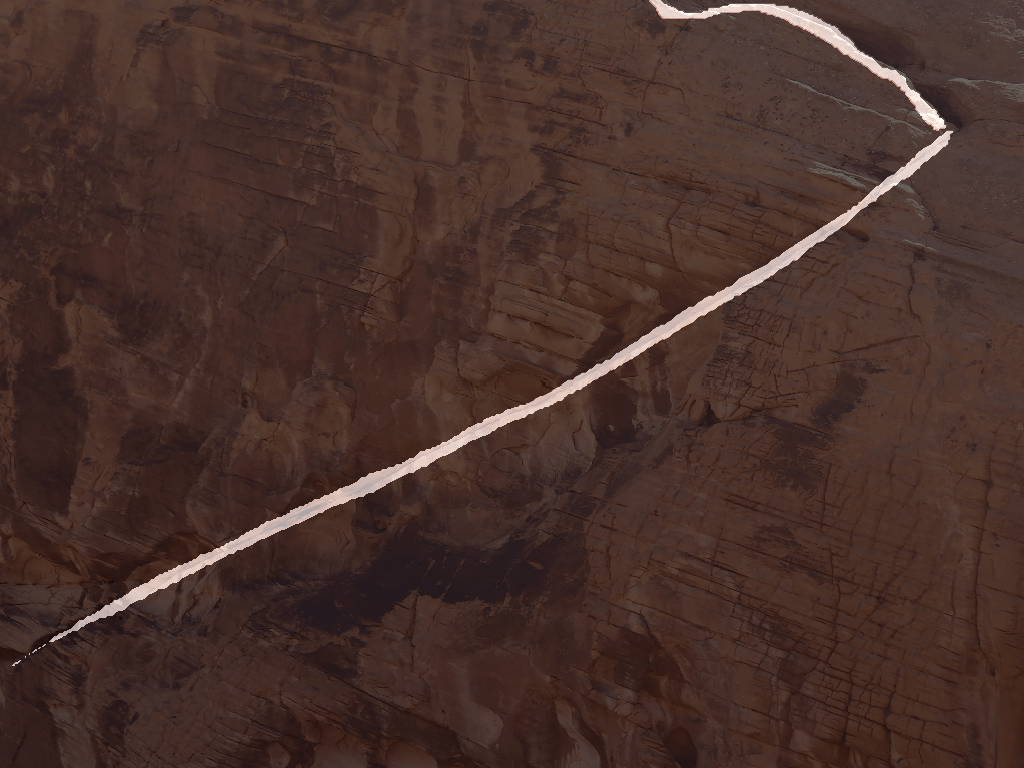}
  \end{subfigure}
  \begin{subfigure}{0.3018\linewidth}
    \includegraphics[width=\linewidth]{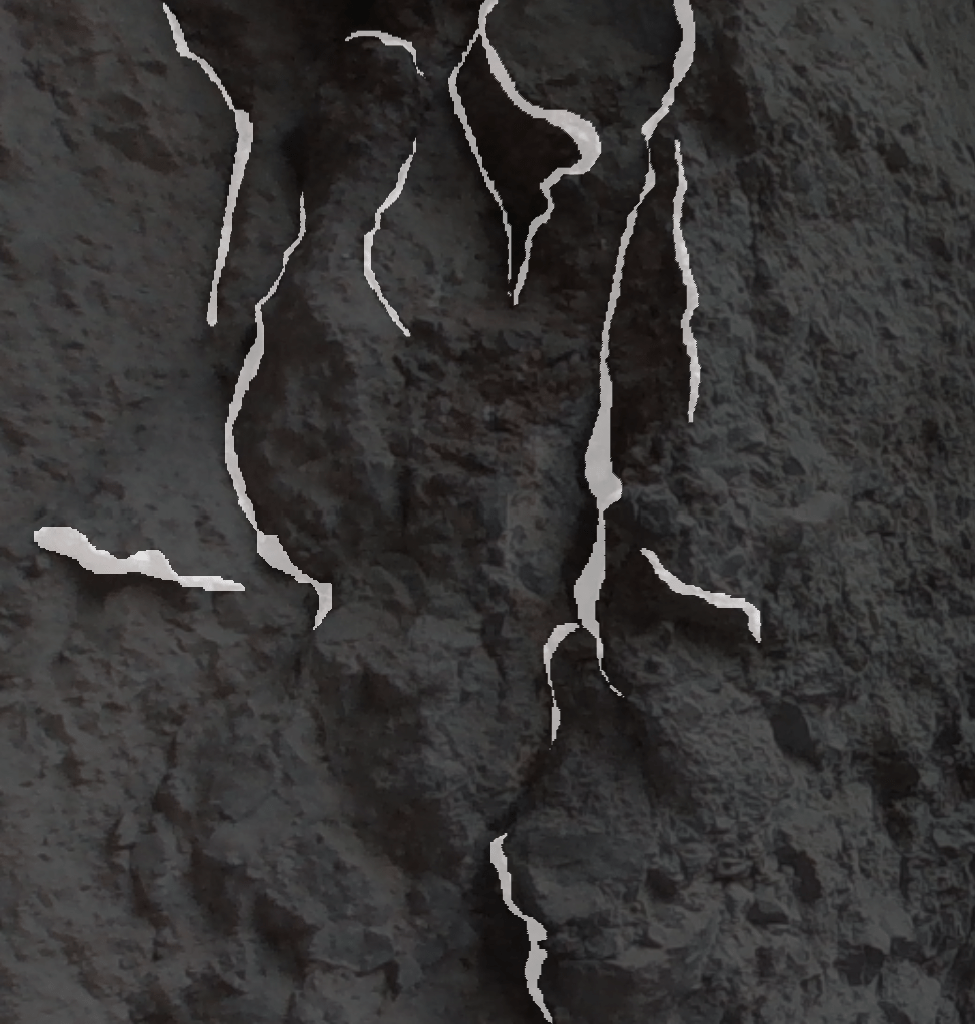}
  \end{subfigure}

  \caption{Crack dataset images. The first row shows raw images of cracks, and the second row shows annotated images.}
  \label{fig:real-data}
\end{figure}

We have collected an initial dataset of 234 rock wall photographs at Pinnacles National Park, Monument Valley, and Castle Rock using an Intel RealSense D435 camera. From these 234 images, we selected and annotated 100 images that presented interesting cracks and edges. Figure \ref{fig:real-data} illustrates some of the images.

\begin{figure}[]
    \includegraphics[width=.32\columnwidth]{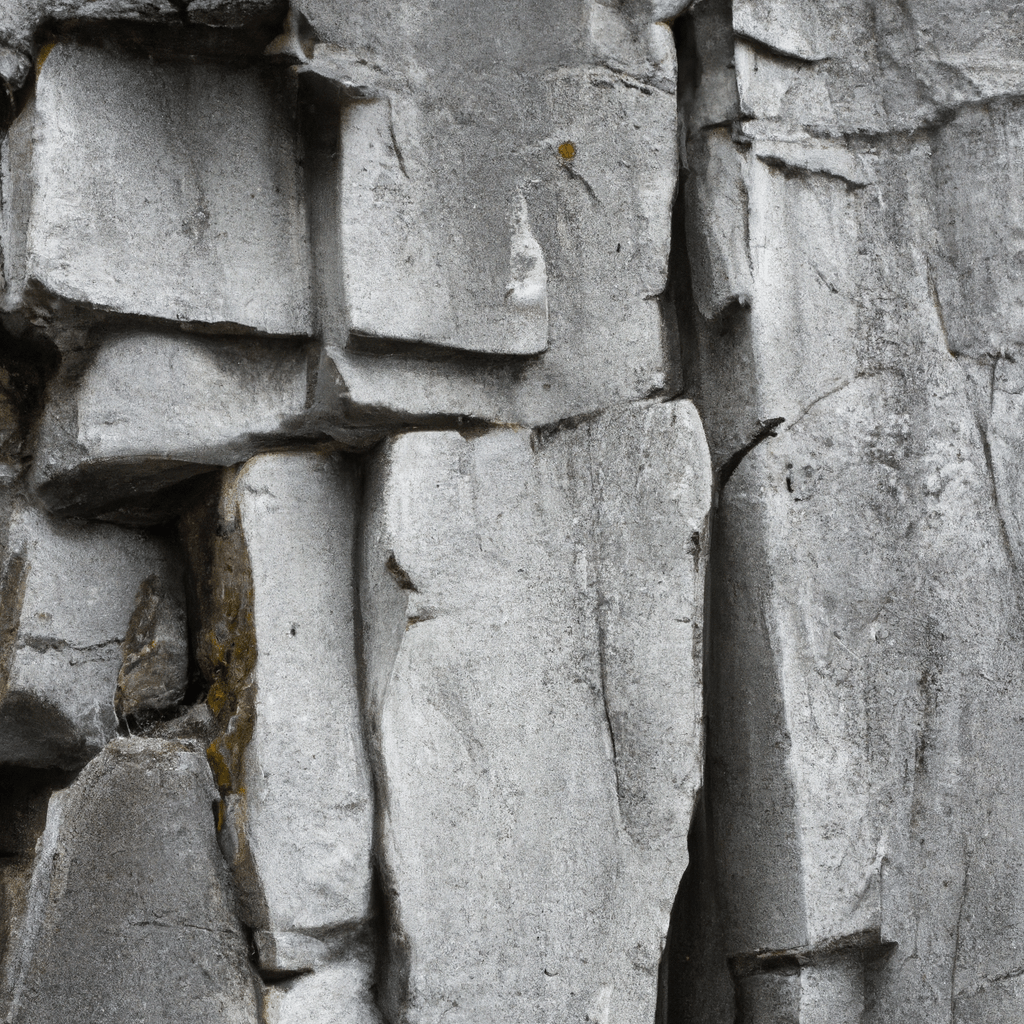}\hfill
  \includegraphics[width=.32\columnwidth]{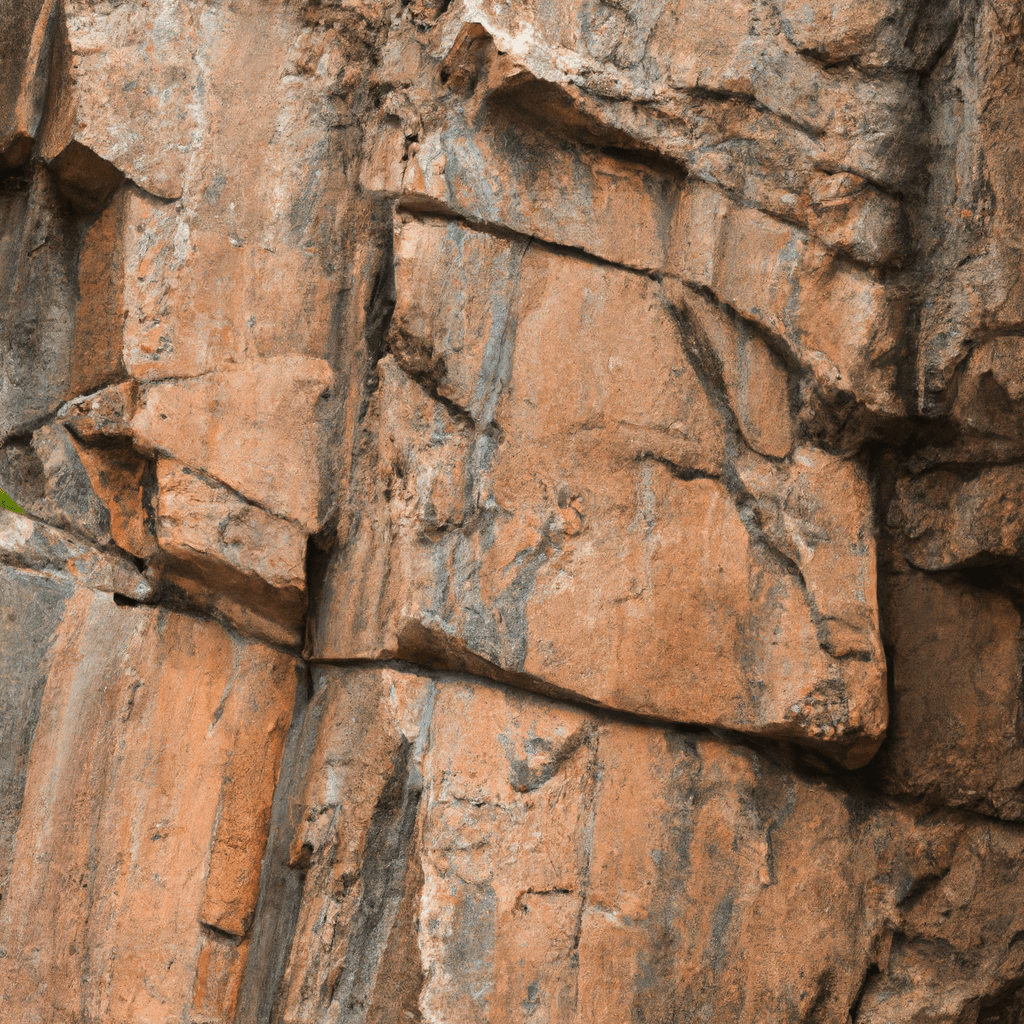}\hfill
  \includegraphics[width=.32\columnwidth]{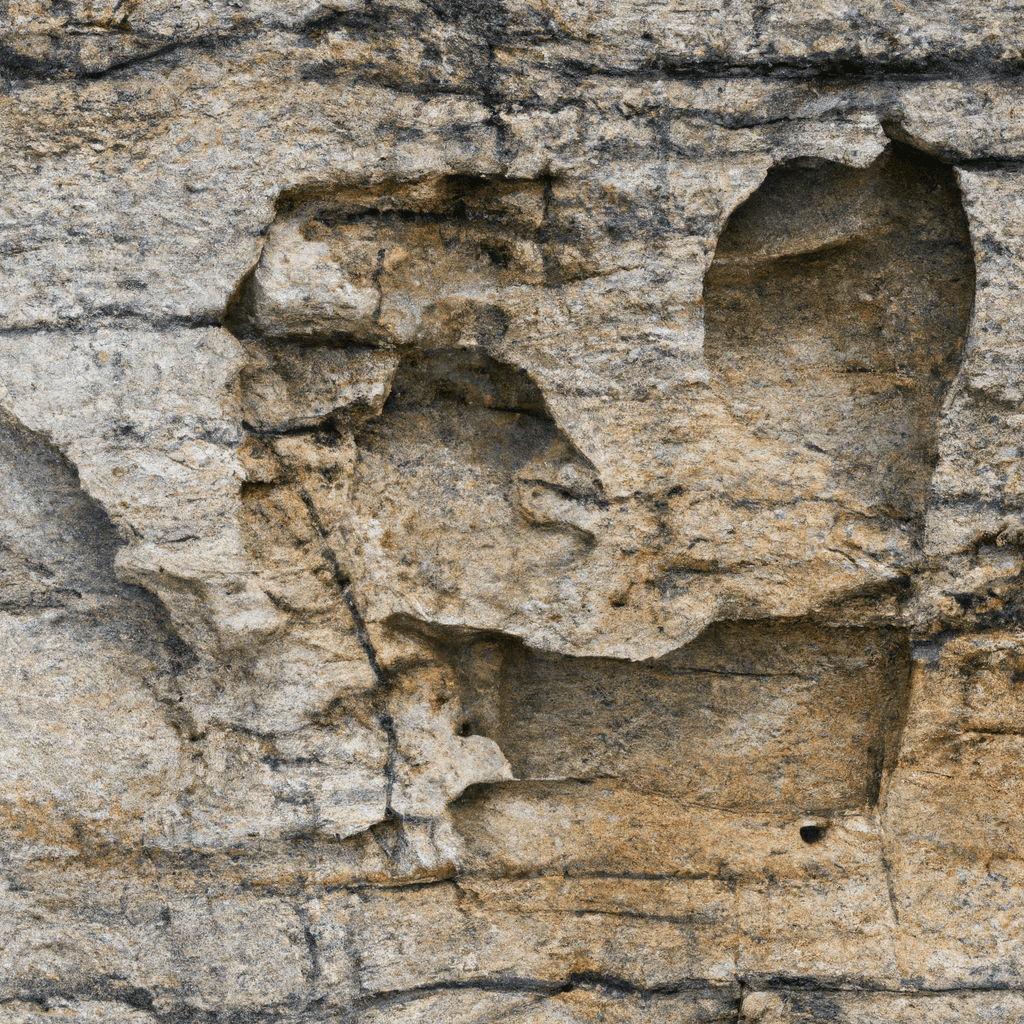}\hfill
  \caption{Examples of images generated using DALL-E 2\cite{dall-e}.}
  \label{fig:generated-images}
\end{figure}

\subsubsection{Generated data}
\label{sec:result-dataset-generated}
In order to add more diverse data, we also generated some images using DALL-E 2 \cite{dall-e}. The prompt used was, \textit{"A natural rock face with no apparent edges, cracks or holds."}. We asked for no visible edges to have more natural-looking rock faces (Figure \ref{fig:generated-images}). 

We generated and annotated an additional 136 images, which we combined into a dataset of real and generated images composed of 50 real images for evaluation and 186 images for training, including 50 real and 136 generated ones. In the rest of this paper, we only consider this augmented dataset as our dataset of 100 real images was too small to evaluate.

\subsection{Metrics Evaluation}
\label{sec:result-metric}
For comparison purposes, we define the combined metric, $\text{LineAcc}_{\text{comb.}}$ (combined):
\begin{equation}
\begin{split}
    \text{LineAcc}_{\text{comb.}} &= 2\text{LineAcc}_{\text{pos.}}
    + 0.5\text{LineAcc}_{\text{width}}\\
    &+ 0.5\text{LineAcc}_{\text{length}} + 0.5\text{Dice} + 0.5\text{IOU}
\end{split}
\end{equation}
The coefficients were chosen manually to prioritize detecting cracks in the correct locations, but not necessarily with very precise width or length.

\begin{figure}[h]
  \captionsetup[subfigure]{justification=centering, labelformat=empty} 
  \centering

  \begin{subfigure}{0.24\linewidth}
    \caption{Image}
    \includegraphics[width=\linewidth]{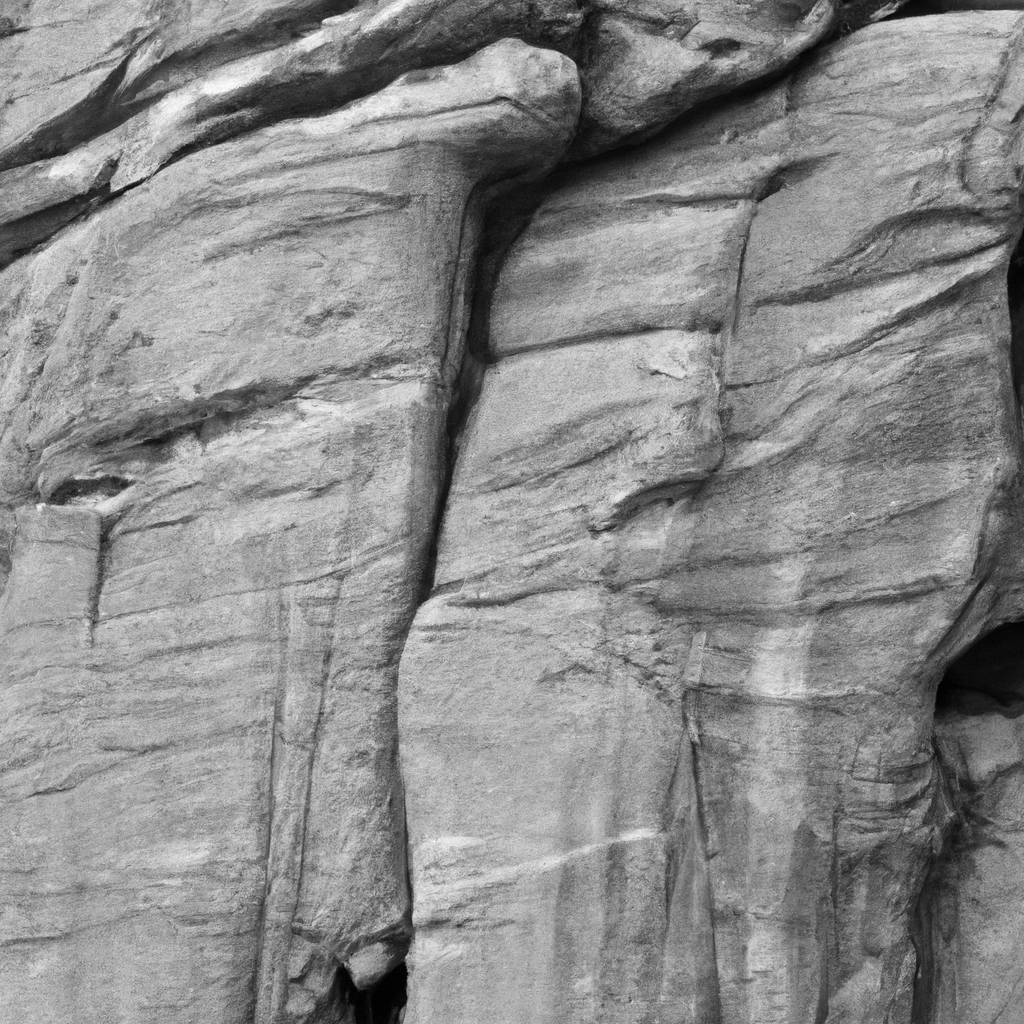}
  \end{subfigure}
  \begin{subfigure}{0.24\linewidth}
    \caption{Label}
    \includegraphics[width=\linewidth]{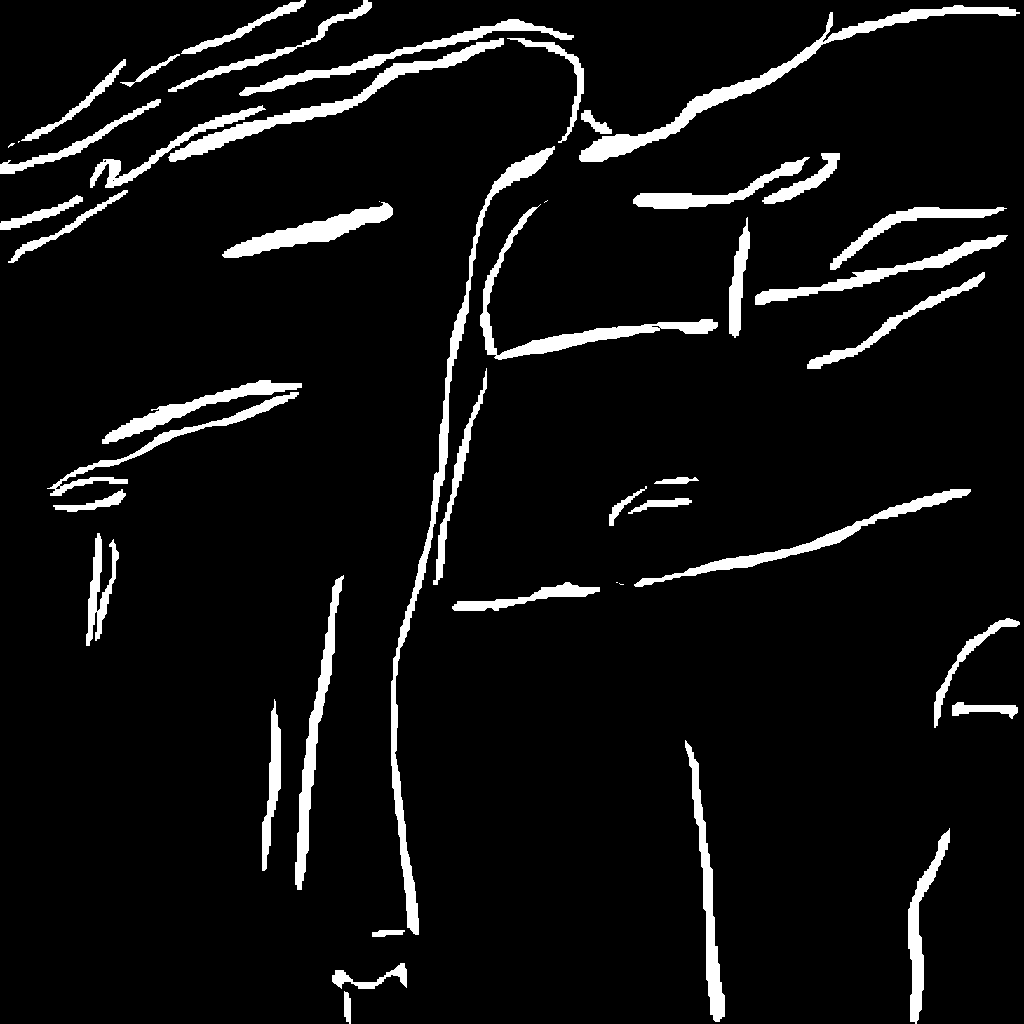}
  \end{subfigure}
  \begin{subfigure}{0.24\linewidth}
    \caption{Pred. 1}
    \includegraphics[width=\linewidth]{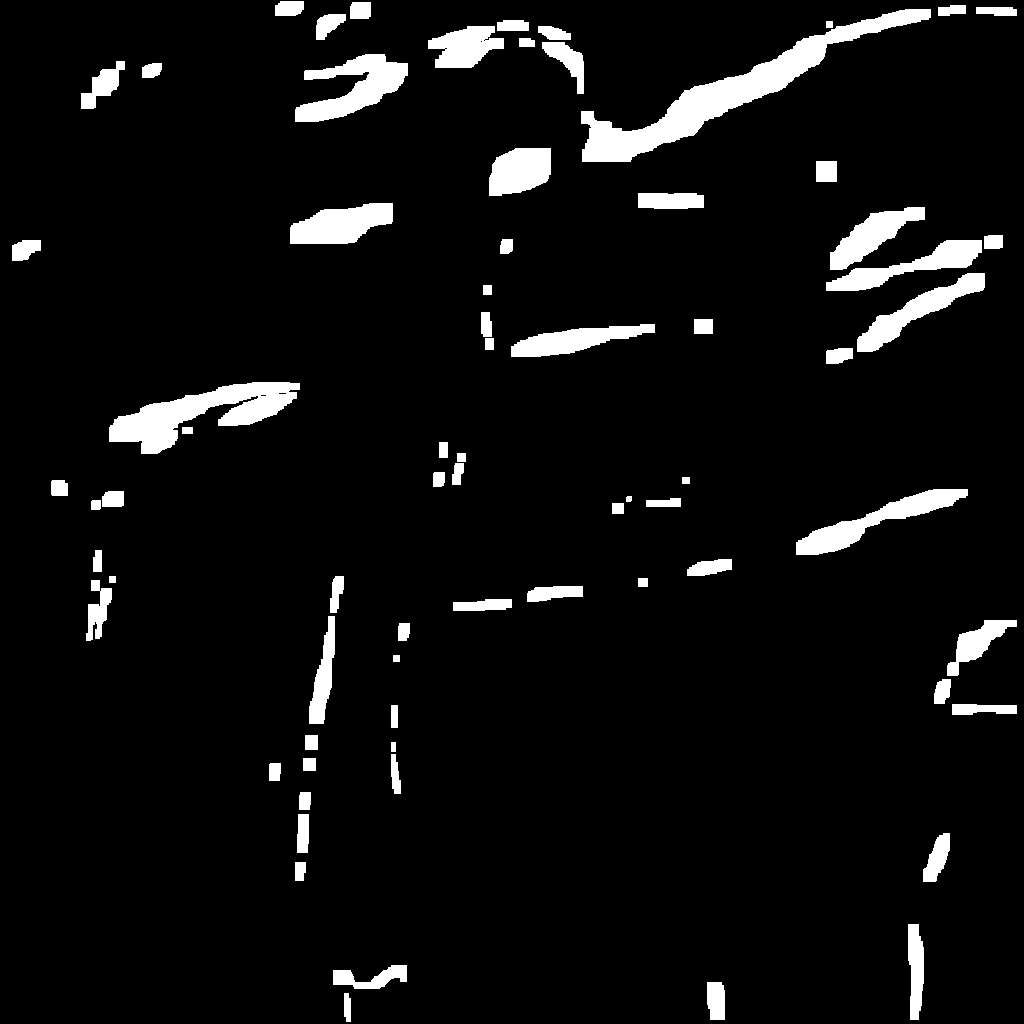}
  \end{subfigure}
  \begin{subfigure}{0.24\linewidth}
    \caption{Pred. 2}
    \includegraphics[width=\linewidth]{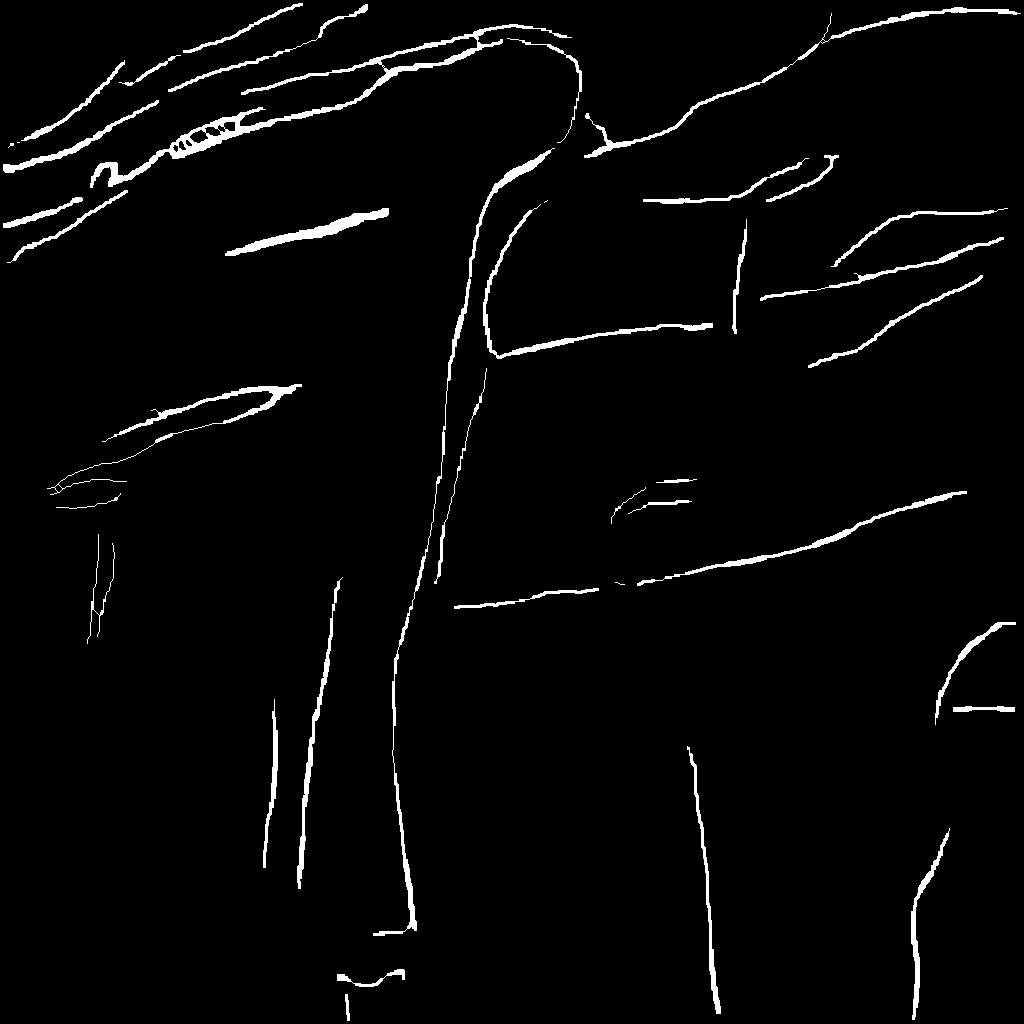}
  \end{subfigure}

\vspace{0.01\linewidth}

    \begin{subfigure}{0.24\linewidth}
    \includegraphics[width=\linewidth]{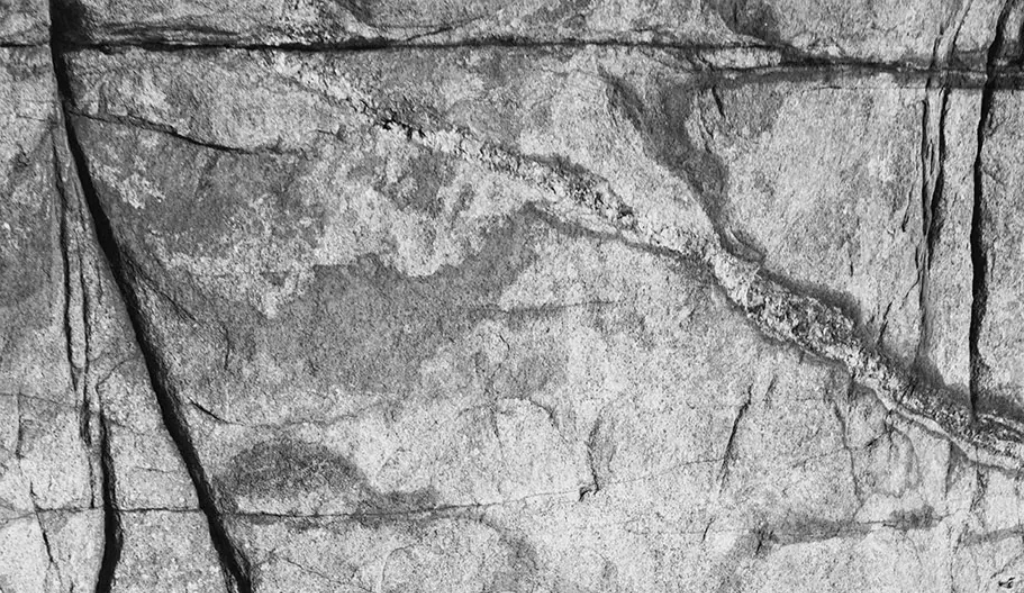}
  \end{subfigure}
  \begin{subfigure}{0.24\linewidth}
    \includegraphics[width=\linewidth]{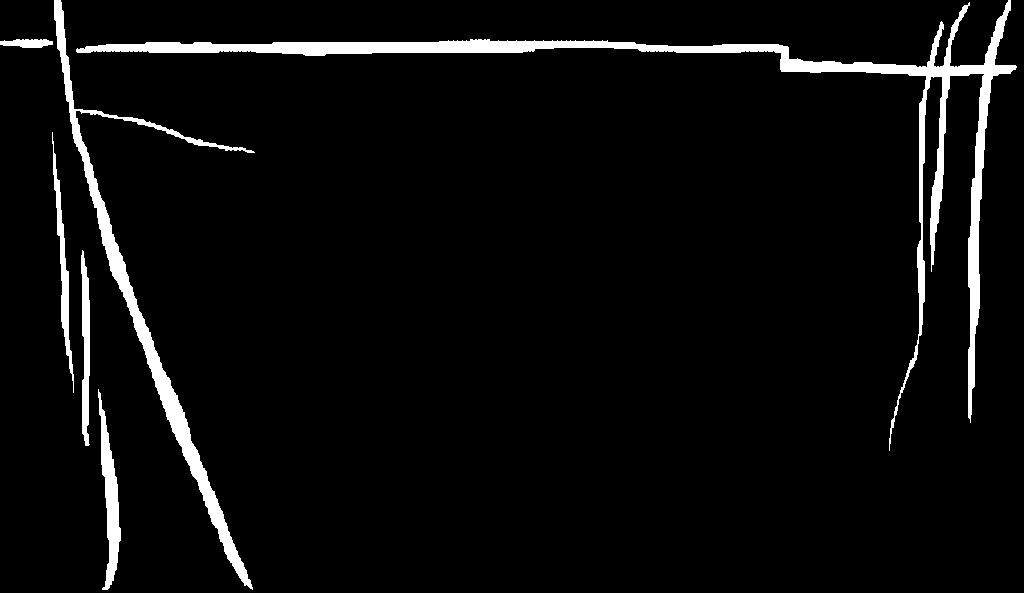}
  \end{subfigure}
  \begin{subfigure}{0.24\linewidth}
    \includegraphics[width=\linewidth]{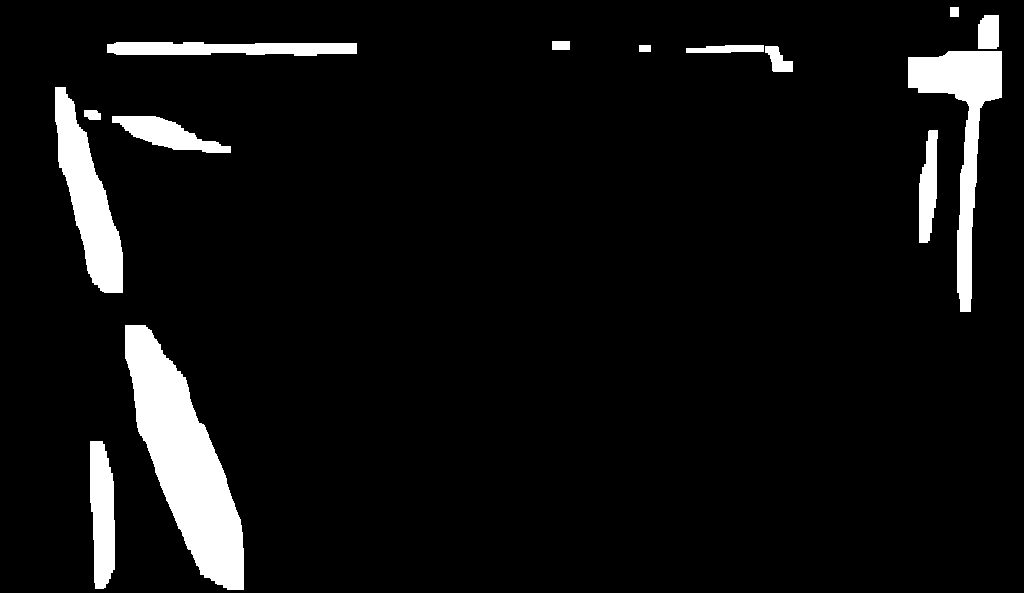}
  \end{subfigure}
  \begin{subfigure}{0.24\linewidth}
    \includegraphics[width=\linewidth]{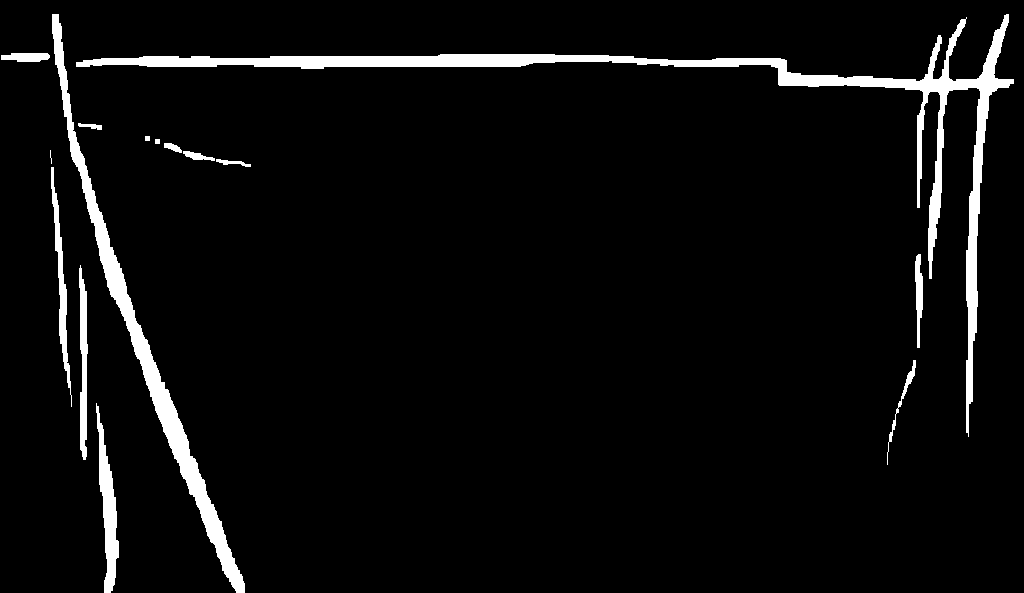}
  \end{subfigure}

\vspace{0.01\linewidth}

    \begin{subfigure}{0.24\linewidth}
    \includegraphics[width=\linewidth]{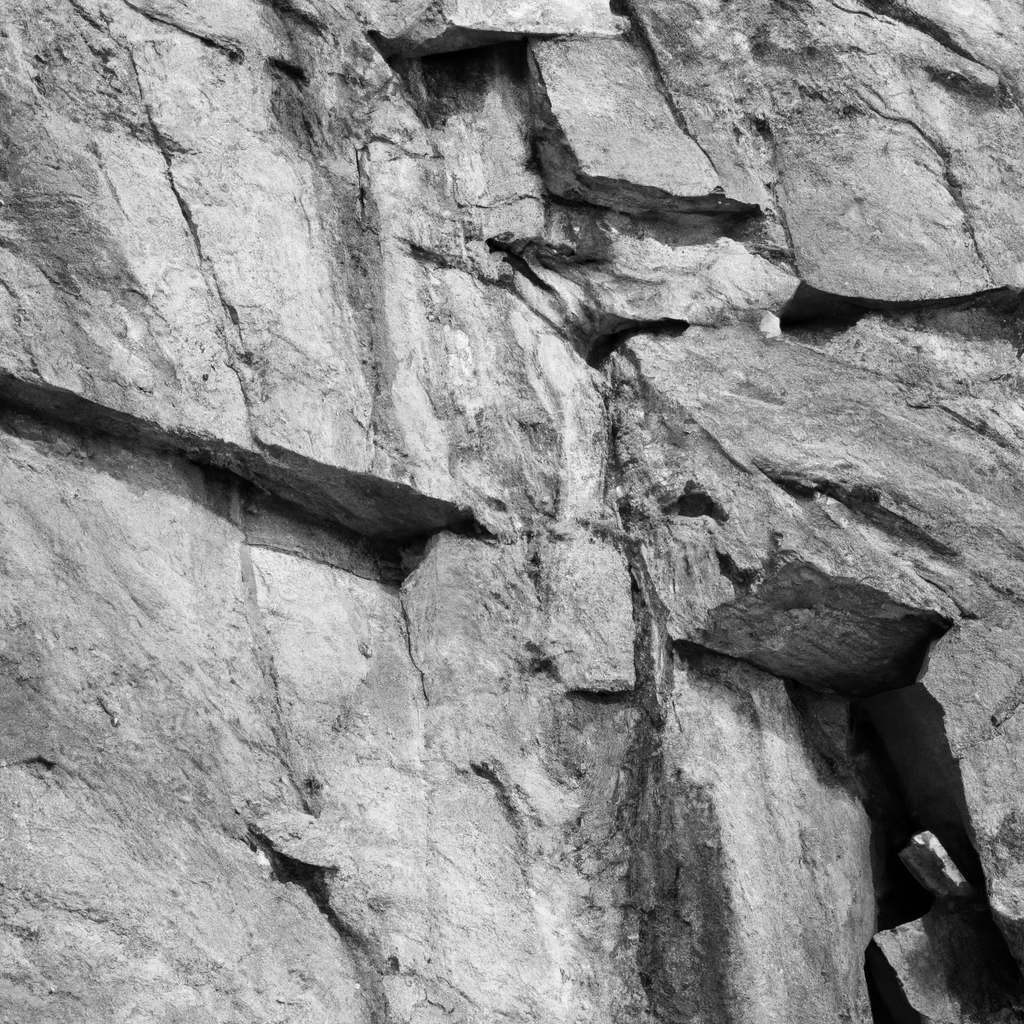}
  \end{subfigure}
  \begin{subfigure}{0.24\linewidth}
    \includegraphics[width=\linewidth]{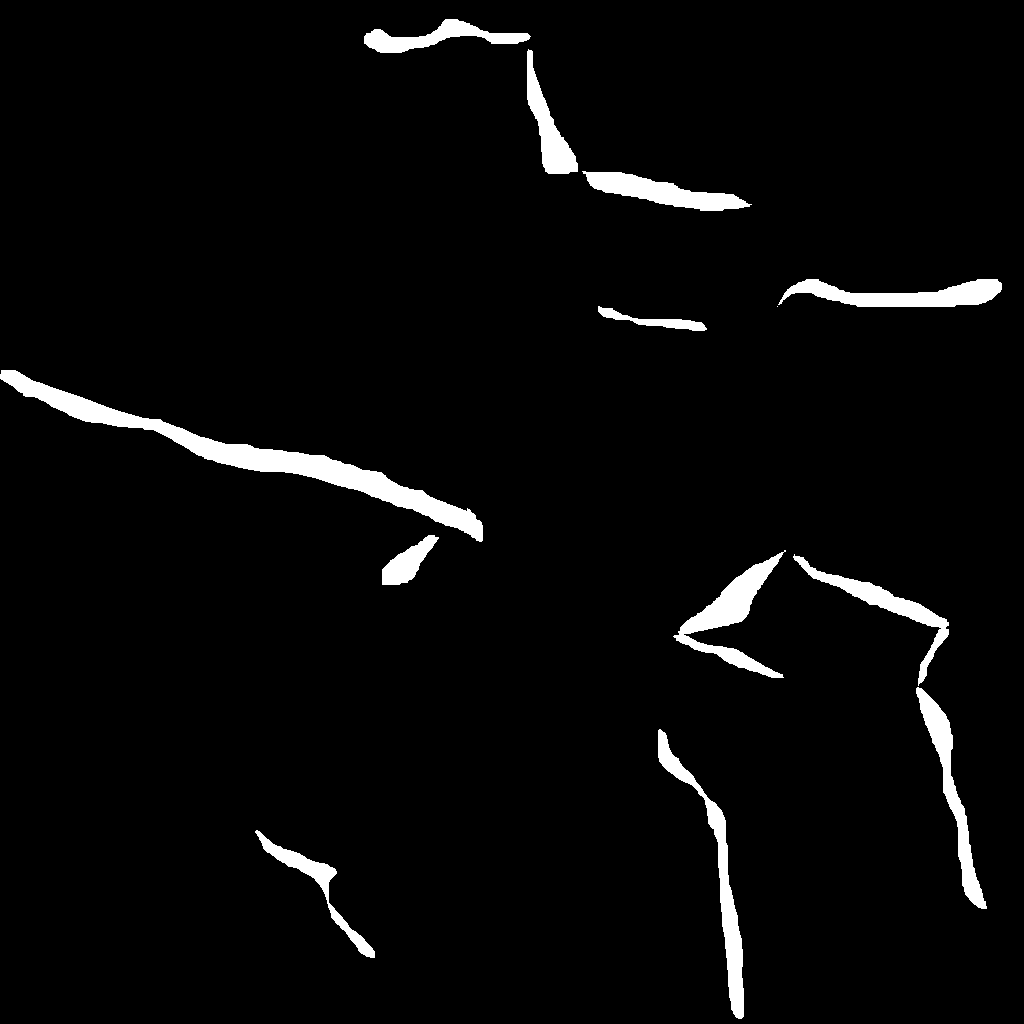}
  \end{subfigure}
  \begin{subfigure}{0.24\linewidth}
    \includegraphics[width=\linewidth]{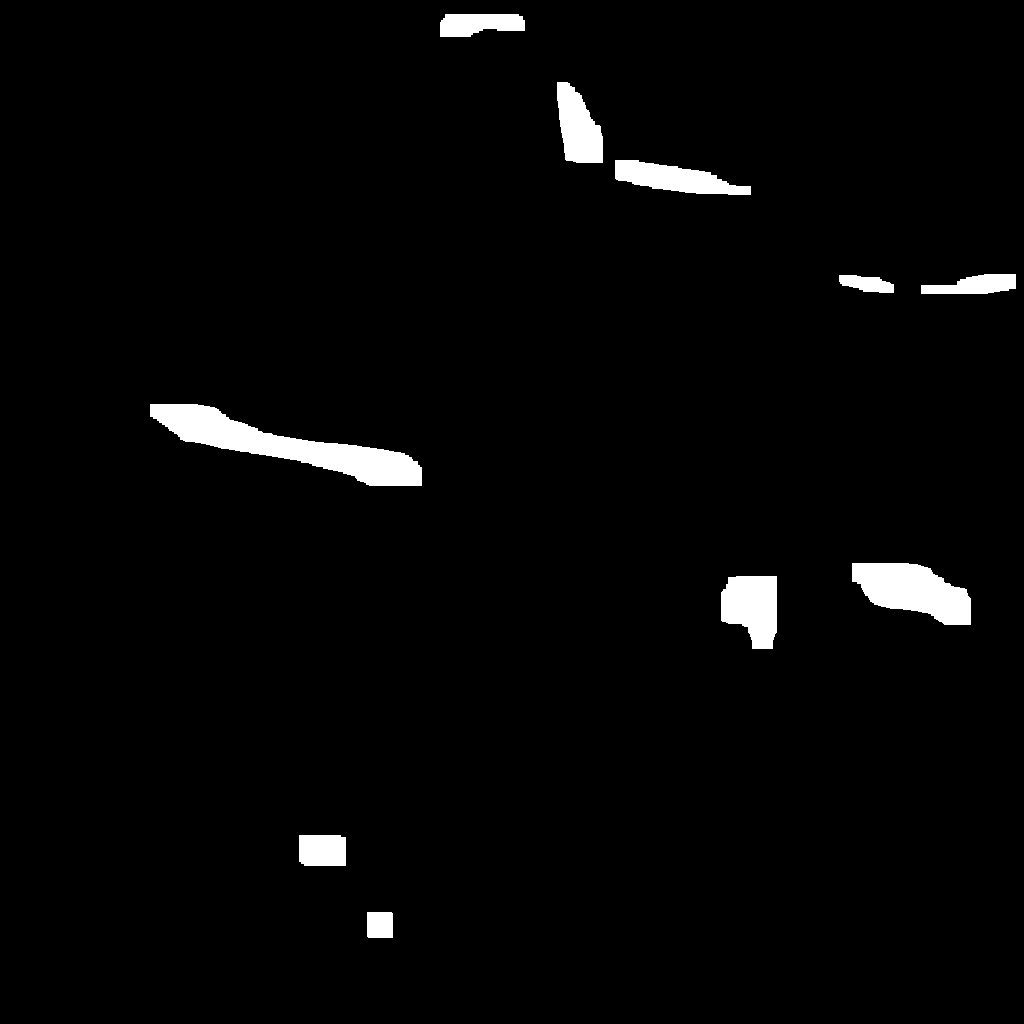}
  \end{subfigure}
  \begin{subfigure}{0.24\linewidth}
    \includegraphics[width=\linewidth]{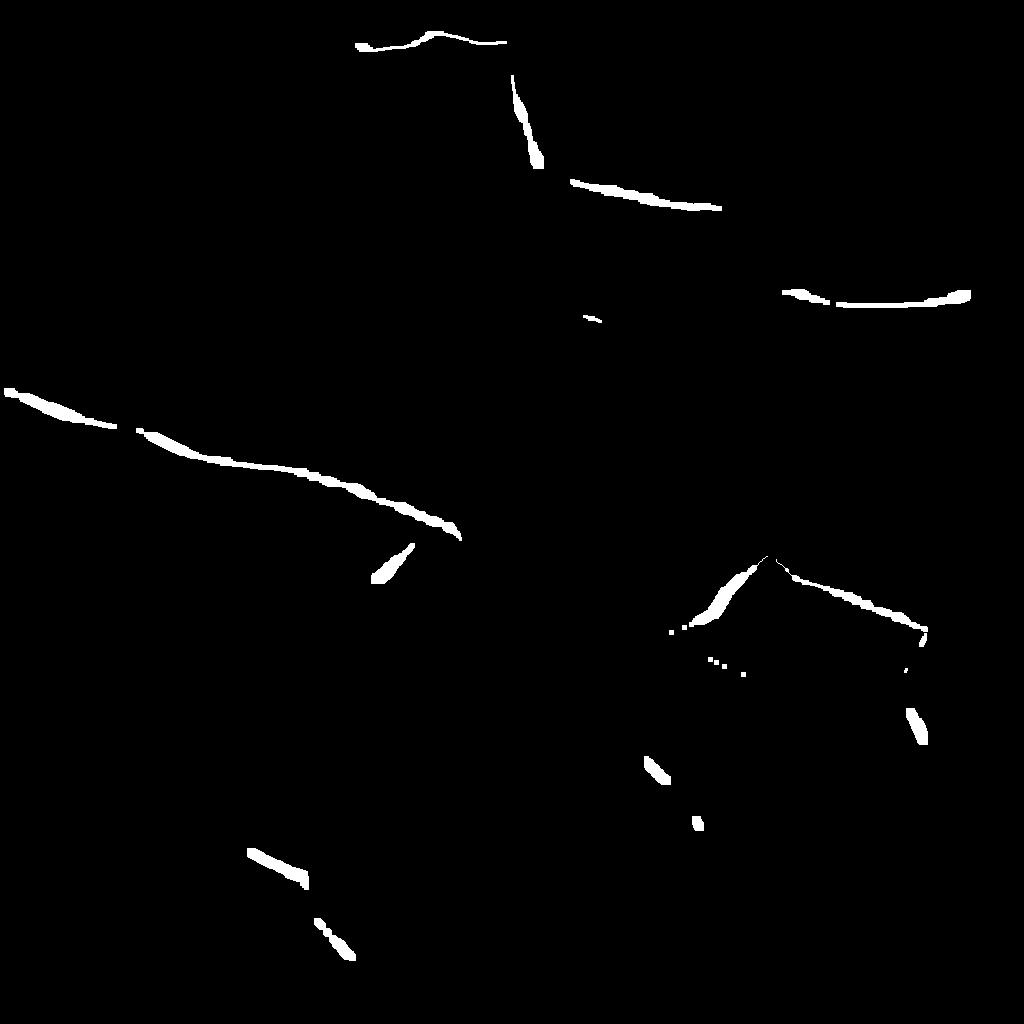}
  \end{subfigure}
  
  \caption{\textbf{Qualitative results:} We show two generated predictions with the same Dice score (absolute error of 1\%) but significantly different performance on $\text{LineAcc}_{\text{comb.}}$. From top to bottom, the Dice scores are 0.6, 0.4, and 0.2. The second prediction has a $\text{LineAcc}_{\text{comb.}}$ score that is $0.23$, $0.29$, and $0.22$ higher than the first, respectively.}
  \label{fig:result-metric}
\end{figure}

To evaluate our new metrics we generated two predictions with the same Dice score but different $\text{LineAcc}_{\text{comb.}}$ to check which one was visually better. The predictions were generated using random deformations as described in Section \ref{sec:result-gen-gen}. Figure \ref{fig:result-metric} shows that $\text{LineAcc}_{\text{comb.}}$ is less sensitive to the prediction width (first row), to small translations (second row), and rewards predictions with the correct line length (last row).

\subsection{Loss performance}
\label{sec:result-loss}
To answer the question: Does SKIL improve training, we trained a ViT-B network on the crack dataset and a combined dataset of blood vessels (Section \ref{sec:result-dataset}). The training was run on an NVIDIA RTX A6000 and averaged on 20 runs each for better precision. We used a linear learning scheduler that reaches the maximum at 25\% of the training. The model was trained for 100 epochs with a maximum learning rate of $1\times10^{-5}$ for the cracks dataset while it was trained for 200 epochs with a maximum learning rate of $5\times10^{-5}$ for the blood vessels dataset. All models were fine-tuned with the Adam\cite{adam} optimizer from a pre-trained model. For all experiments, the parameters used for the SKIL loss (both Dice and Product) were $s_{\text{border}} = 20$, $n_{\text{iter}}^{\text{max}} = 50$, $f = 0.82$ and $s = 10$. These parameters were qualitatively chosen and manually tuned to offer the behavior we wanted, and parameter search optimization will be part of future work. \\

\subsubsection{Quantitative results}
\label{sec:result-loss-quantitative}
For each training, we used a mixture of the Dice loss with a coefficient of 3 and the studied loss with a coefficient of 1 \textit{(For the Dice-only training we, therefore, had a coefficient of 4)}. In the tables below we will only reference the studied loss even though the Dice loss was also used for all of them.

\begin{table}[H]
\centering
    \scalebox{0.88}{
\ra{0.8}
\setlength\tabcolsep{3pt}
\begin{tabular}{@{}lcccccccc@{}}\toprule
& \multicolumn{2}{c}{\textbf{Classical}} & & \multicolumn{5}{c}{\textbf{LineAcc}} \\ 
\cmidrule{2-3} \cmidrule{5-9} 
& IOU & Dice && Pos. & Width & Length && Combined\\ \midrule
\phantom{ab} Dice & 0.255 & 0.396 ± 0.001 && 0.555 & 0.608 & 0.726 && 0.548 ± 0.001\\
\phantom{ab} CL-Dice & 0.253 & 0.394 ± 0.001 && 0.548 & 0.560 & 0.712 && 0.536 ± 0.001\\
\phantom{ab} SKIL-Dice  & \textbf{0.261} & \textbf{0.404 ± 0.001} && \textbf{0.570} & \textbf{0.648} & \textbf{0.764} && \textbf{0.569 ± 0.001}\\
\phantom{ab} SKIL-Prod.  & 0.261 & 0.403 ± 0.001 && 0.560 & 0.627 & 0.721 && 0.553 ± 0.001\\
\bottomrule
\end{tabular}
}
\caption{\textbf{Quantitative results:} Comparison of different metrics on the cracks dataset (averaged over 20 runs) after training for 100 epochs with 4 losses: Dice, CL-Dice, SKIL-Dice \textit{(This Work)} and SKIL-Product \textit{(This Work)}. The standard error is reported for the Dice and $\text{LineAcc}_{\text{comb.}}$ score.}
\label{tab:results-cracks}
\end{table}

\begin{table}[h]
\centering
    \scalebox{0.86}{
\ra{0.8}
\setlength\tabcolsep{3pt}
\begin{tabular}{@{}lcccccccc@{}}\toprule
& \multicolumn{2}{c}{\textbf{Classical}} & & \multicolumn{5}{c}{\textbf{LineAcc}} \\ 
\cmidrule{2-3} \cmidrule{5-9} 
& IOU & Dice && Pos. & Width & Length && Combined\\ \midrule
\phantom{ab} Dice & 0.264 & 0.400 ± 0.001 && 0.746 & 0.631 & 0.643 && 0.618 ± 0.001\\
\phantom{ab} CL-Dice & \textbf{0.270} & \textbf{0.410 ± 0.001} && 0.745 & 0.584 & 0.657 && 0.618 ± 0.001\\
\phantom{ab} SKIL-Dice  & 0.257 & 0.389 ± 0.001 && \textbf{0.823} & \textbf{0.685} & \textbf{0.746} && \textbf{0.679 ± 0.001}\\
\phantom{ab} SKIL (Prod) & 0.259 & 0.391 ± 0.001 && 0.803 & 0.665 & 0.711 && 0.661 ± 0.001\\
\bottomrule
\end{tabular}
}
\caption{\textbf{Quantitative results:} Comparison of different metrics on the \textbf{vessels datasets (aggregation of DRIVE, HRF, STARE, CHASE-DB1)} \textit{(averaged over 20 runs)} after training for 200 epochs with 4 losses: Dice, CL-Dice, SKIL-Dice \textit{(This Work)} and SKIL-Product \textit{(This Work)}. The standard error is reported for the Dice and $\text{LineAcc}_{\text{comb.}}$ score.}
\label{tab:results-vessels}
\end{table}

Table \ref{tab:results-cracks} shows that SKIL-Dice and SKIL-Product both improve the Dice score and $\text{LineAcc}_{\text{comb.}}$ score. The results of both SKIL variants are close, but SKIL-Dice offers slightly better performances, so in the rest of this paper we will only consider SKIL-Dice and refer to it as SKIL. 

Table \ref{tab:results-vessels} shows that for the blood vessels dataset, SKIL slightly decreases the Dice score but the Combined score is substantially increased. We report visual predictions in the qualitative results later, Section \ref{sec:result-loss-vessels}.

\subsubsection{Qualitative results on cracks}
\label{sec:result-loss-cracks}

\begin{figure}[h]
  \captionsetup[subfigure]{justification=centering, labelformat=empty} 
  \centering

  \begin{subfigure}{0.19\linewidth}
    \caption{Image}
    \includegraphics[width=\linewidth]{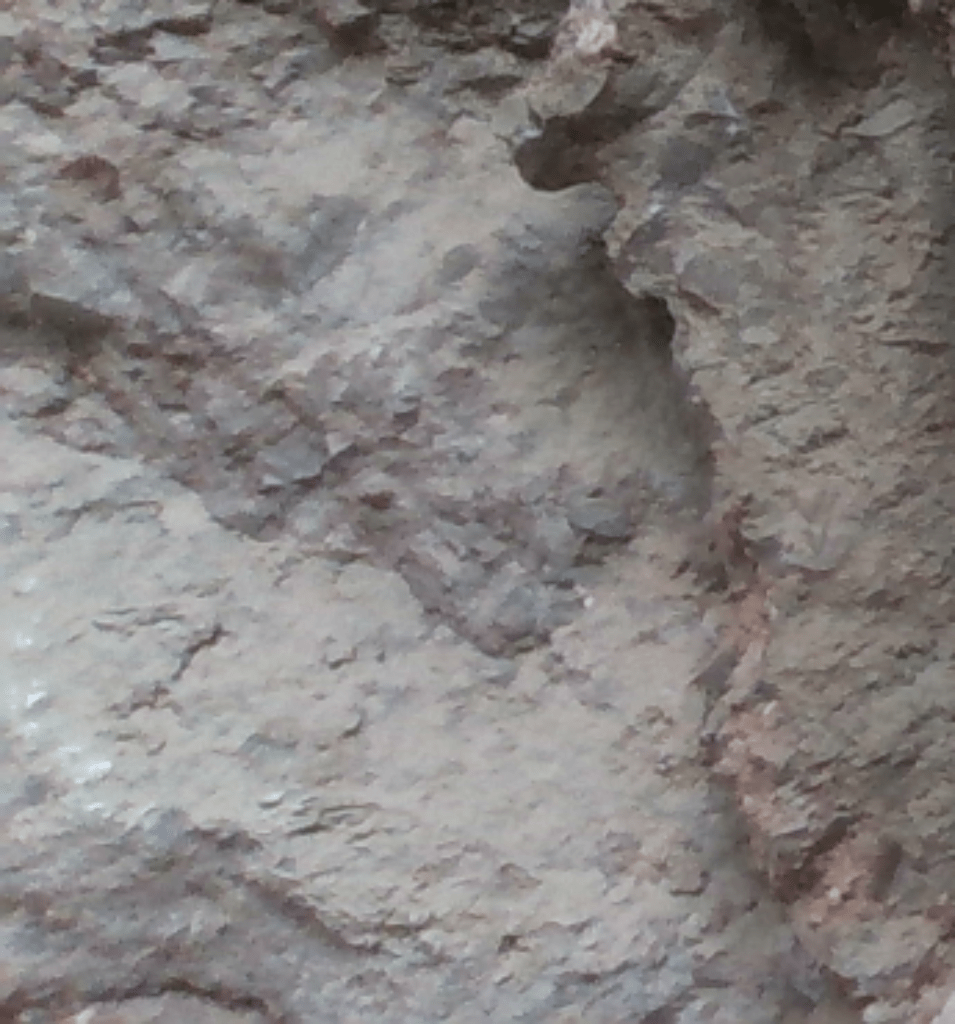}
  \end{subfigure}
  \begin{subfigure}{0.19\linewidth}
    \caption{Label}
    \includegraphics[width=\linewidth]{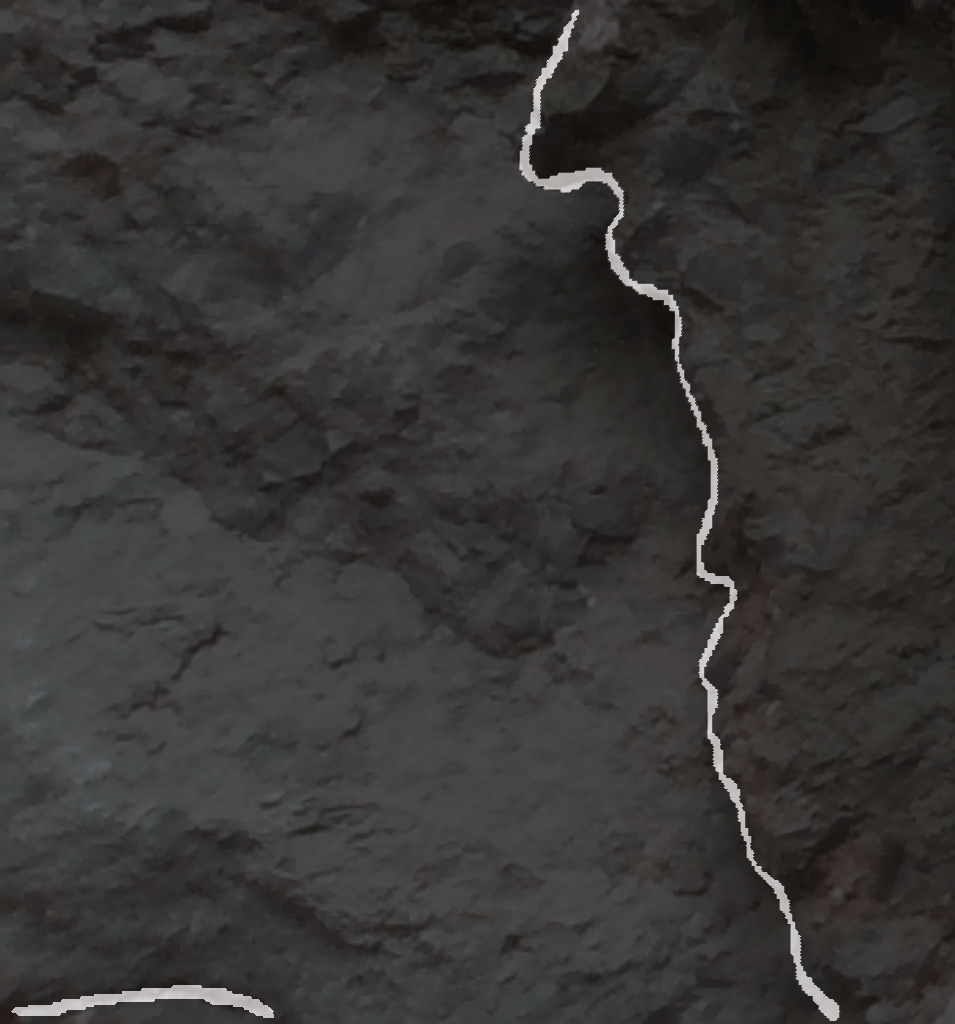}
  \end{subfigure}
  \begin{subfigure}{0.19\linewidth}
    \caption{Dice}
    \includegraphics[width=\linewidth]{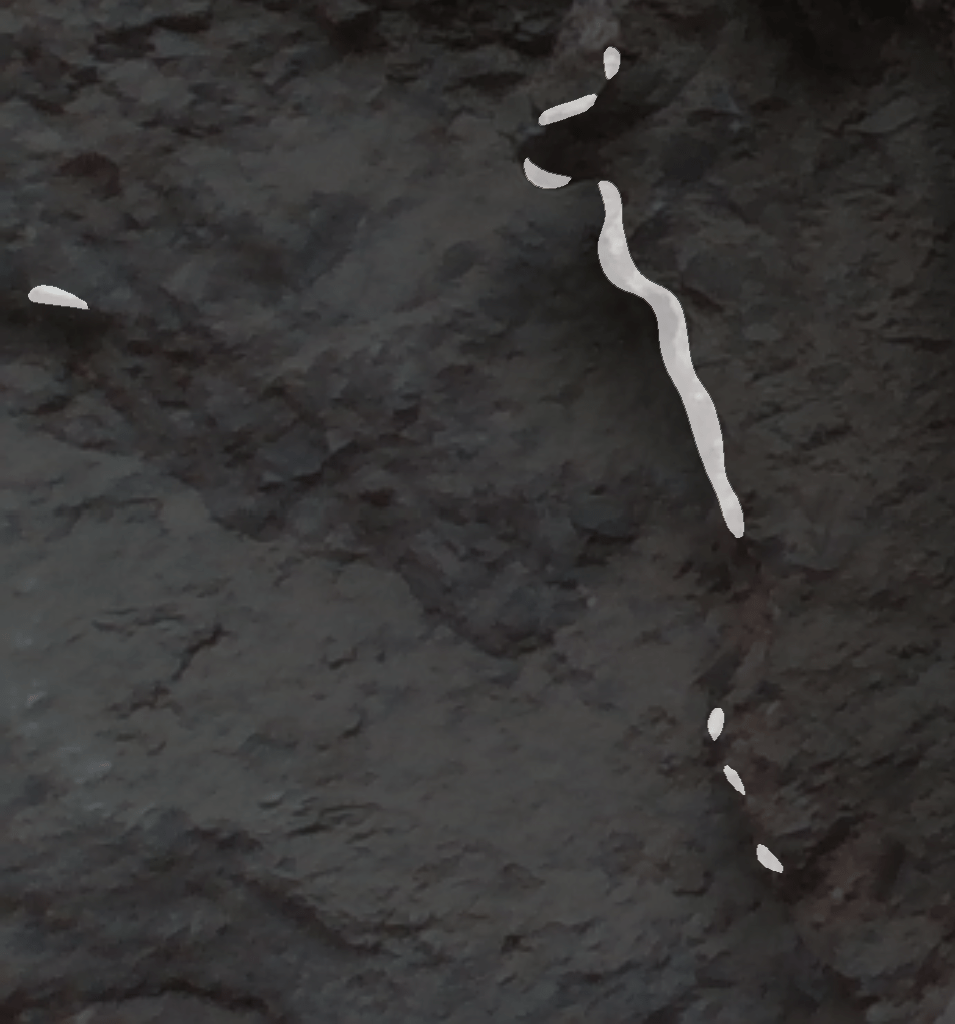}
  \end{subfigure}
  \begin{subfigure}{0.19\linewidth}
    \caption{CL-Dice}
    \includegraphics[width=\linewidth]{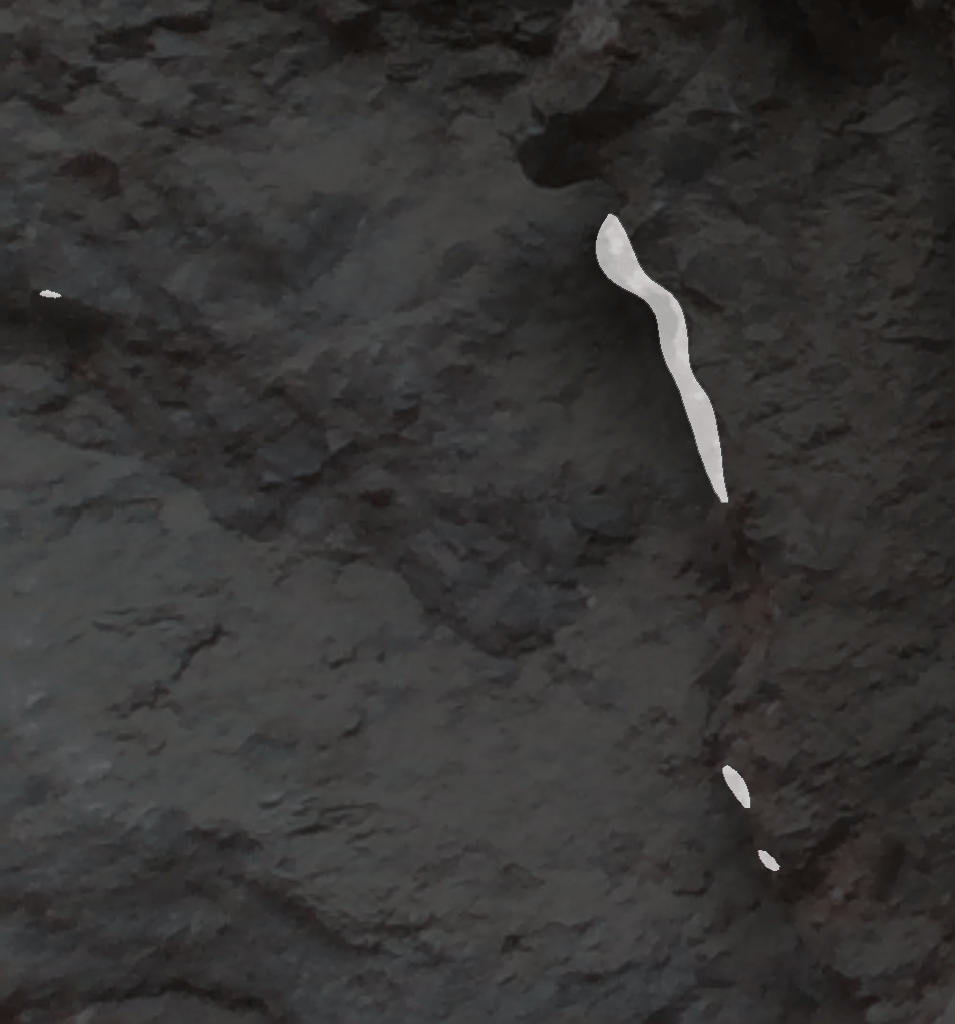}
  \end{subfigure}
  \begin{subfigure}{0.19\linewidth}
    \caption{SKIL}
    \includegraphics[width=\linewidth]{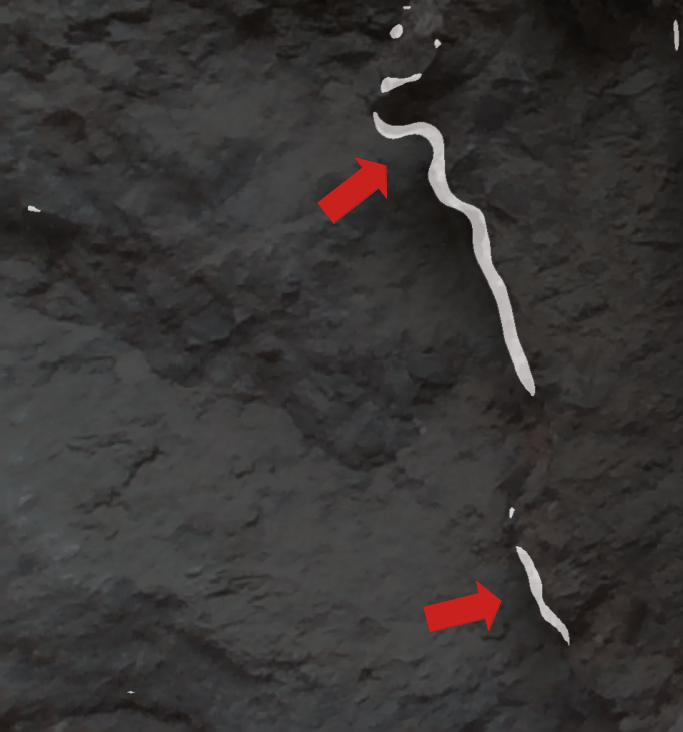}
  \end{subfigure}

\vspace{0.01\linewidth}

    \begin{subfigure}{0.19\linewidth}
    \includegraphics[width=\linewidth]{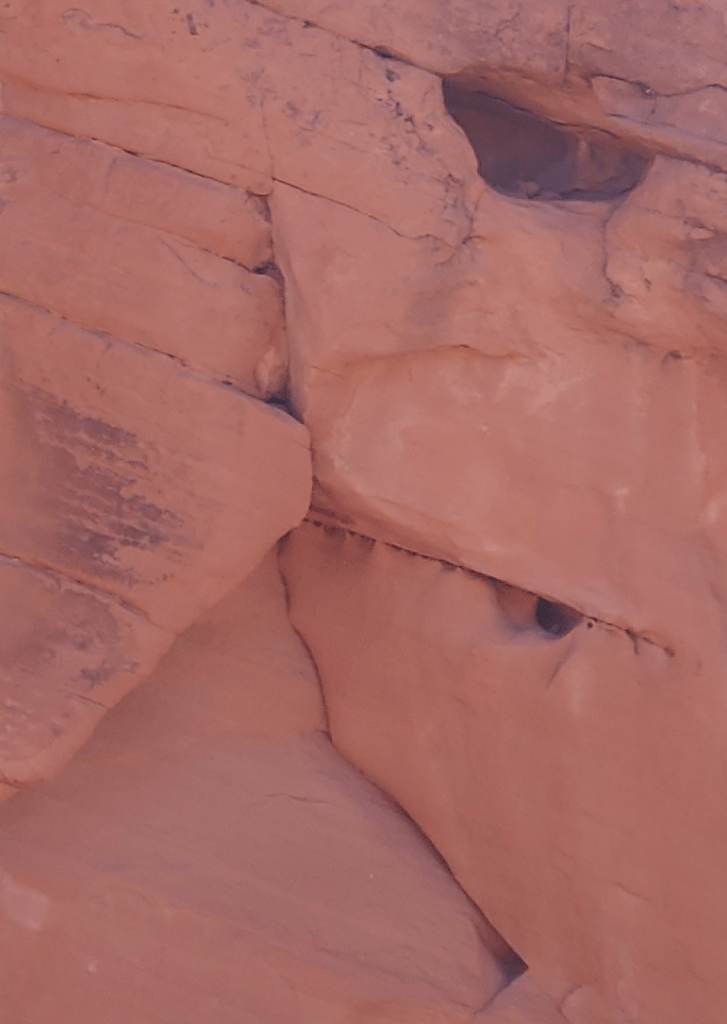}
  \end{subfigure}
  \begin{subfigure}{0.19\linewidth}
    \includegraphics[width=\linewidth]{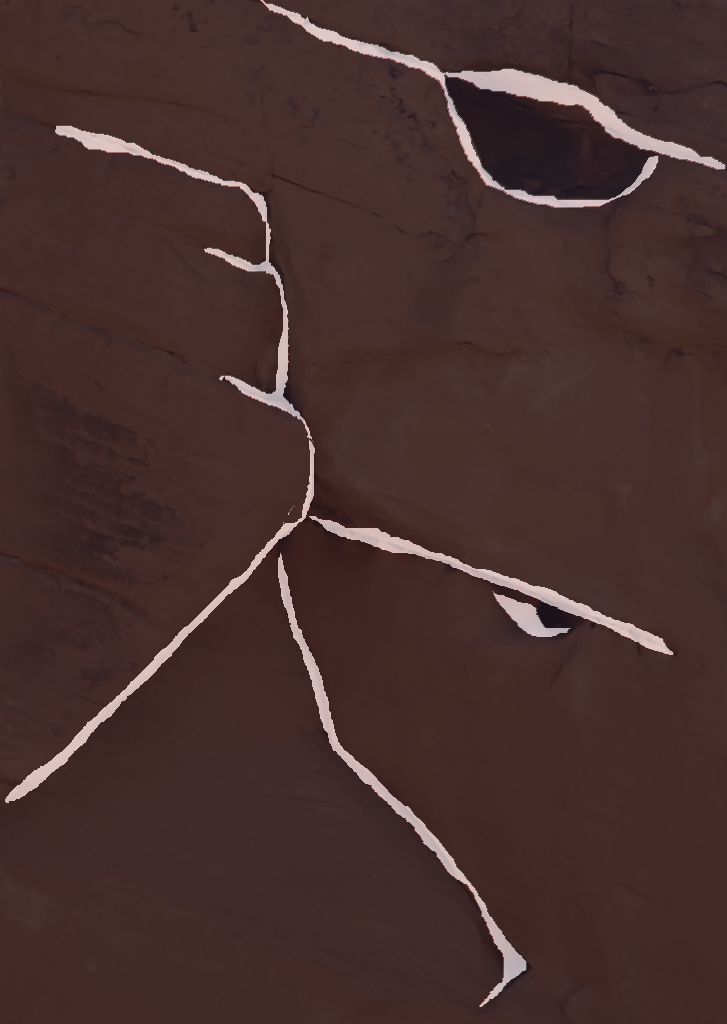}
  \end{subfigure}
  \begin{subfigure}{0.19\linewidth}
    \includegraphics[width=\linewidth]{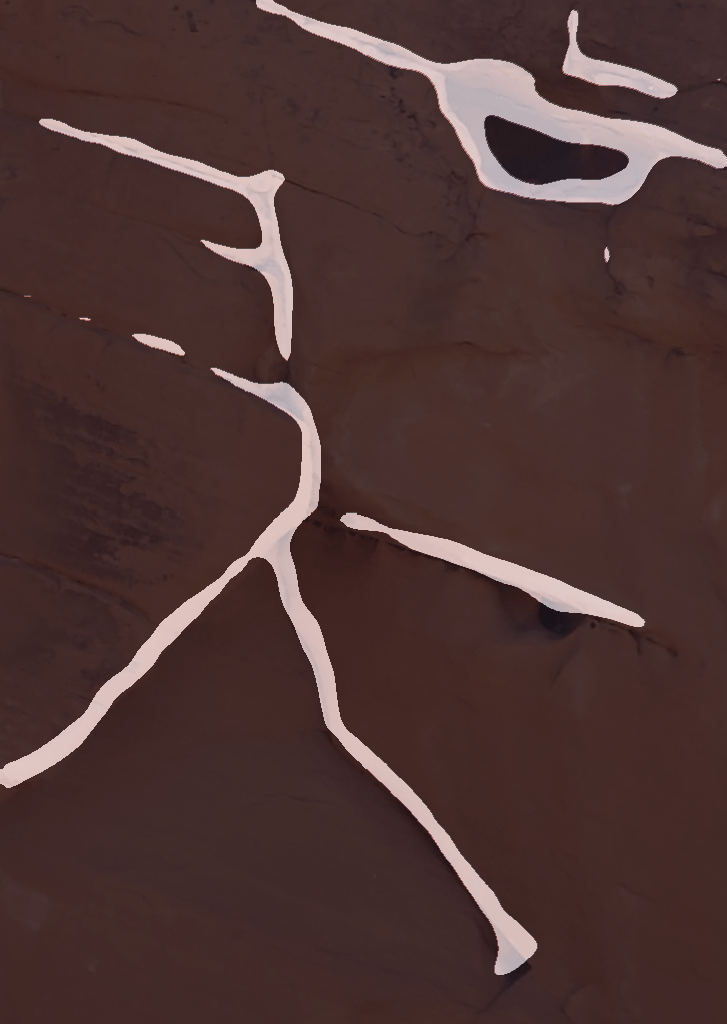}
  \end{subfigure}
  \begin{subfigure}{0.19\linewidth}
    \includegraphics[width=\linewidth]{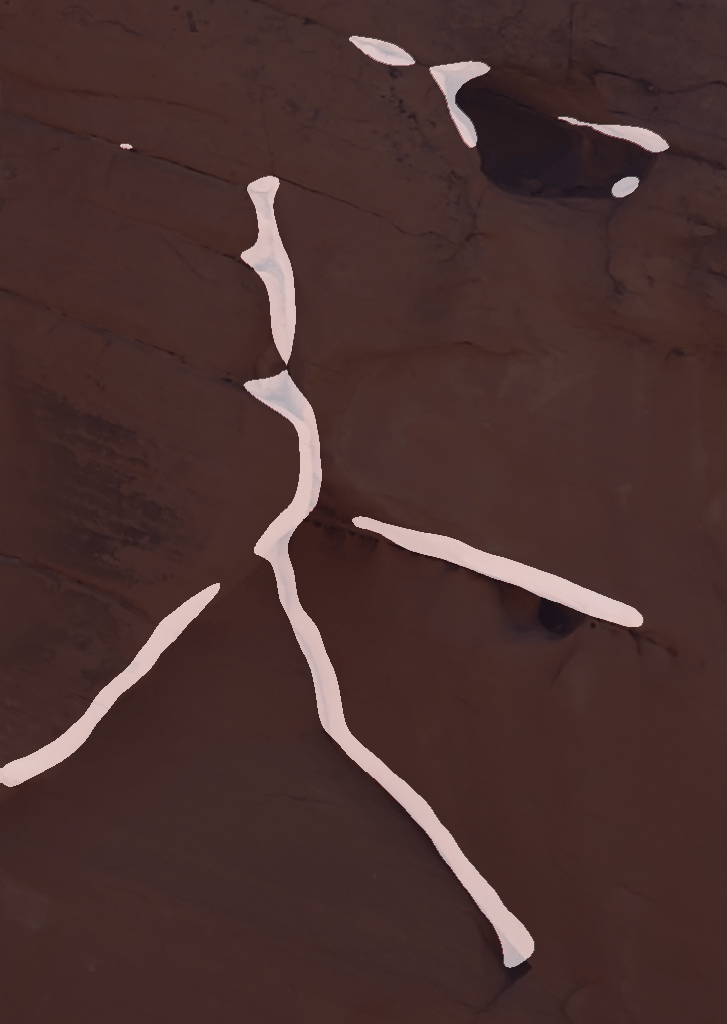}
  \end{subfigure}
  \begin{subfigure}{0.19\linewidth}
    \includegraphics[width=\linewidth]{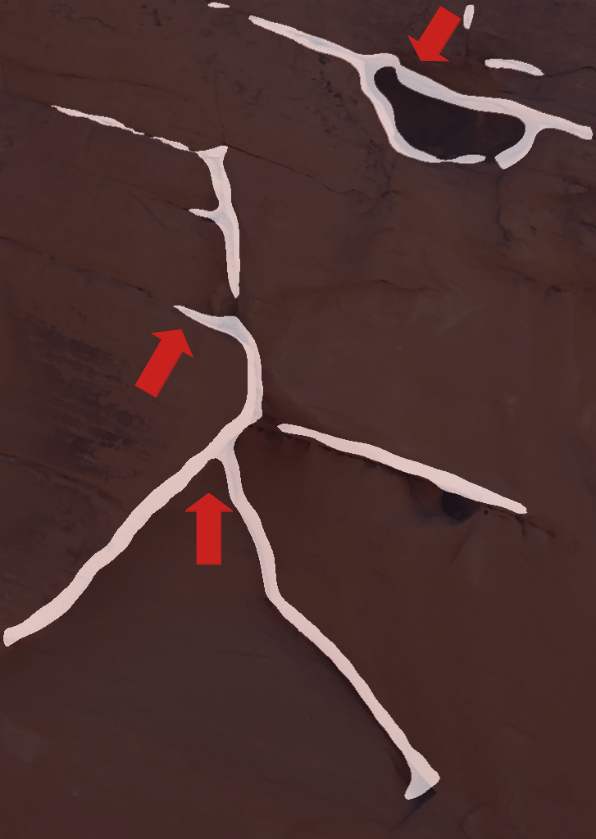}
  \end{subfigure}

\vspace{0.01\linewidth}

    \begin{subfigure}{0.19\linewidth}
    \includegraphics[width=\linewidth]{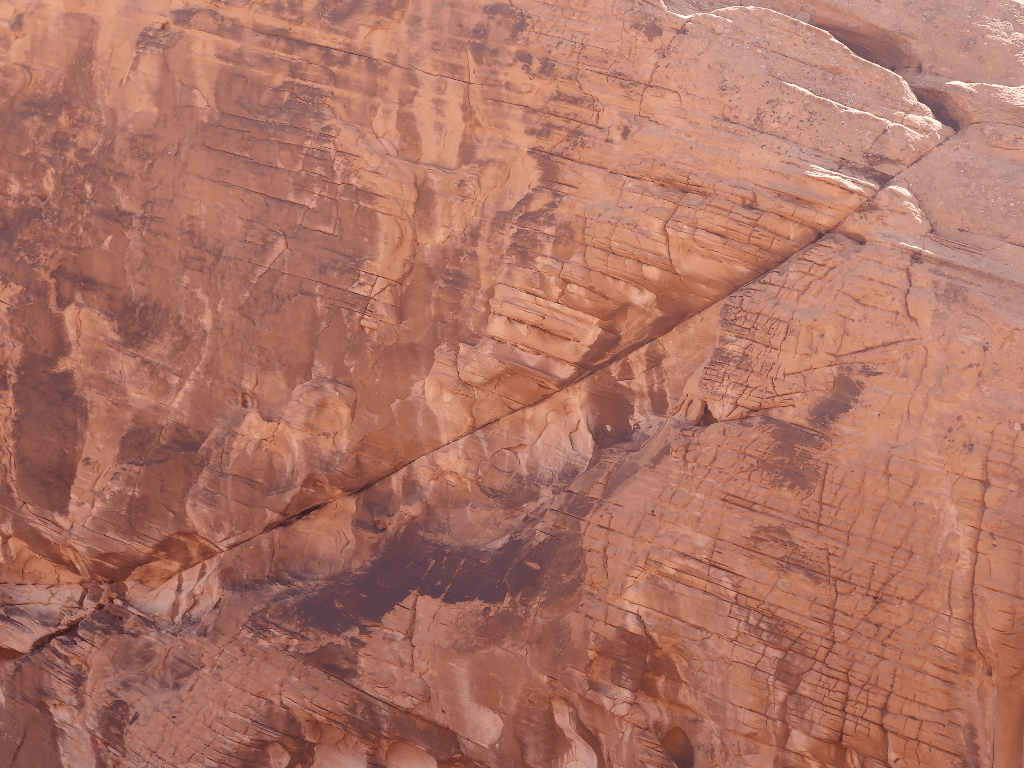}
  \end{subfigure}
  \begin{subfigure}{0.19\linewidth}
    \includegraphics[width=\linewidth]{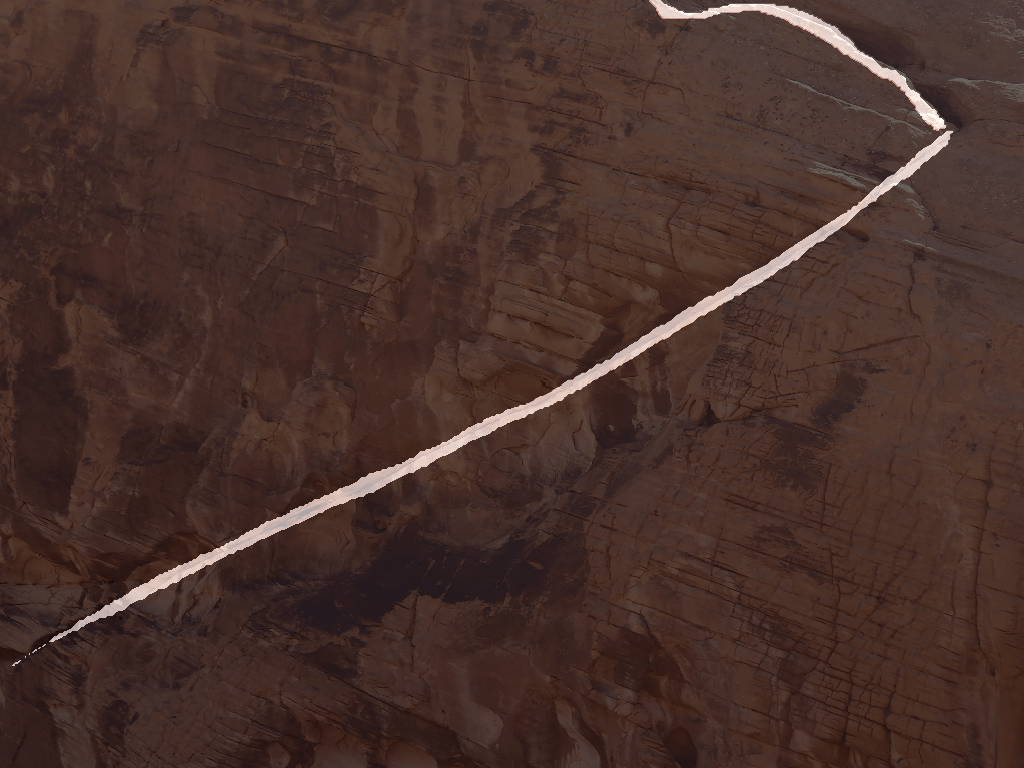}
  \end{subfigure}
  \begin{subfigure}{0.19\linewidth}
    \includegraphics[width=\linewidth]{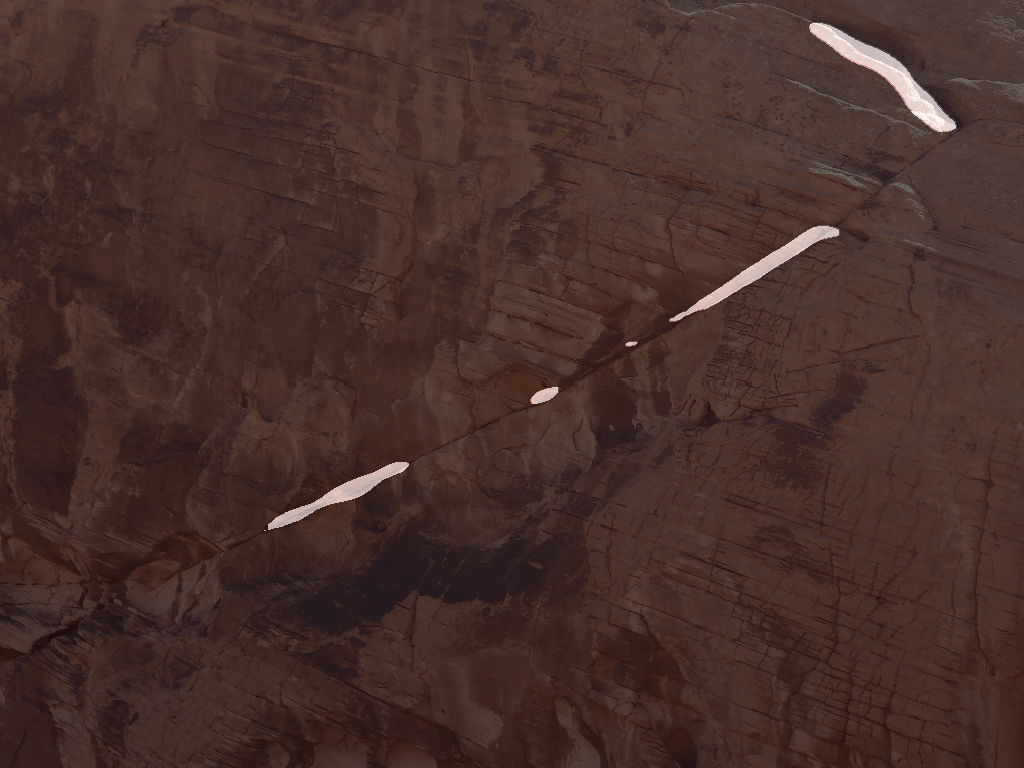}
  \end{subfigure}
  \begin{subfigure}{0.19\linewidth}
    \includegraphics[width=\linewidth]{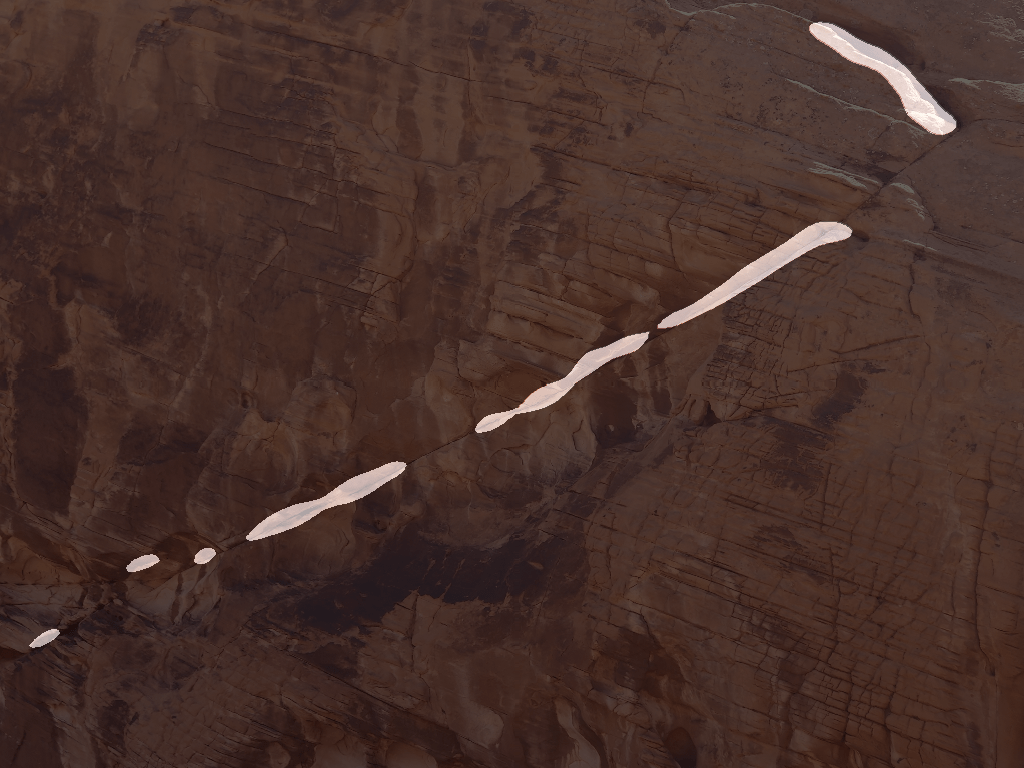}
  \end{subfigure}
  \begin{subfigure}{0.19\linewidth}
    \includegraphics[width=\linewidth]{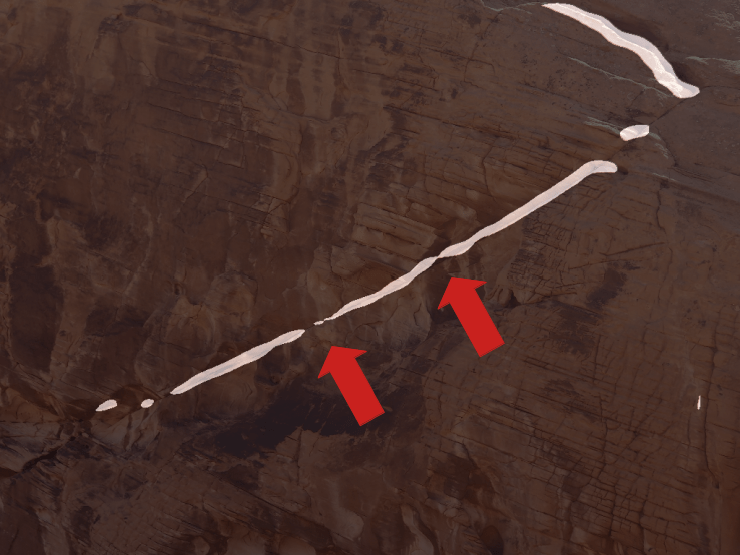}
  \end{subfigure}

\vspace{0.01\linewidth}

    \begin{subfigure}{0.19\linewidth}
    \includegraphics[width=\linewidth]{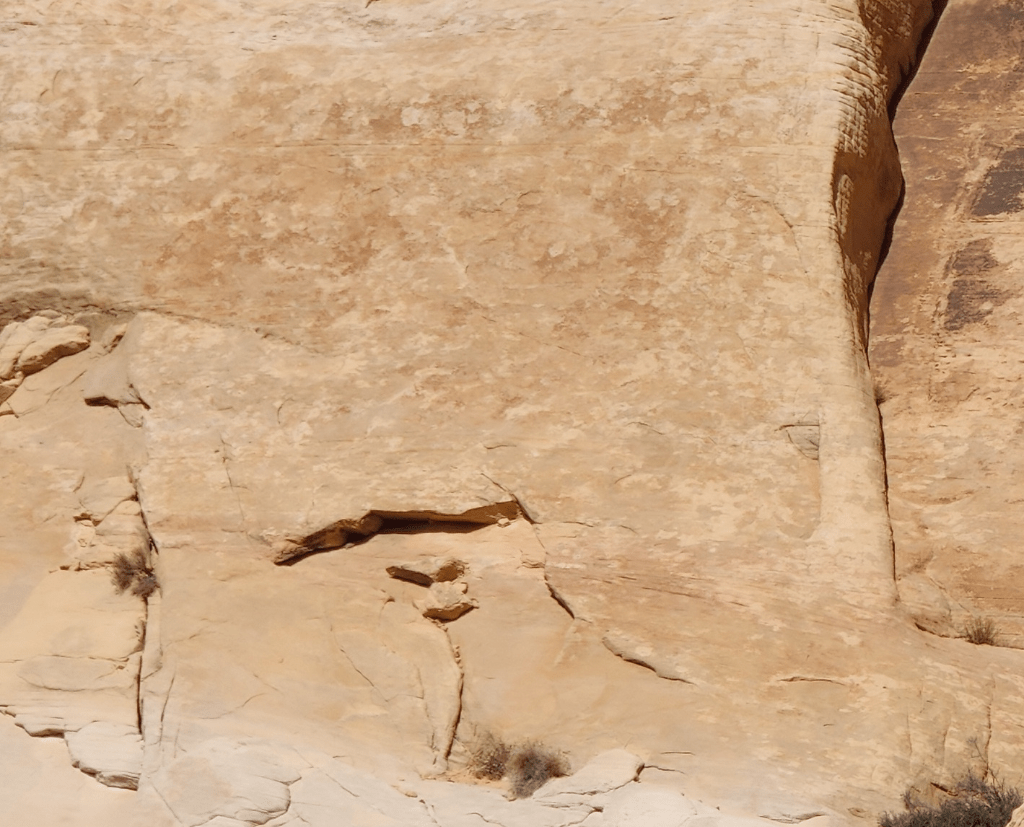}
  \end{subfigure}
  \begin{subfigure}{0.19\linewidth}
    \includegraphics[width=\linewidth]{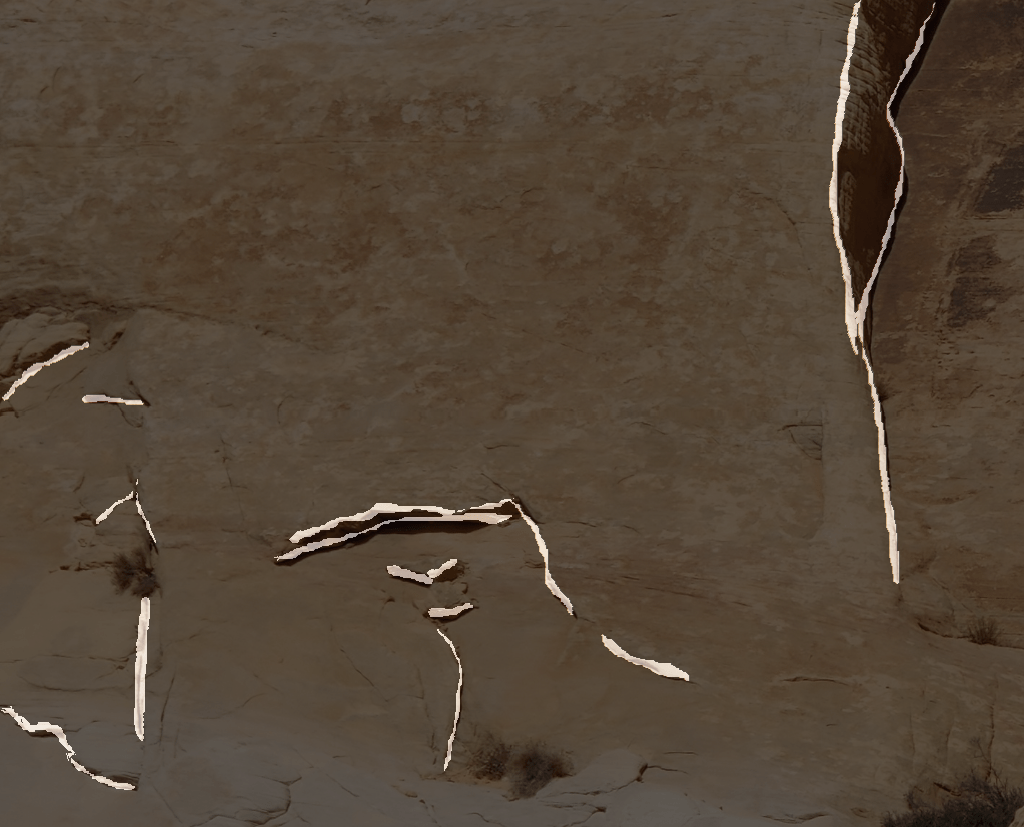}
  \end{subfigure}
  \begin{subfigure}{0.19\linewidth}
    \includegraphics[width=\linewidth]{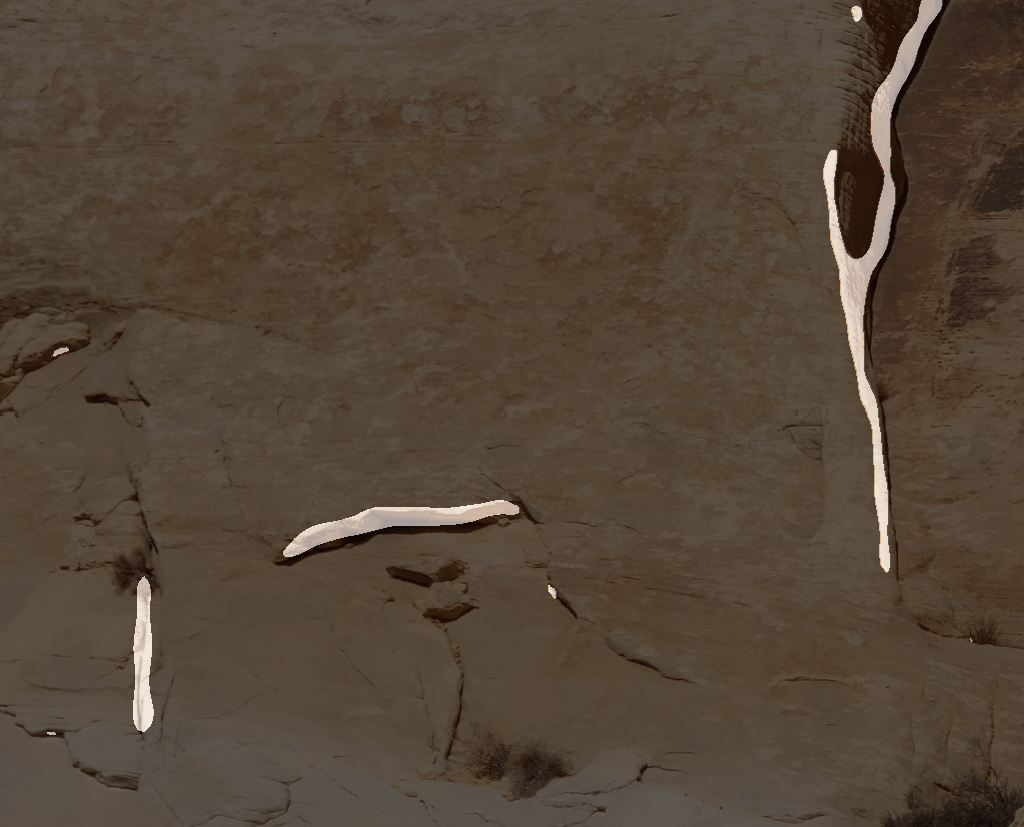}
  \end{subfigure}
  \begin{subfigure}{0.19\linewidth}
    \includegraphics[width=\linewidth]{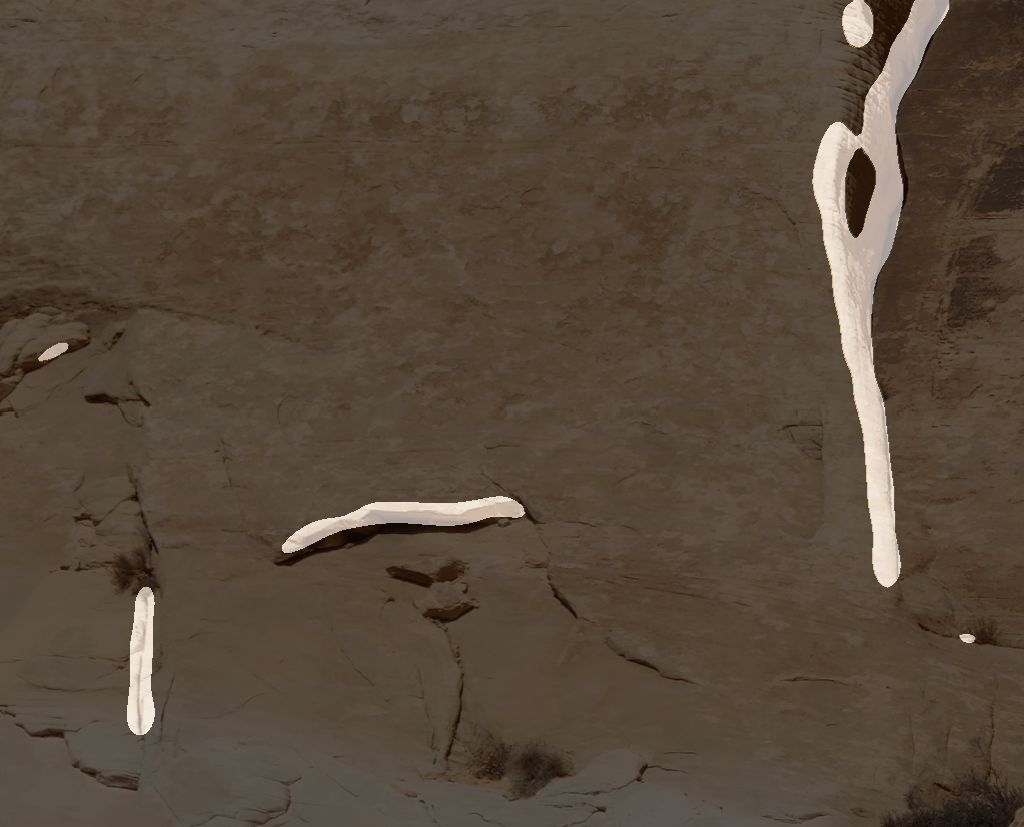}
  \end{subfigure}
  \begin{subfigure}{0.19\linewidth}
    \includegraphics[width=\linewidth]{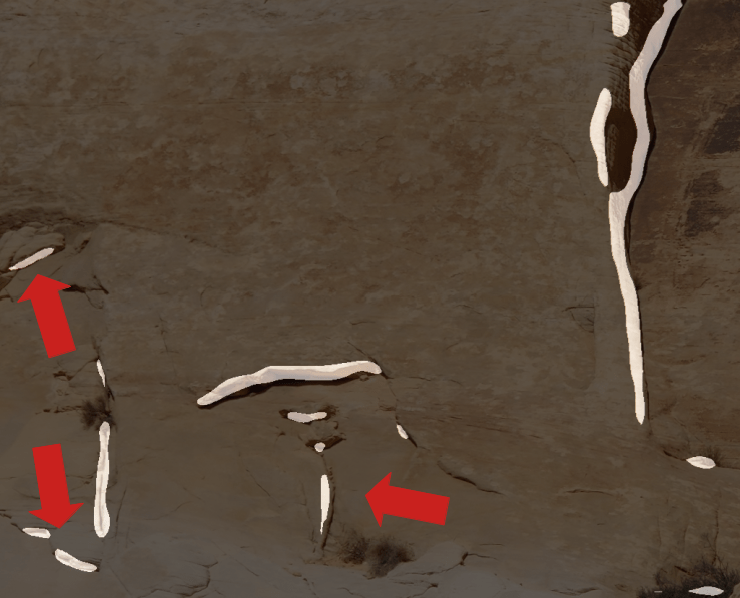}
  \end{subfigure}

\vspace{0.01\linewidth}

    \begin{subfigure}{0.19\linewidth}
    \includegraphics[width=\linewidth]{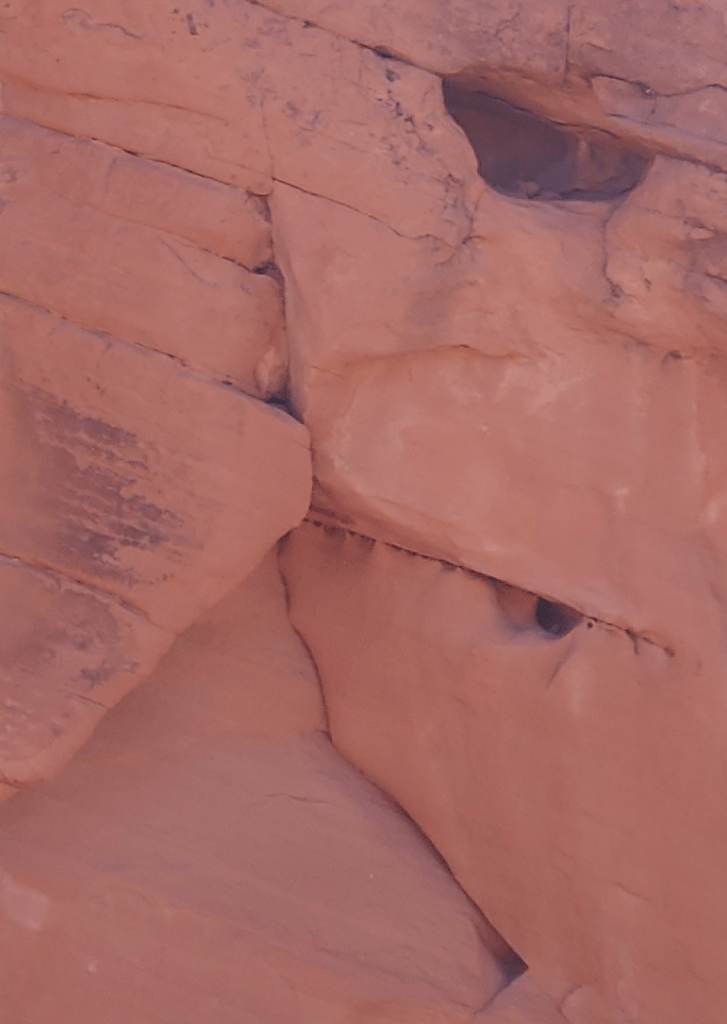}
  \end{subfigure}
  \begin{subfigure}{0.19\linewidth}
    \includegraphics[width=\linewidth]{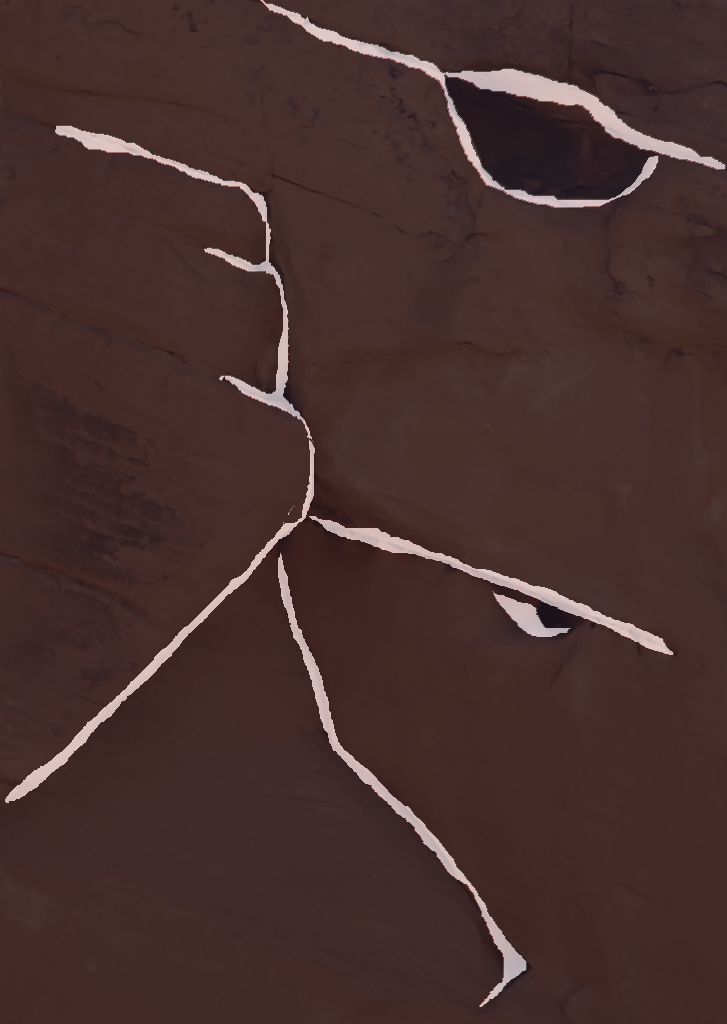}
  \end{subfigure}
  \begin{subfigure}{0.19\linewidth}
    \includegraphics[width=\linewidth]{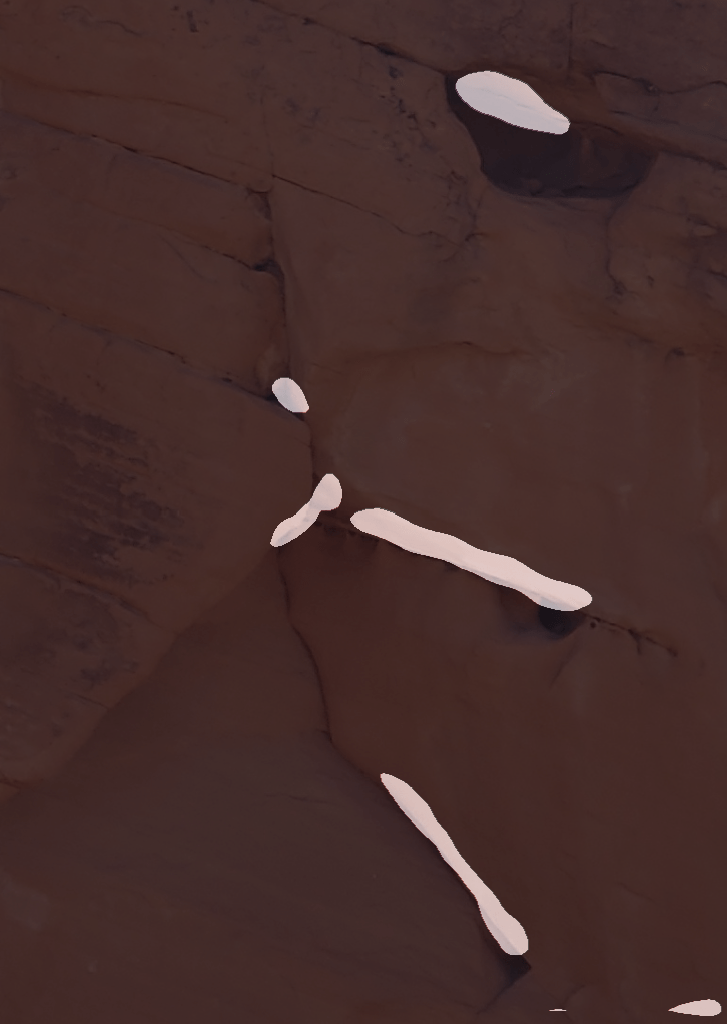}
  \end{subfigure}
  \begin{subfigure}{0.19\linewidth}
    \includegraphics[width=\linewidth]{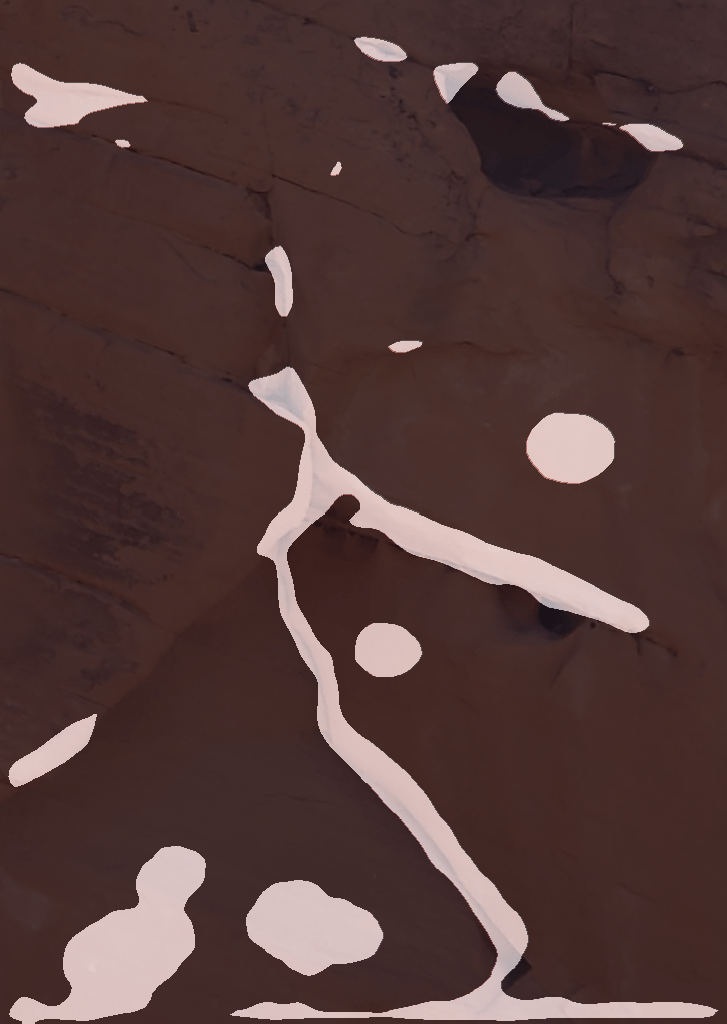}
  \end{subfigure}
  \begin{subfigure}{0.19\linewidth}
    \includegraphics[width=\linewidth]{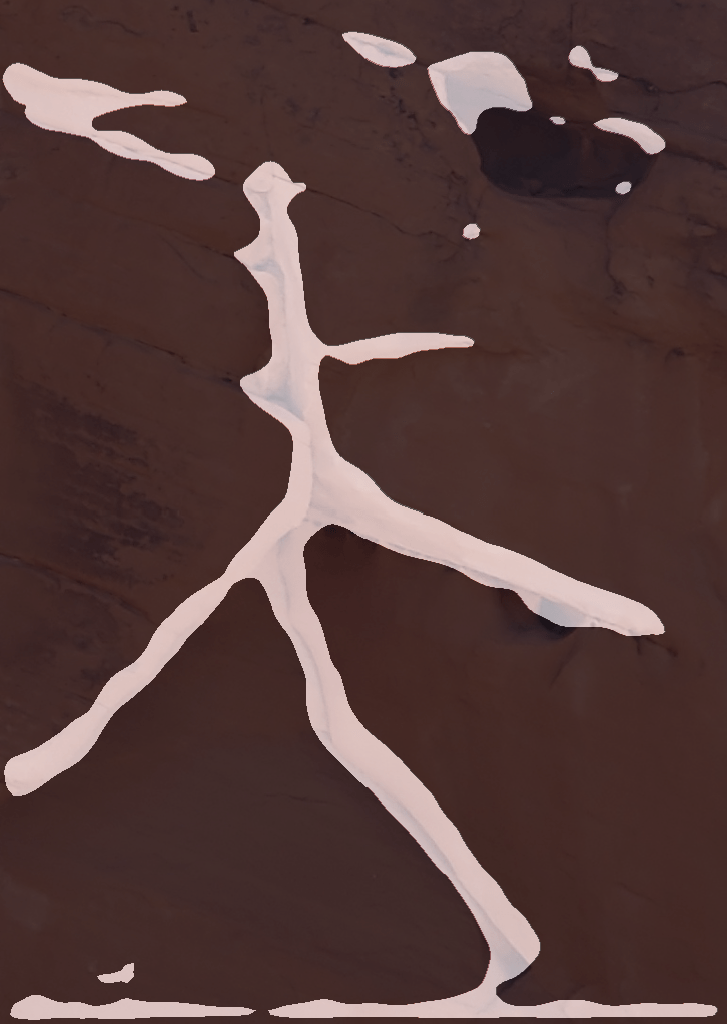}
  \end{subfigure}

  \caption{\textbf{Qualitative results:} From the top, we show results on 4 crack images (after 100 epochs) and one result on the last crack (after 10 epochs). With SKIL, the predicted cracks are more connected and the network is able to learn much faster \textit{(see last row)}. Visual improvements are marked by red arrows.}
  \label{fig:results-cracks}
\end{figure}

Figure \ref{fig:results-cracks} shows the qualitative results obtained on different crack images. From the top, images (1) -- (3) show that SKIL detects more details, with fewer gaps in predictions, and image (4) shows a more accurate predicted width. Image (5) is trained after 10 epochs and illustrates the sample efficiency with SKIL.

\subsubsection{Qualitative results on blood vessels}
\label{sec:result-loss-vessels}

\begin{figure}[H]
  \captionsetup[subfigure]{justification=centering, labelformat=empty} 
  \centering

  \begin{subfigure}{0.1902\linewidth}
      \caption{Image}
      \includegraphics[width=\linewidth]{images/results/vessels/img_0-min.png}  
  \end{subfigure}
  \begin{subfigure}{0.1902\linewidth}
    \caption{Label}
    \includegraphics[width=\linewidth]{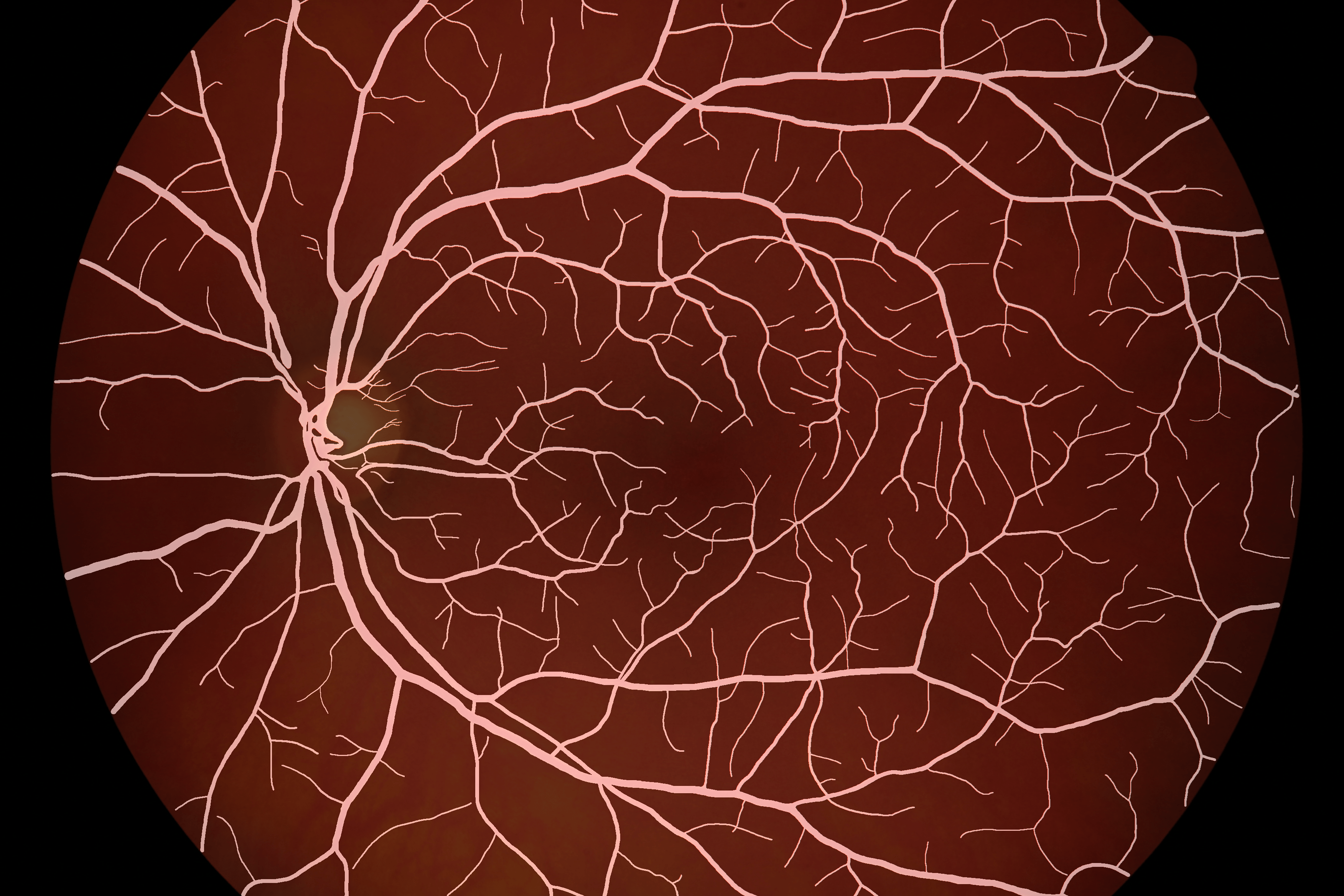}
  \end{subfigure}
  \begin{subfigure}{0.1902\linewidth}
    \caption{Dice}
    \includegraphics[width=\linewidth]{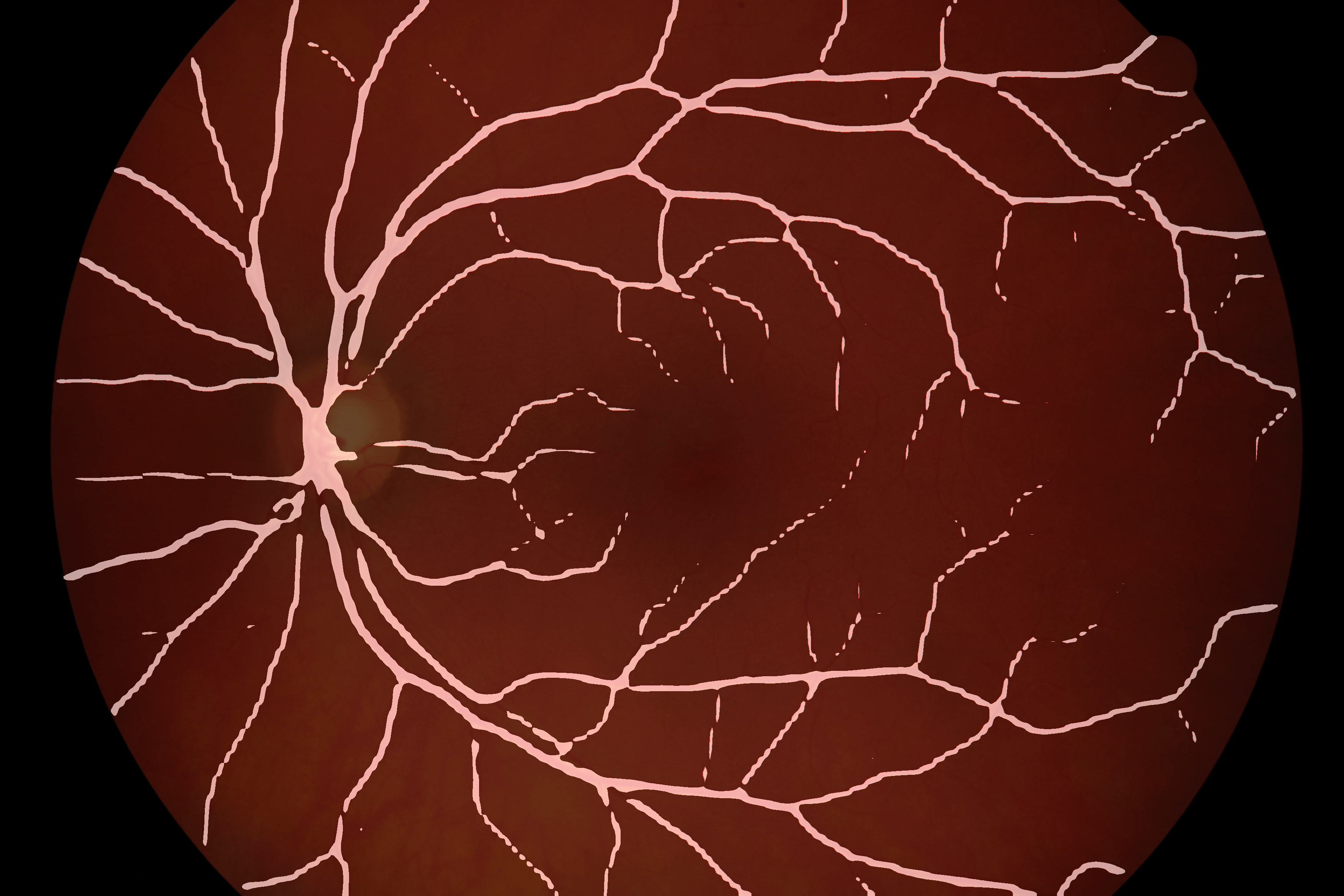}
  \end{subfigure}
  \begin{subfigure}{0.1902\linewidth}
    \caption{CL Dice}
    \includegraphics[width=\linewidth]{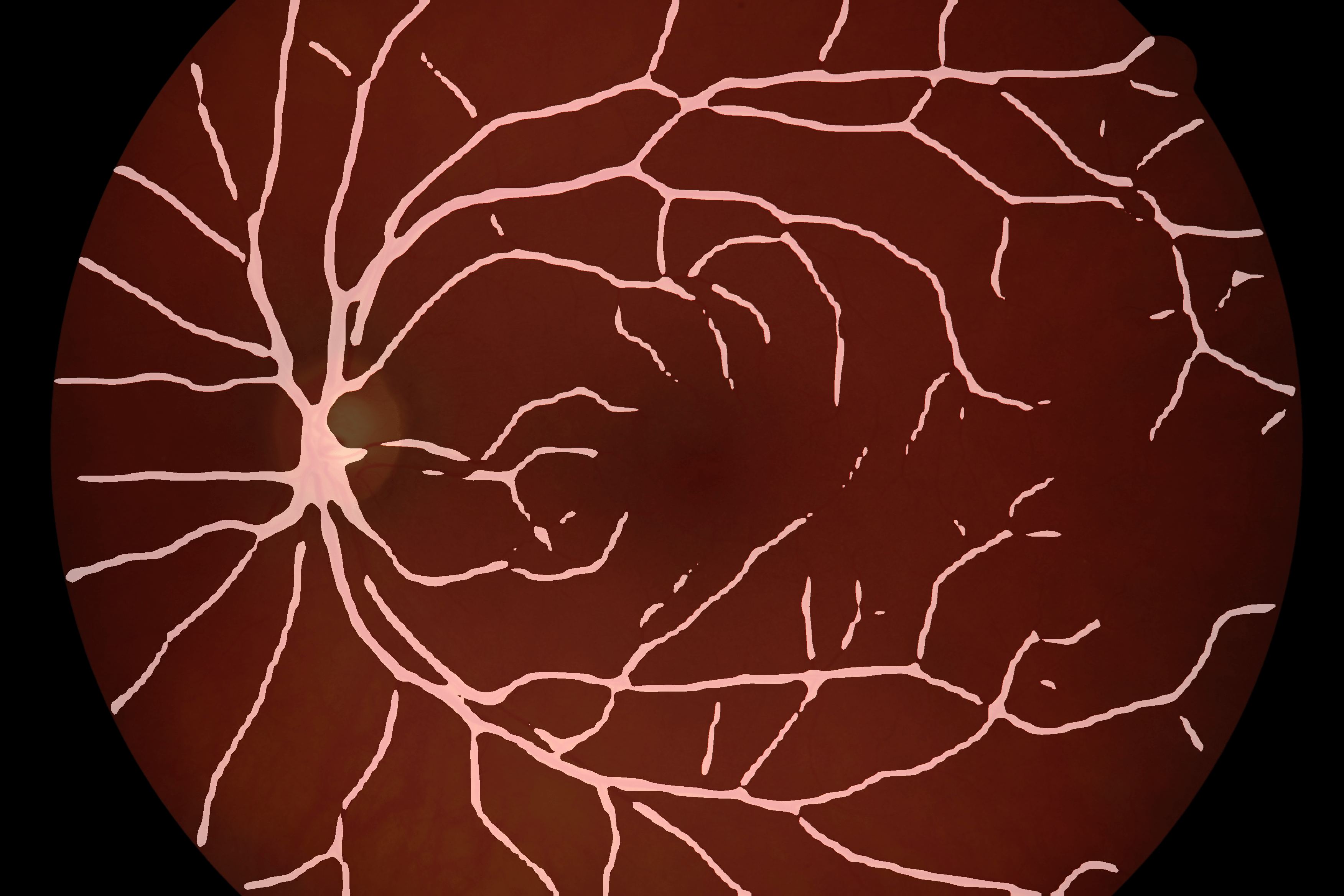}
  \end{subfigure}
  \begin{subfigure}{0.1902\linewidth}
    \caption{SKIL}
    \includegraphics[width=\linewidth]{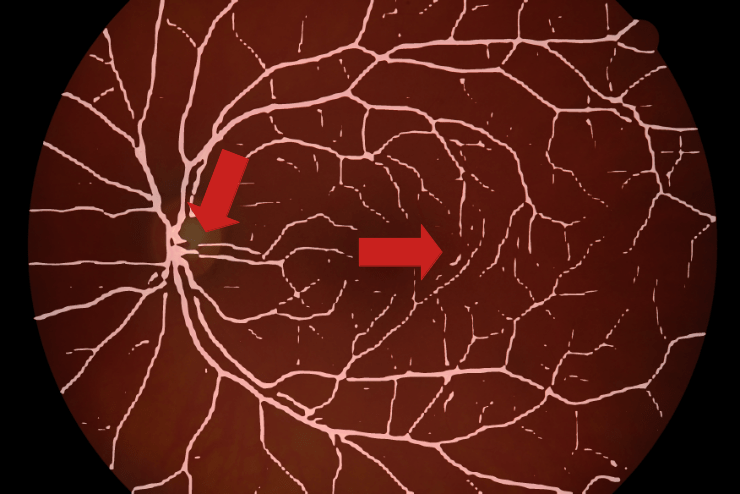}
  \end{subfigure}

\vspace{0.01\linewidth}

    \begin{subfigure}{0.1902\linewidth}
    \includegraphics[width=\linewidth]{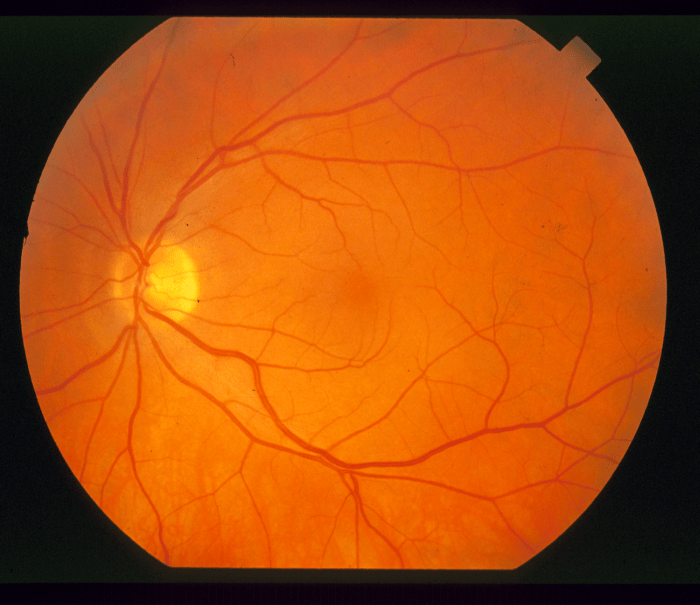}
  \end{subfigure}
  \begin{subfigure}{0.1902\linewidth}
    \includegraphics[width=\linewidth]{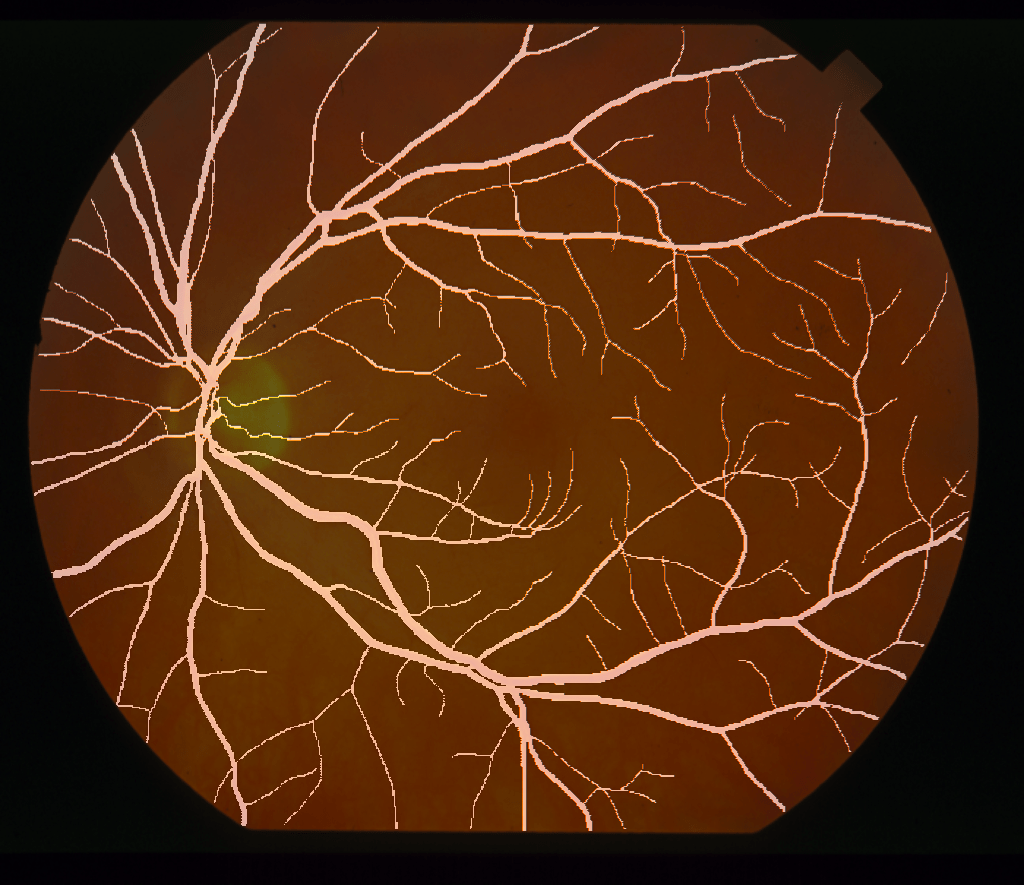}
  \end{subfigure}
  \begin{subfigure}{0.1902\linewidth}
    \includegraphics[width=\linewidth]{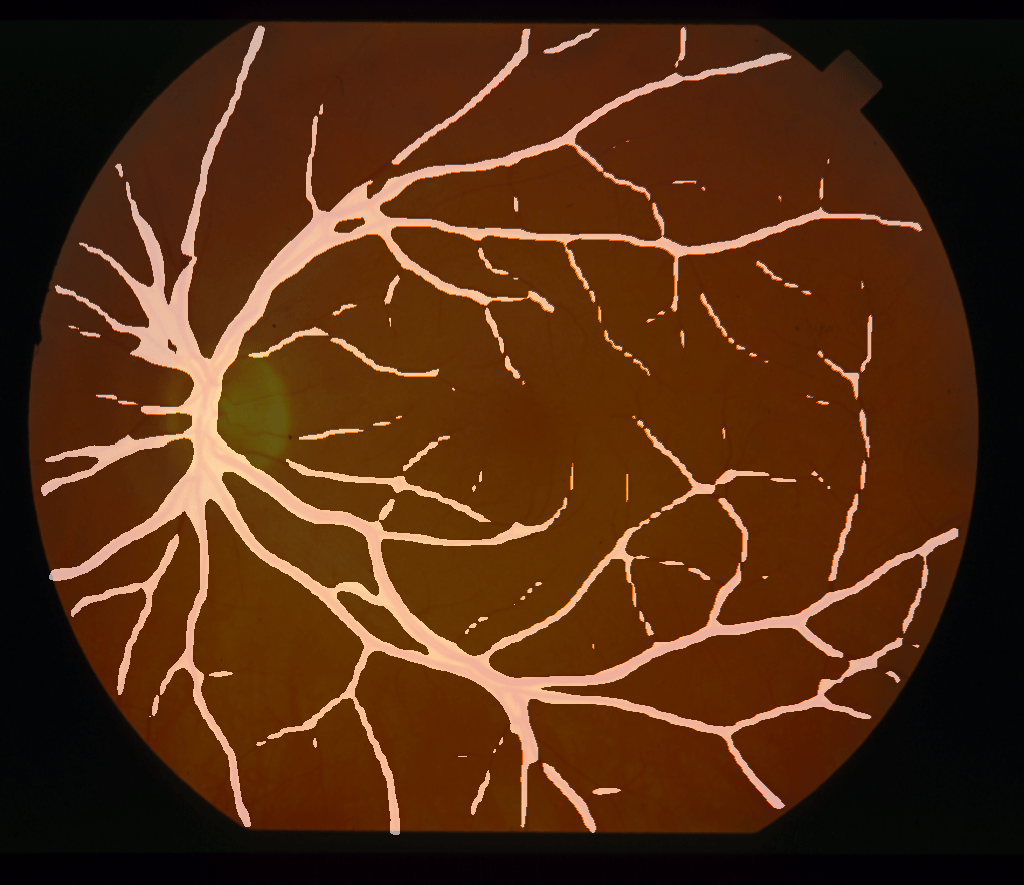}
  \end{subfigure}
  \begin{subfigure}{0.1902\linewidth}
    \includegraphics[width=\linewidth]{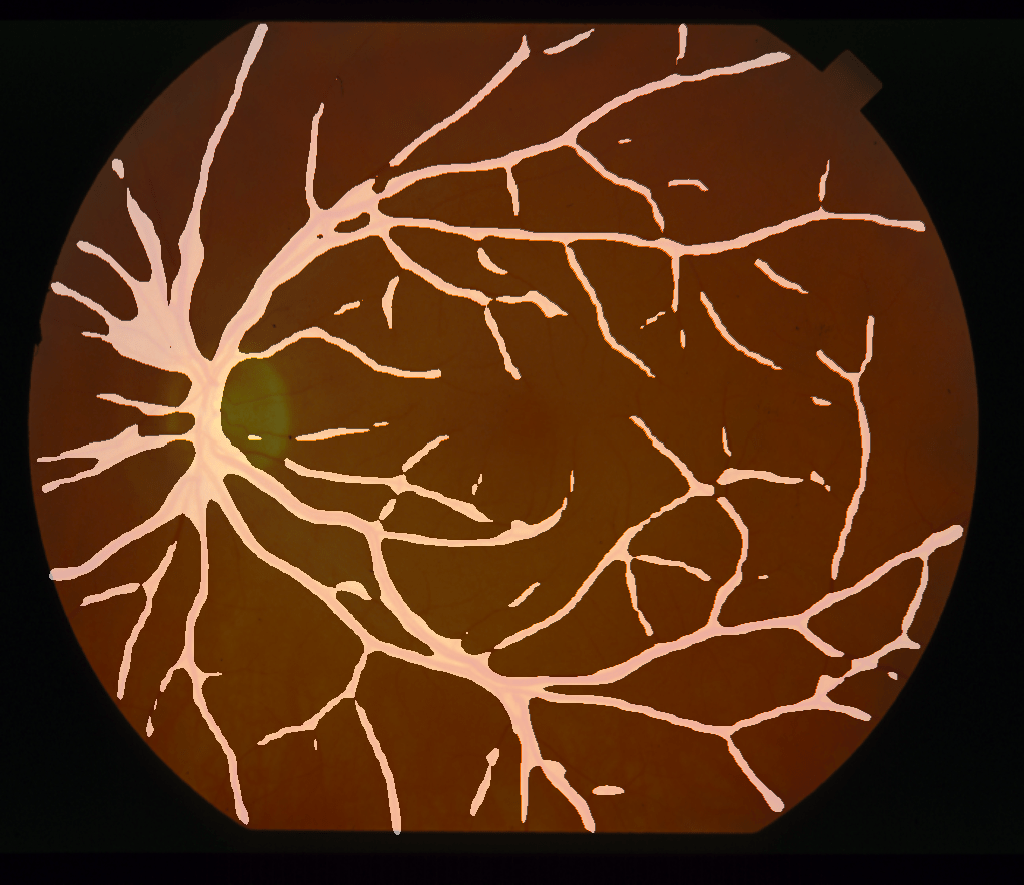}
  \end{subfigure}
  \begin{subfigure}{0.1902\linewidth}
    \includegraphics[width=\linewidth]{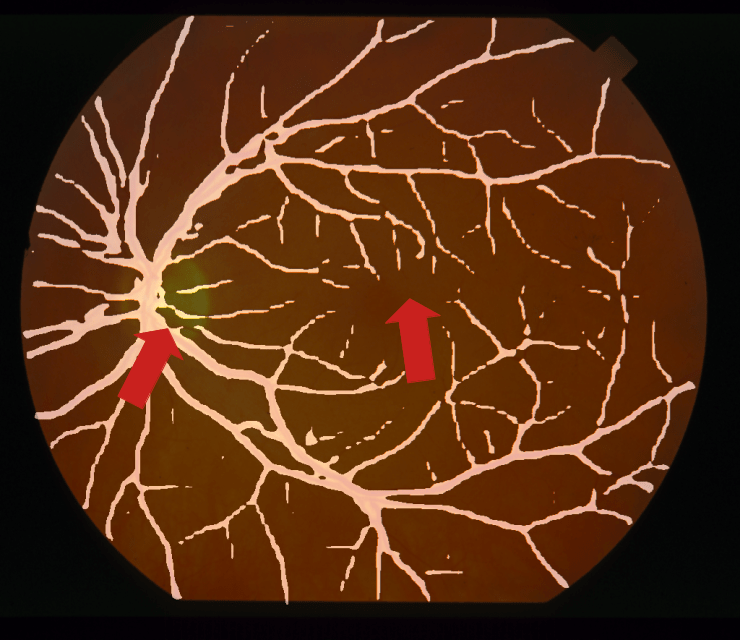}
  \end{subfigure}

\vspace{0.01\linewidth}

    \begin{subfigure}{0.1902\linewidth}
    \includegraphics[width=\linewidth]{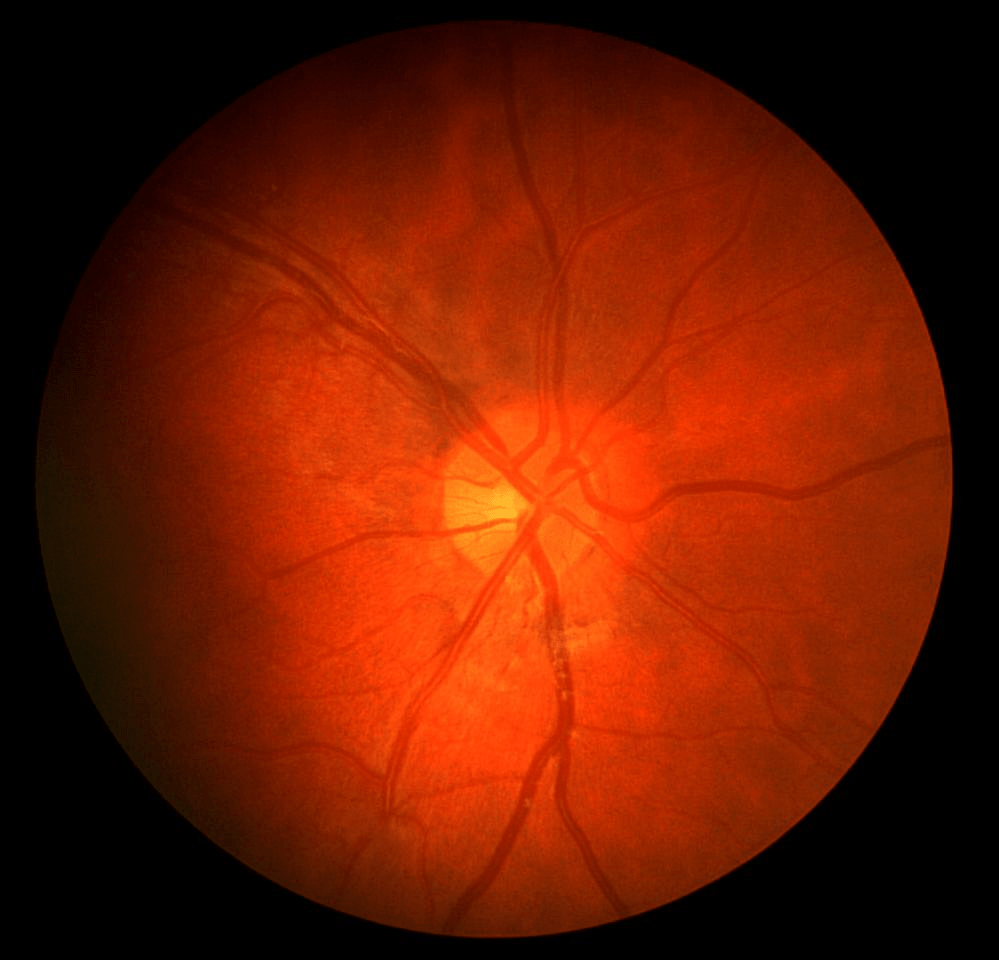}
  \end{subfigure}
  \begin{subfigure}{0.1902\linewidth}
    \includegraphics[width=\linewidth]{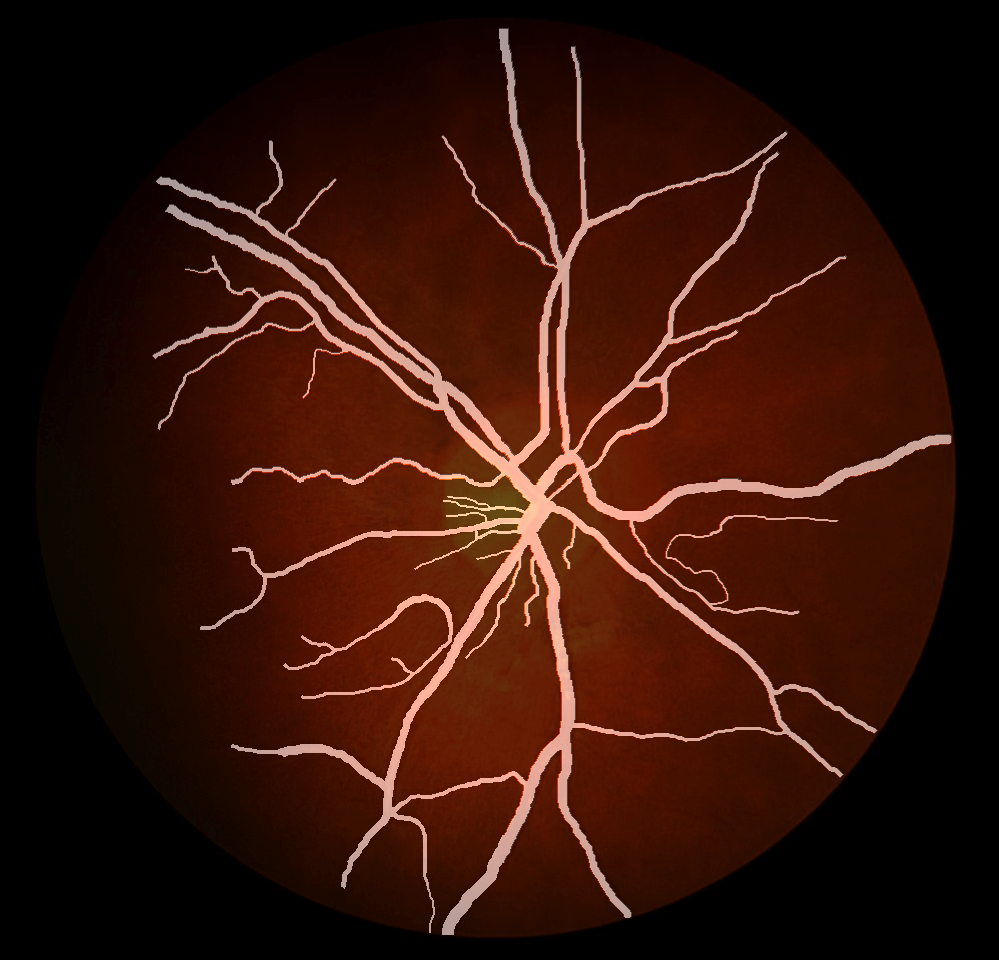}
  \end{subfigure}
  \begin{subfigure}{0.1902\linewidth}
    \includegraphics[width=\linewidth]{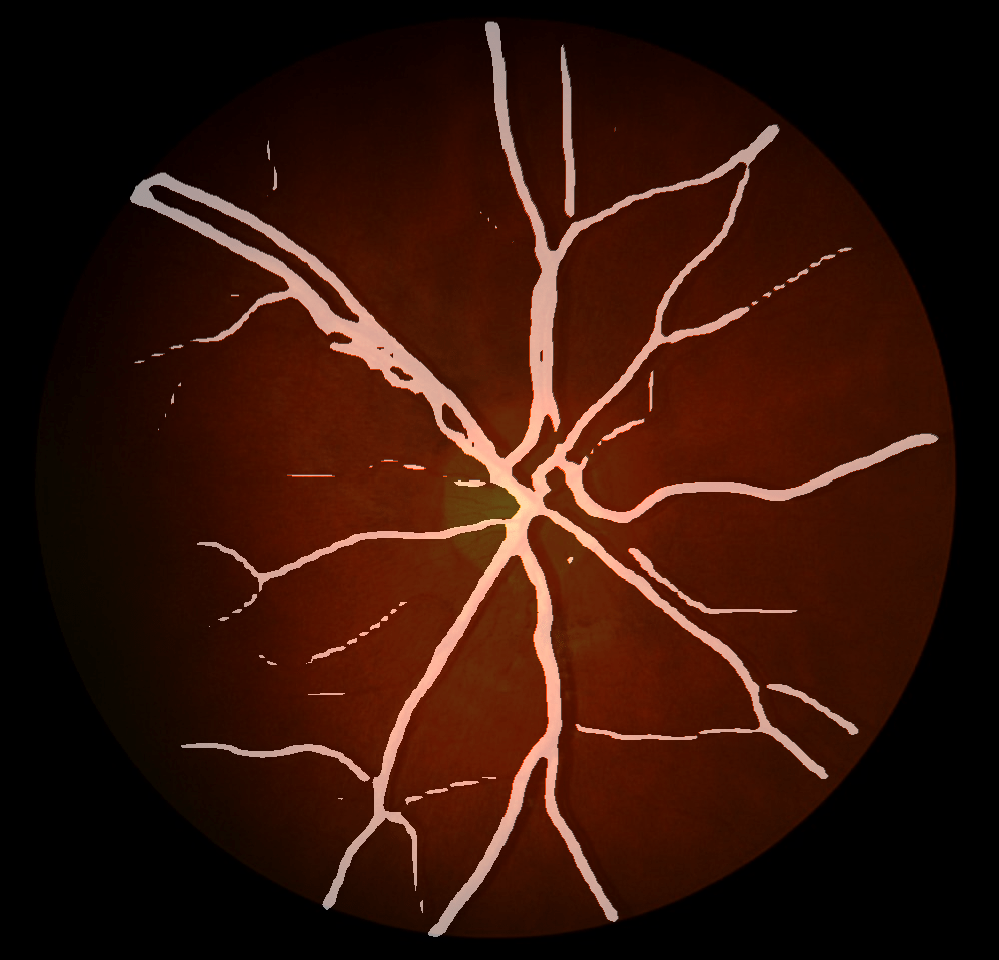}
  \end{subfigure}
  \begin{subfigure}{0.1902\linewidth}
    \includegraphics[width=\linewidth]{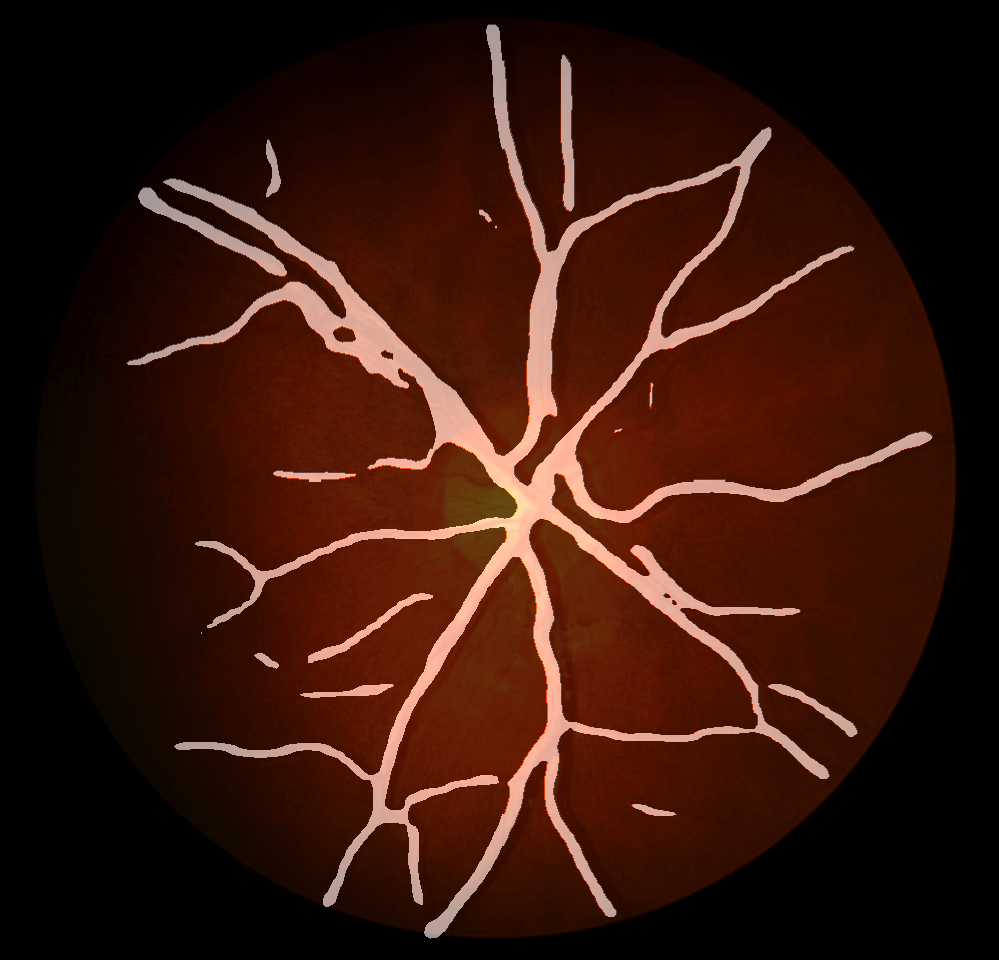}
  \end{subfigure}
  \begin{subfigure}{0.1902\linewidth}
    \includegraphics[width=\linewidth]{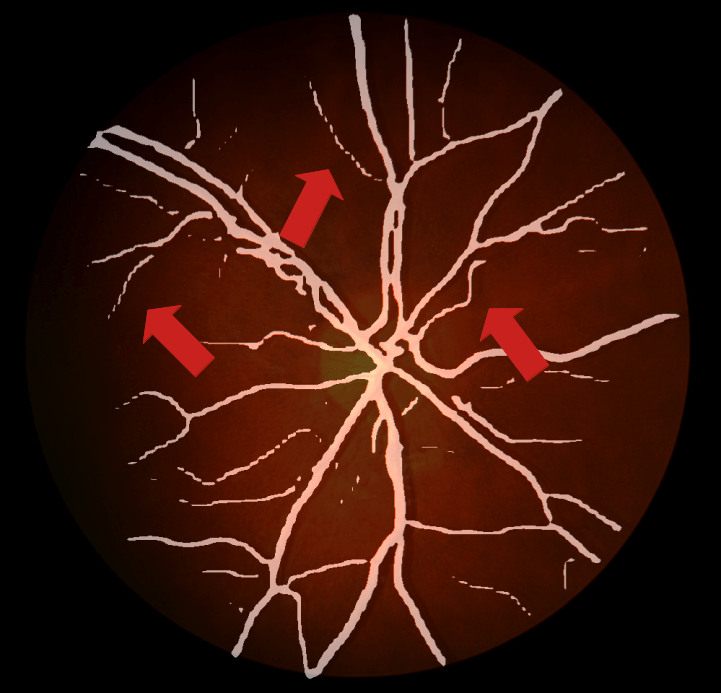}
  \end{subfigure}

\vspace{0.01\linewidth}

    \begin{subfigure}{0.1902\linewidth}
    \includegraphics[width=\linewidth]{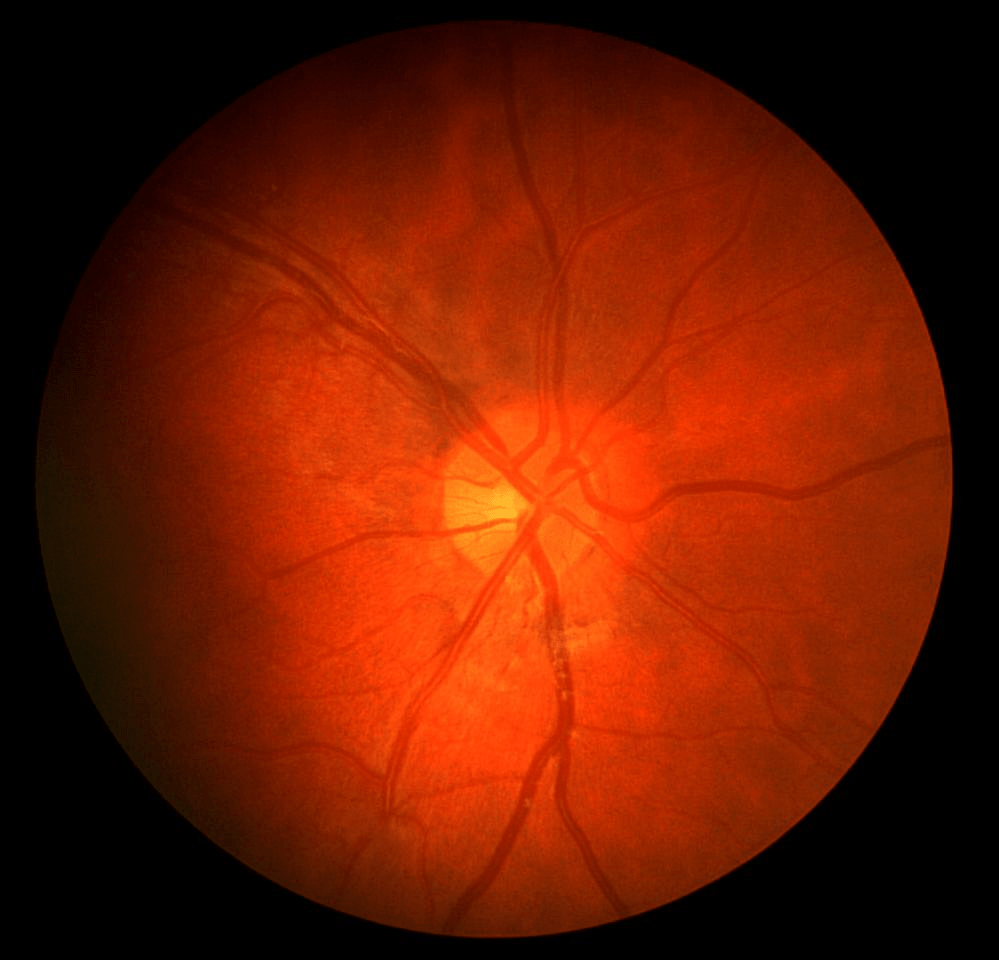}
  \end{subfigure}
  \begin{subfigure}{0.1902\linewidth}
    \includegraphics[width=\linewidth]{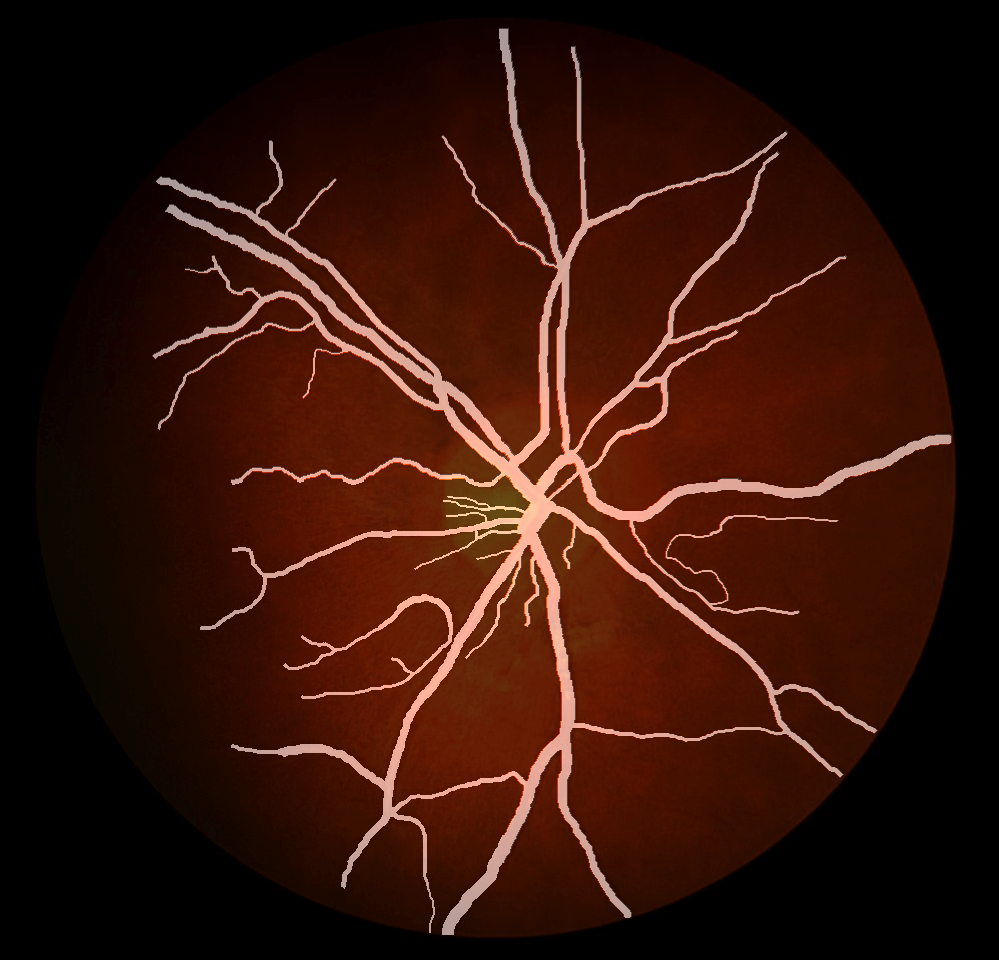}
  \end{subfigure}
  \begin{subfigure}{0.1902\linewidth}
    \includegraphics[width=\linewidth]{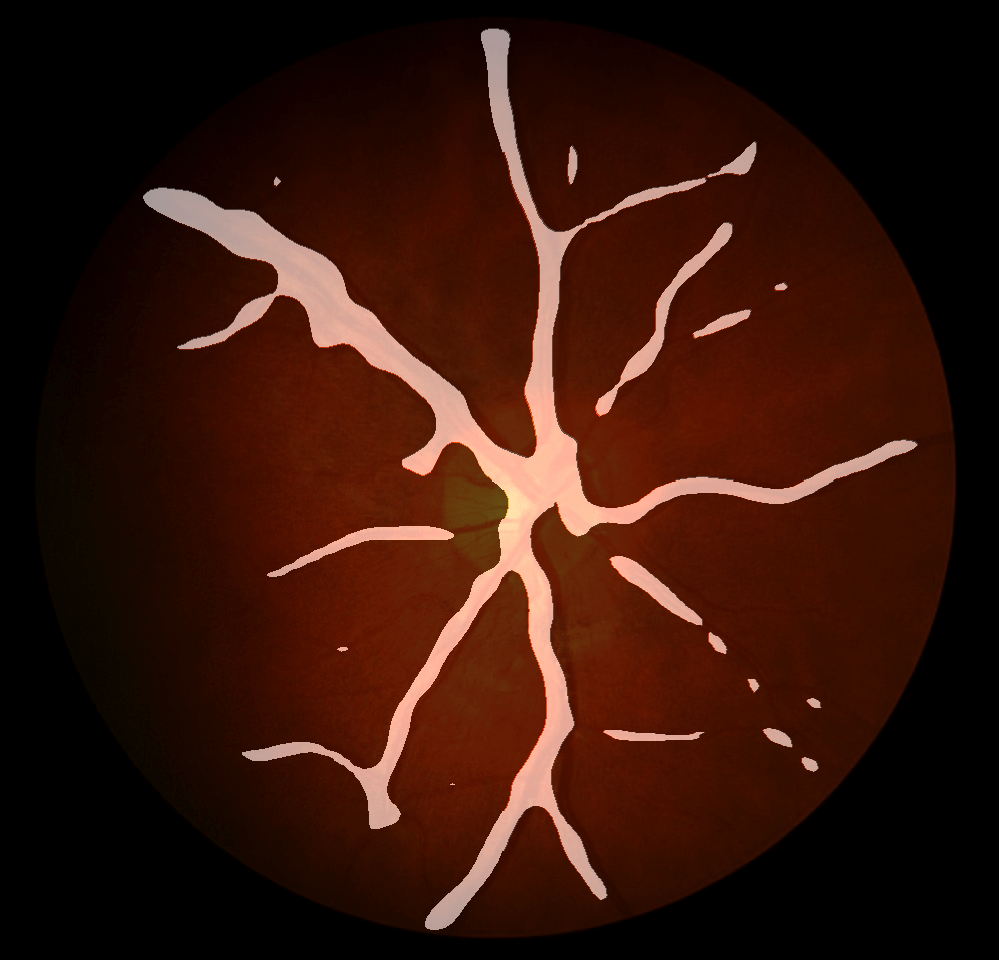}
  \end{subfigure}
  \begin{subfigure}{0.1902\linewidth}
    \includegraphics[width=\linewidth]{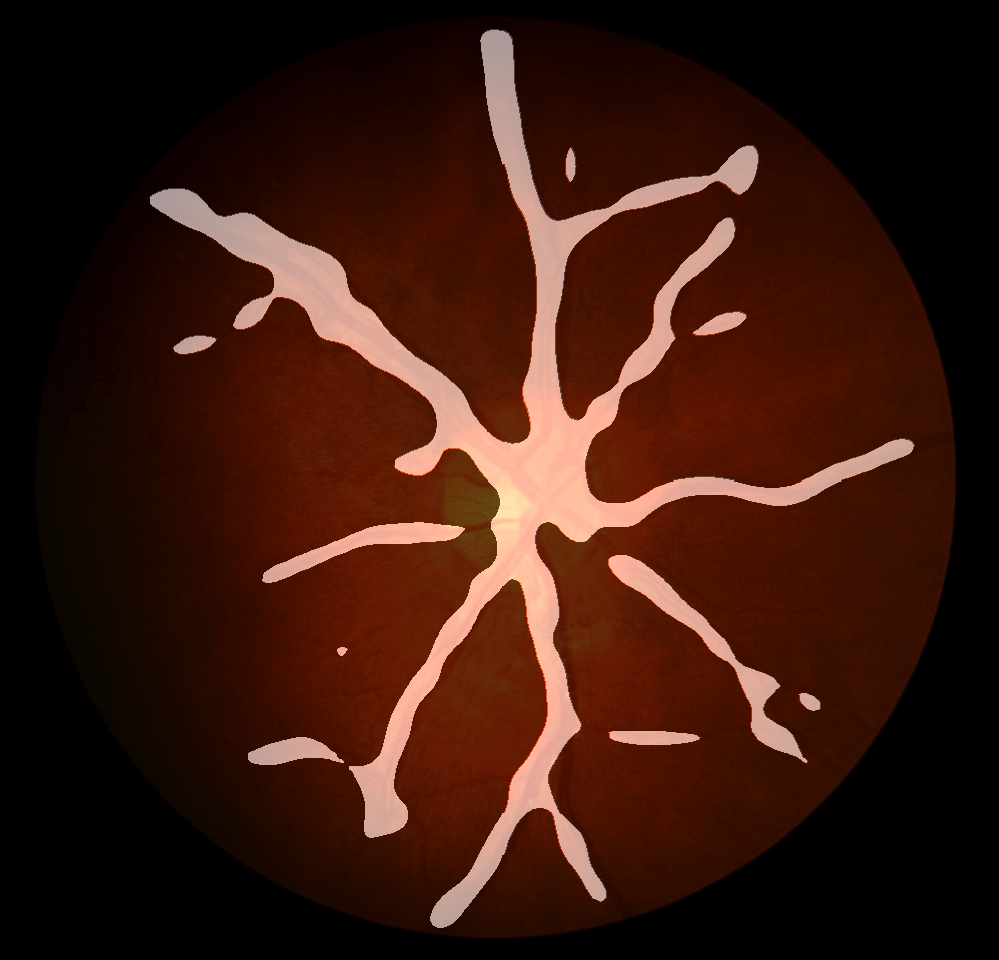}
  \end{subfigure}
  \begin{subfigure}{0.1902\linewidth}
    \includegraphics[width=\linewidth]{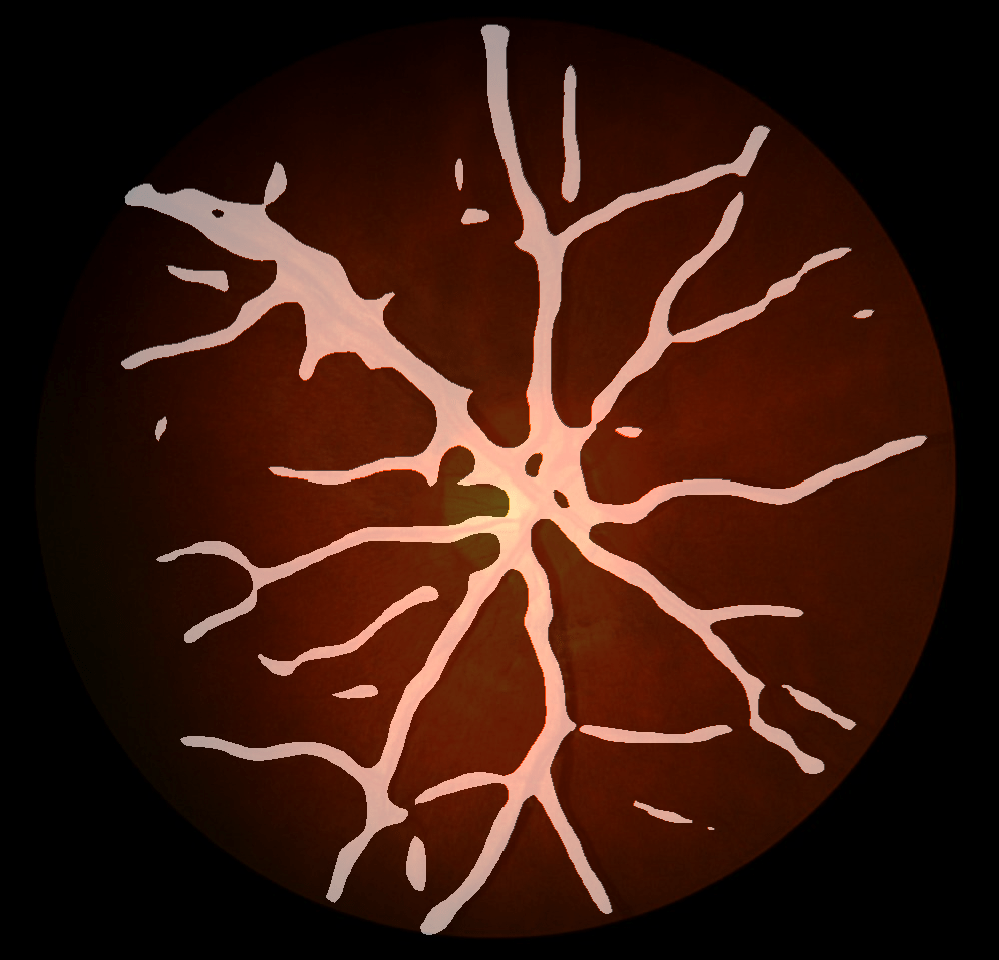}
  \end{subfigure}
  
    \caption{\textbf{Qualitative results:} From the top, we show results from HRF, STARE, CHASE-DB1 (after 200 epochs), and finally CHASE-DB1 (after 20 epochs). The predictions show that with SKIL (SKILL-Dice), the vessels are thinner and more detailed (Dice and CL-Dice tend to miss small vessels). Visual improvements are marked by red arrows.}
  \label{fig:results-vessels}
\end{figure}

Figure \ref{fig:results-vessels} illustrates how much better the model performs with SKIL (SKIL-Dice). The first two rows show that when trained with SKIL, the network is able to detect thin vessels that are not detected when trained with just Dice or CL-Dice. In addition, the third row also shows that even for large vessels, SKIL does a better job at predicting the correct width. For example, in the third row from the top, the middle diagonal vessel is actually two parallel vessels which are mostly fused when trained with Dice or CL-Dice, but cleanly separated when trained with SKIL. Finally, the last row illustrates the fact that SKIL enables the network to converge faster. After only 10\% of its training, the network identifies more vessels with SKIL.

\subsubsection{SKIL with U-Net}
\label{sec:unet}
\begin{figure}[h]
  \captionsetup[subfigure]{justification=centering, labelformat=empty} 
  \centering

  \begin{subfigure}{0.184\linewidth}
      \includegraphics[width=\linewidth]{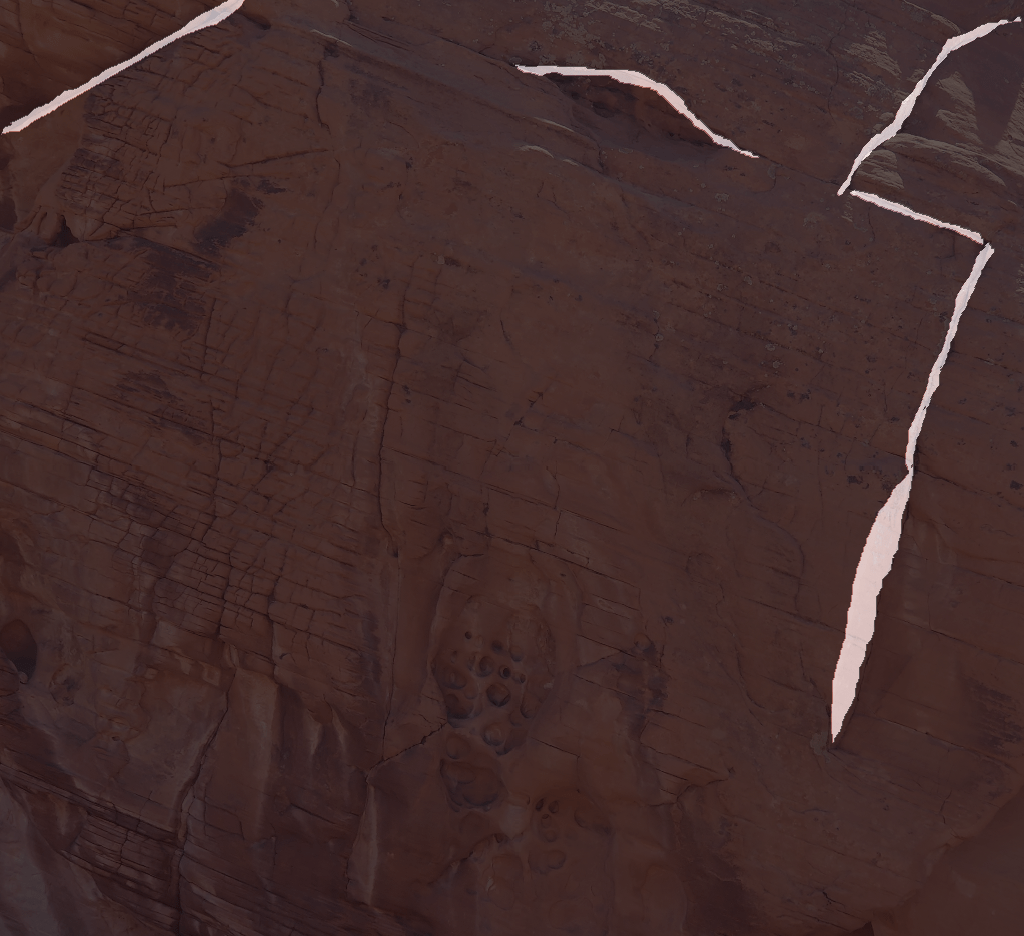}  
  \end{subfigure}
  \begin{subfigure}{0.300\linewidth}
    \includegraphics[width=\linewidth]{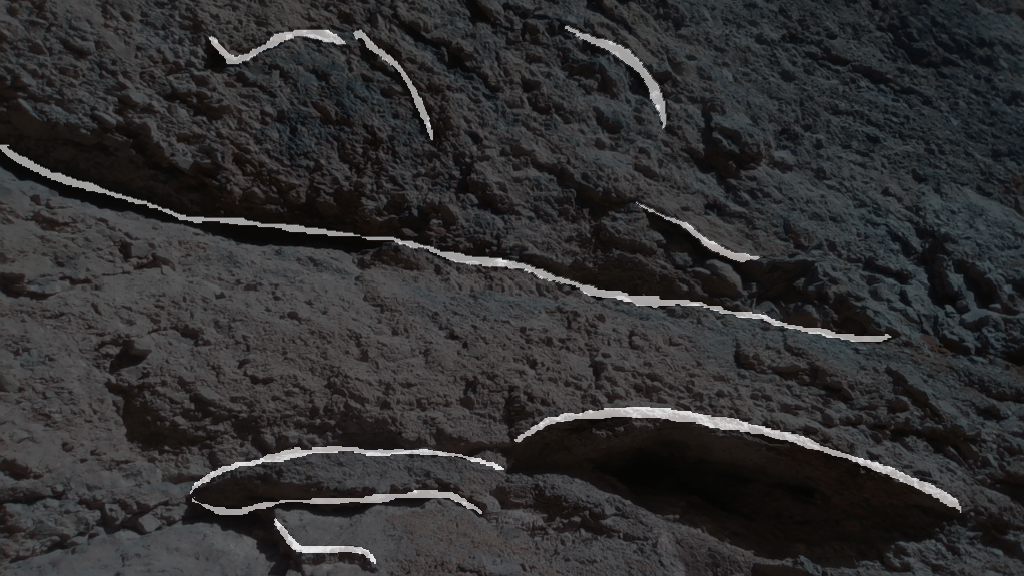}
  \end{subfigure}
  \begin{subfigure}{0.275\linewidth}
    \includegraphics[width=\linewidth]{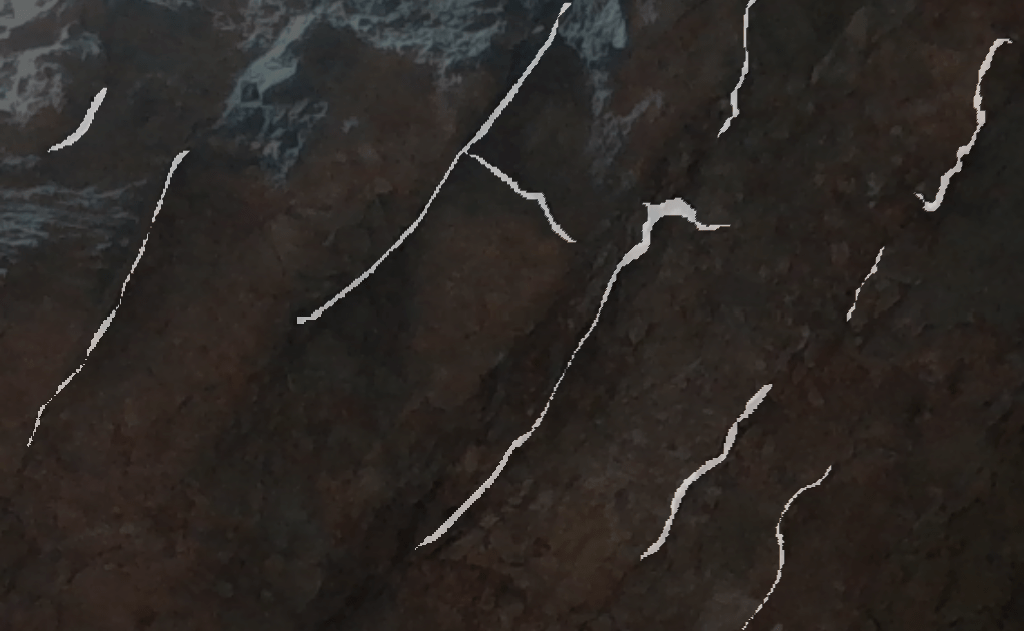}
  \end{subfigure}
  \begin{subfigure}{0.203\linewidth}
    \includegraphics[width=\linewidth]{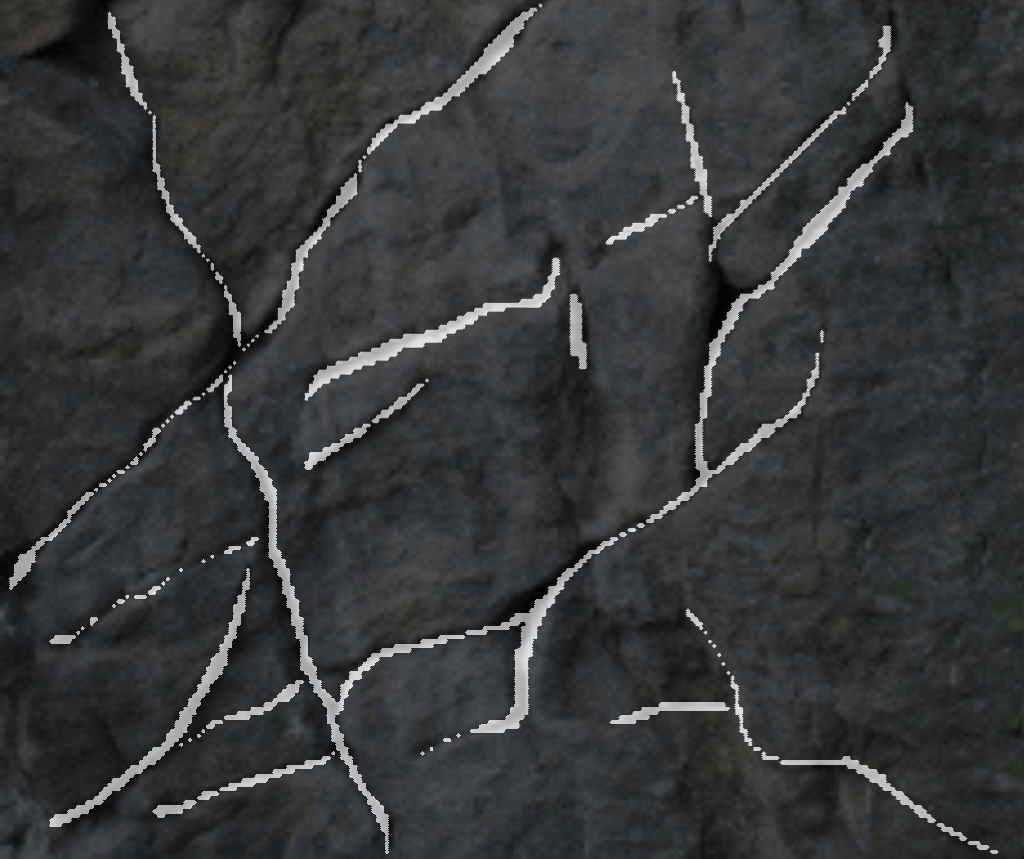}
  \end{subfigure}

\vspace{0.01\linewidth}

    \begin{subfigure}{0.184\linewidth}
    \includegraphics[width=\linewidth]{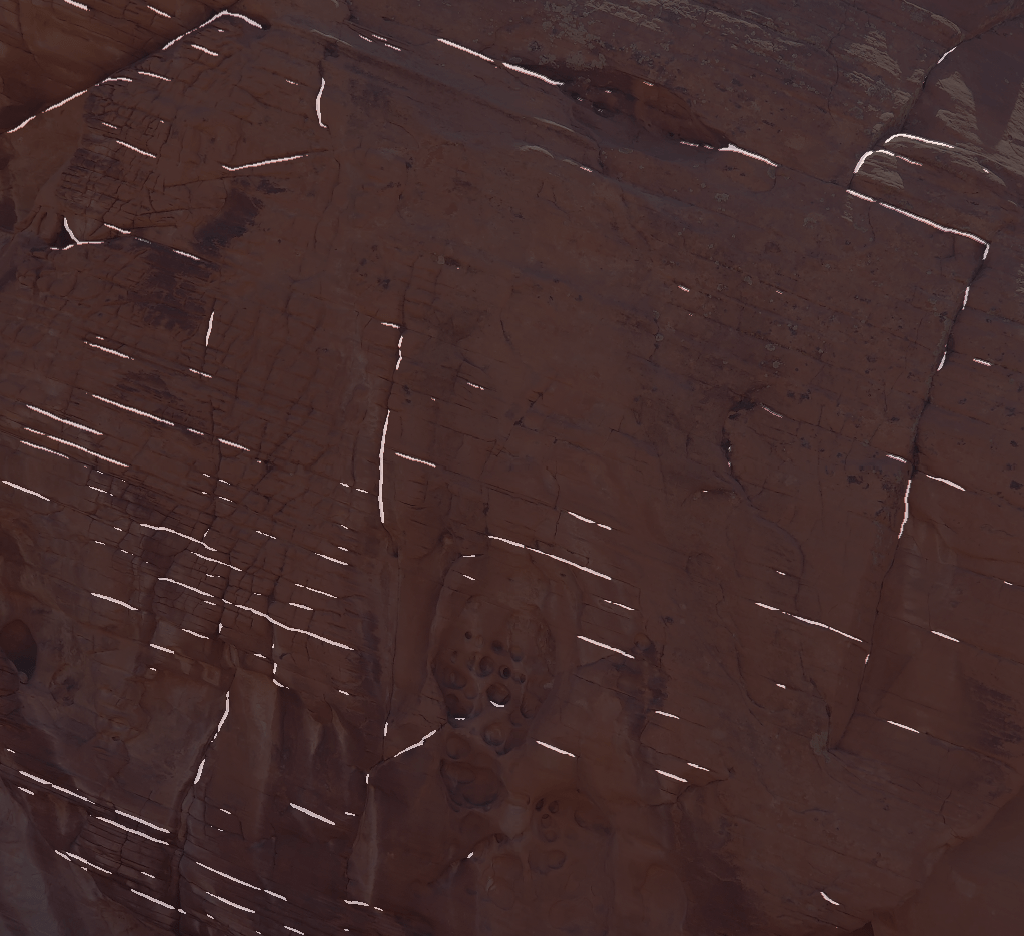}
  \end{subfigure}
  \begin{subfigure}{0.300\linewidth}
    \includegraphics[width=\linewidth]{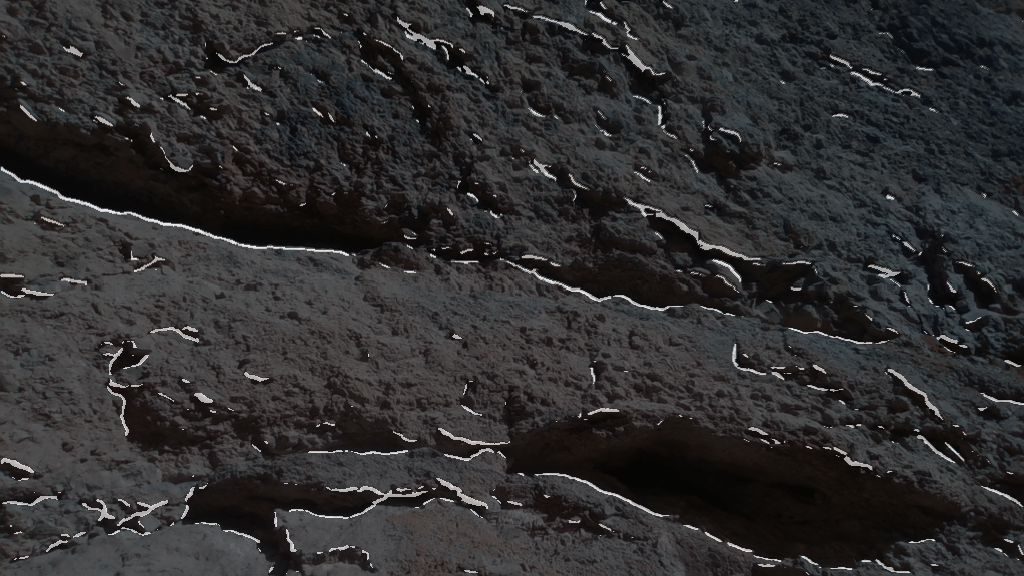}
  \end{subfigure}
  \begin{subfigure}{0.275\linewidth}
    \includegraphics[width=\linewidth]{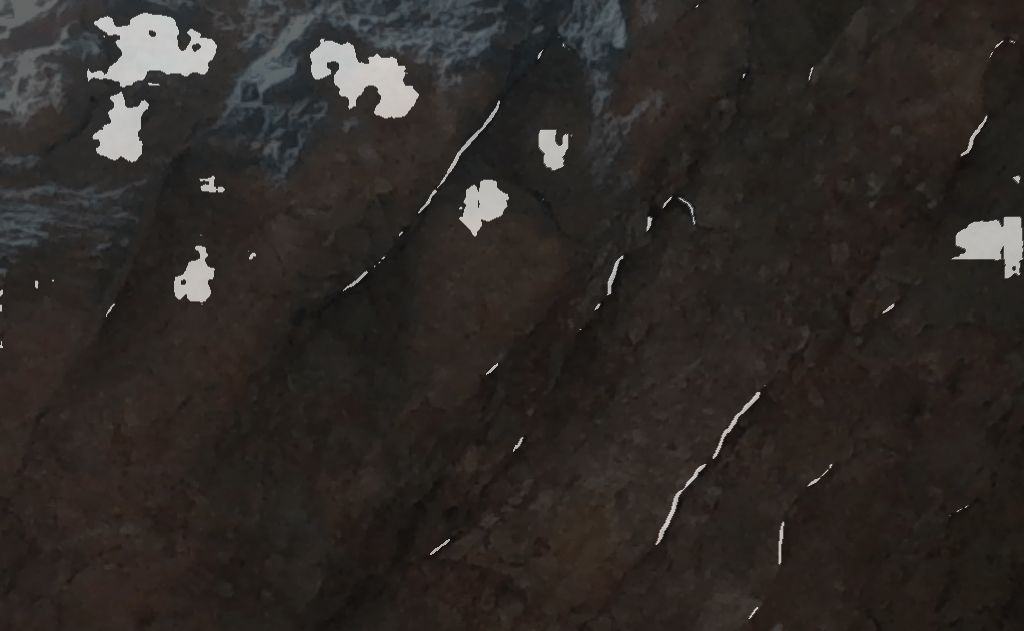}
  \end{subfigure}
  \begin{subfigure}{0.203\linewidth}
    \includegraphics[width=\linewidth]{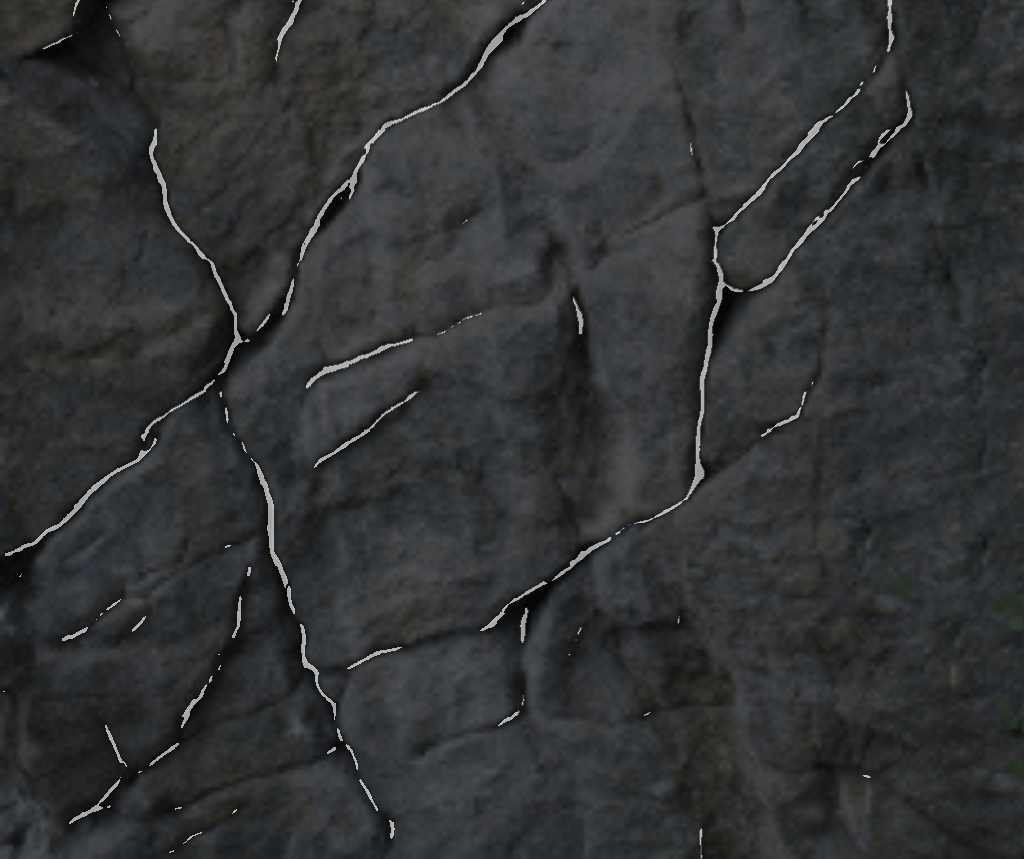}
  \end{subfigure}
  
  \caption{Predictions \textit{(bottom row)} from U-Net versus ground truth \textit{(top row)} on 4 images. From left, in images (1) and (2) the model detects all the lines including lines that are not cracks; in (3) predicts patches rather than cracks; and in (4) predicts correct position but incorrect crack width.}
  \label{fig:unet-cracks}
\end{figure}

The quantitative results in Table \ref{tab:results-vessels} are not state of the art, so we further investigate the choice of the ViT-B\cite{vit} architecture for our crack detection problem given that the U-Net\cite{unet} architecture performs better for blood vessel segmentation (See Table \ref{tab:unet-vessels}). We show that the U-Net architecture does not perform as well on the crack segmentation task across the four losses. We believe this is because while rock crack detection and blood vessel detection are similar tasks for benchmarking, they still have nuanced differences, such as blood vessels having a more radial configuration as they emerge from a central optic nerve \cite{kolbanatomy1995}. Figure \ref{fig:unet-cracks} and Table \ref{tab:unet-cracks} report U-Net results.

For completeness, a comparison of the losses is reported in Table \ref{tab:unet-cracks} and \ref{tab:unet-vessels}. The networks are trained from scratch with a maximum learning rate of $1\times10^{-2}$ for $40000$ steps (approximately $2000$ epochs depending on the size of the dataset) using the Adam\cite{adam} optimizer. While the results slightly favor SKIL, it is not possible to conclude statistical significance given the relative error.

\begin{table}[h]
\centering
    \scalebox{0.92}{
\ra{0.8}
\setlength\tabcolsep{3pt}
\begin{tabular}{@{}lcccccccc@{}}\toprule
& \multicolumn{2}{c}{\textbf{Classical}} & & \multicolumn{5}{c}{\textbf{LineAcc}} \\ 
\cmidrule{2-3} \cmidrule{5-9} 
& IOU & Dice && Pos. & Width & Length && Combined\\ \midrule
\phantom{ab} Dice & 0.093 & 0.166 ± 0.003 && 0.397 & 0.218 & \textbf{0.640} && 0.372 ± 0.003\\
\phantom{ab} CL-Dice & \textbf{0.116} & \textbf{0.203 ± 0.003} && 0.393 & \textbf{0.374} & 0.625 && \textbf{0.391 ± 0.003}\\
\phantom{ab} SKIL-Dice & 0.109 & 0.192 ± 0.002 && 0.400 & 0.249 & 0.610 && 0.374 ± 0.004\\
\phantom{ab} SKIL-Prod. & 0.106 & 0.188 ± 0.003 && \textbf{0.403} & 0.227 & 0.620 && 0.375 ± 0.004\\
\bottomrule
\end{tabular}
}
\caption{Metrics when training a U-Net architecture \textit{(averaged over 10 runs)} on the \textbf{cracks dataset} with 4 losses: Dice, CL-Dice, SKIL-Dice \textit{(This Work)} and SKIL-Product \textit{(This Work)}. The standard error is reported for the Dice and $\text{LineAcc}_{\text{comb.}}$ score. U-Net performs worse on the cracks dataset across all 4 losses compared to Table \ref{tab:results-cracks}.}
\label{tab:unet-cracks}
\end{table}

\begin{table}[]
\centering
    \scalebox{0.82}{
\ra{0.8}
\setlength\tabcolsep{3pt}
\begin{tabular}{@{}lccccc@{}}\toprule
& \multicolumn{2}{c}{\textbf{STARE}} & & \multicolumn{2}{c}{\textbf{DRIVE}}\\
\cmidrule{2-3} \cmidrule{5-6}
& Dice & LineAcc-Comb. && Dice & LineAcc-Comb.\\ \midrule
\phantom{ab} Cross Entropy & 0.8200 ± 0.0008 & 0.8932 ± 0.0013 && 0.7684 ± 0.0009 & 0.8680 ± 0.0009\\
\phantom{ab} CL-Dice & 0.8204 ± 0.0009 & 0.8860 ± 0.0009 && \textbf{0.7750 ± 0.0002} & 0.8639 ± 0.0008\\
\phantom{ab} SKIL-Dice & 0.8175 ± 0.0009 & 0.9024 ± 0.0014 && 0.7679 ± 0.0009 & \textbf{0.8868 ± 0.0008}\\
\phantom{ab} SKIL-Product & \textbf{0.8205 ± 0.0008} & \textbf{0.9027 ± 0.0011} && 0.7718 ± 0.0005 & 0.8850 ± 0.0008\\
\bottomrule
\end{tabular}
}
\caption{Comparison of different loss scores trained on a U-Net architecture \textit{(averaged over 10 runs)} and standard errors when trained separately on \textbf{STARE} and \textbf{DRIVE} \textit{(HRF and CHASE DB1 have been omitted for readability)}. The loss used consisted of the Dice loss (coefficient of 3) and the loss in this Table (coefficient of 1).}
\label{tab:unet-vessels}
\end{table}

\subsection{Generalization on poor quality annotations}
\label{sec:result-gen}
One of the motivations of SKIL and $\text{LineAcc}_{\text{pos}}$ was to abstract the width of the cracks from the training and evaluation. This was important as the true width of a graspable crack is subjective (to the point that it was not always consistent between labelers). More generally, given the stochasticity of grasping with microspine grippers, crack detection for robotic grasping may always be poorly annotated. To test SKIL's performance on poorly annotated data, we created some random deformations to the blood vessel labels, trained on them and evaluated them on clean labels.

\subsubsection{Generation of poorly annotated blood vessels}
\label{sec:result-gen-gen}
\begin{figure}[h]
  \captionsetup[subfigure]{justification=centering, labelformat=empty} 
  \centering

  \begin{subfigure}{0.235\linewidth}
    \caption{Shift}
    \includegraphics[width=\linewidth]{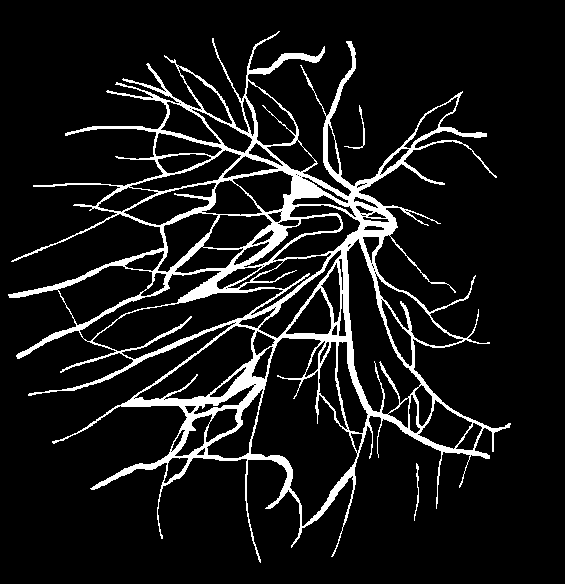}
  \end{subfigure}
  \begin{subfigure}{0.235\linewidth}
    \caption{Random Width}
    \includegraphics[width=\linewidth]{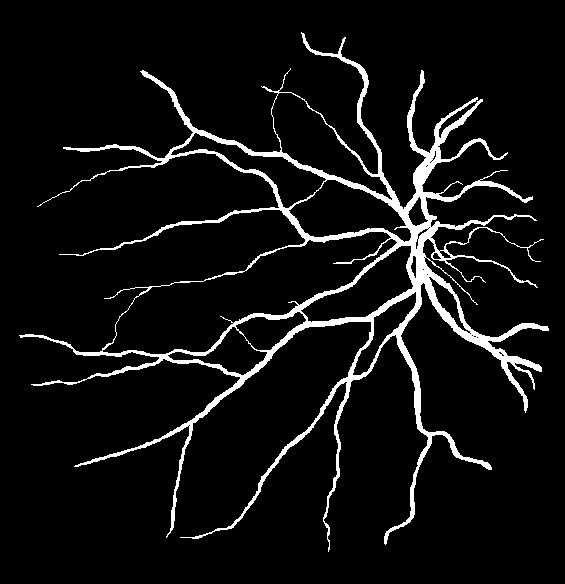}
  \end{subfigure}
  \begin{subfigure}{0.235\linewidth}
    \caption{Branch Crop}
    \includegraphics[width=\linewidth]{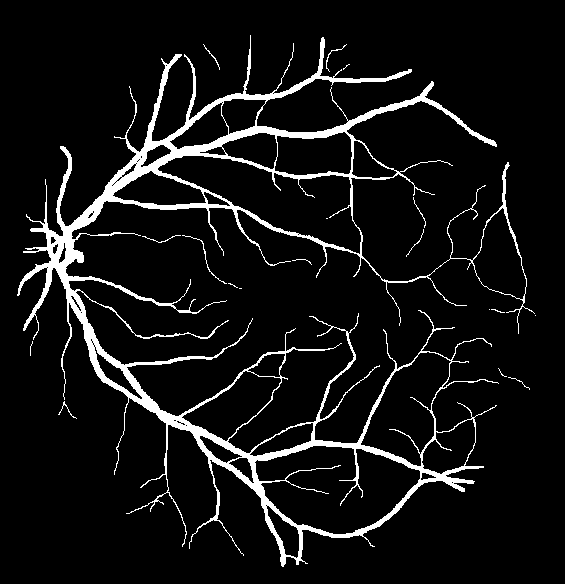}
  \end{subfigure}
  \begin{subfigure}{0.235\linewidth}
    \caption{Combination}
    \includegraphics[width=\linewidth]{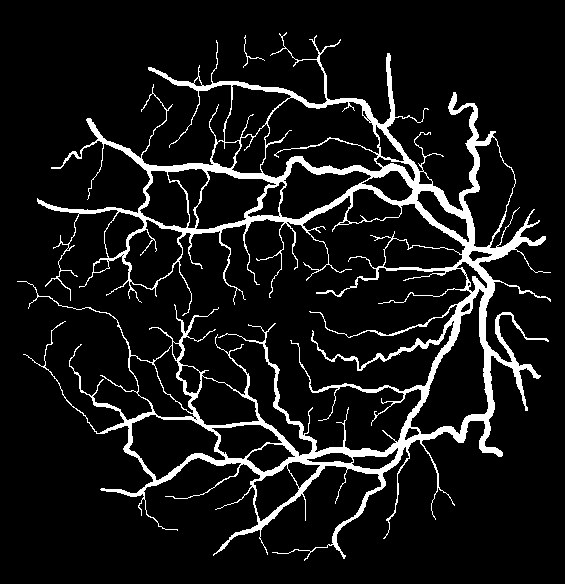}
  \end{subfigure}

\vspace{0.03\linewidth}

  \begin{subfigure}{0.235\linewidth}
    \includegraphics[width=\linewidth]{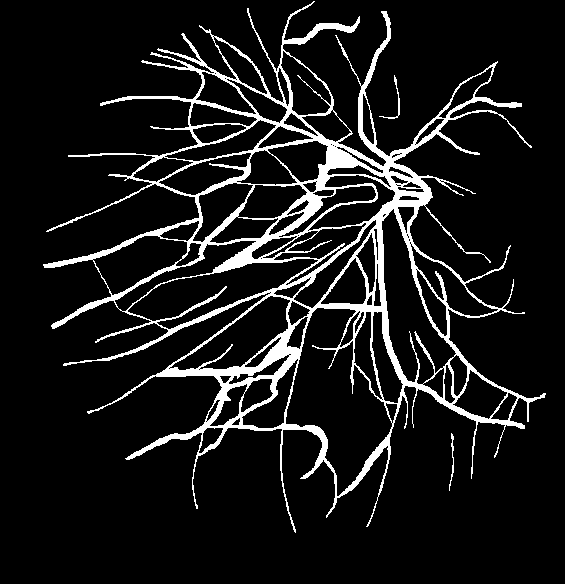}
    \end{subfigure}
  \begin{subfigure}{0.235\linewidth}
    \includegraphics[width=\linewidth]{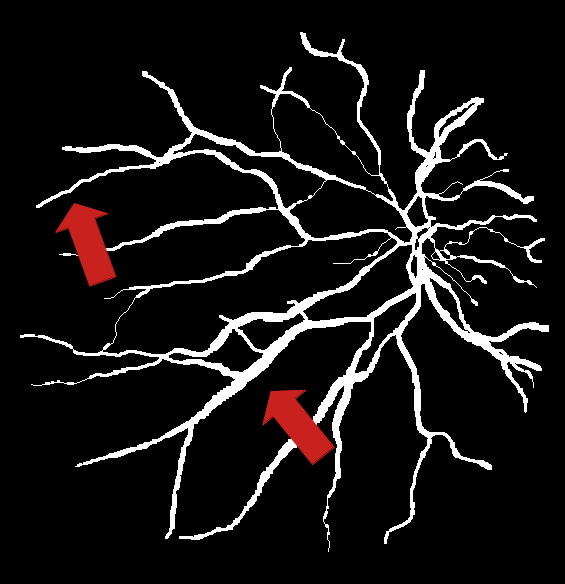}
  \end{subfigure}
\begin{subfigure}{0.235\linewidth}
    \includegraphics[width=\linewidth]{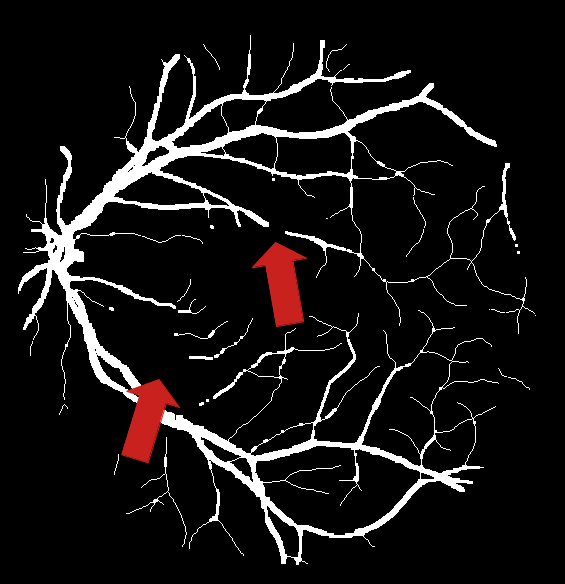}
  \end{subfigure}
  \begin{subfigure}{0.235\linewidth}
    \includegraphics[width=\linewidth]{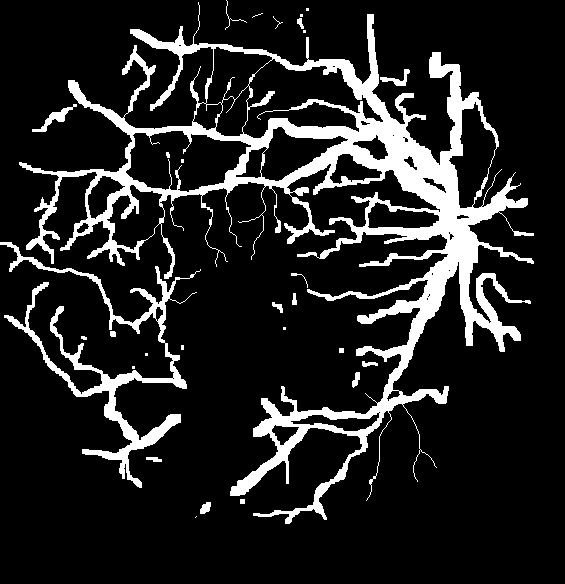}
  \end{subfigure}

\vspace{0.01\linewidth}

  \begin{subfigure}{0.235\linewidth}
    \includegraphics[width=\linewidth]{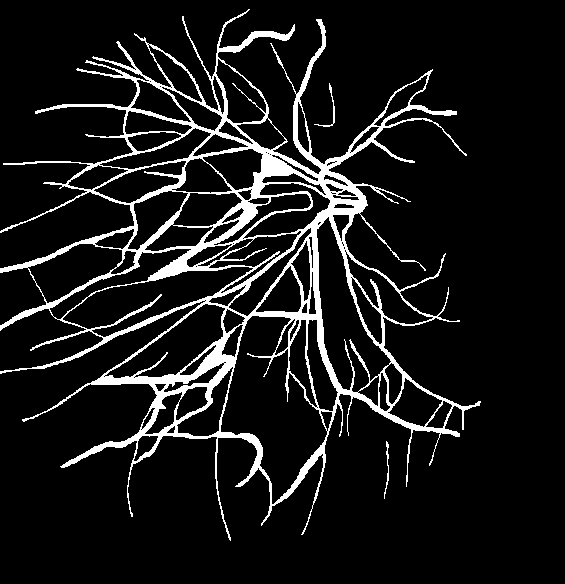}
  \end{subfigure}
  \begin{subfigure}{0.235\linewidth}
    \includegraphics[width=\linewidth]{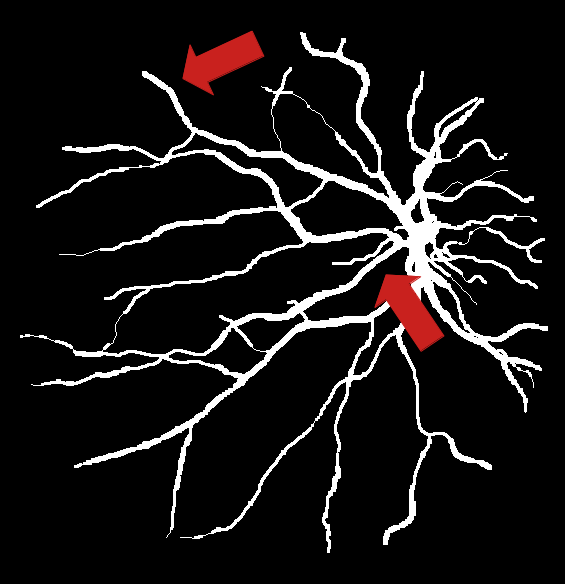}
  \end{subfigure}
  \begin{subfigure}{0.235\linewidth}
    \includegraphics[width=\linewidth]{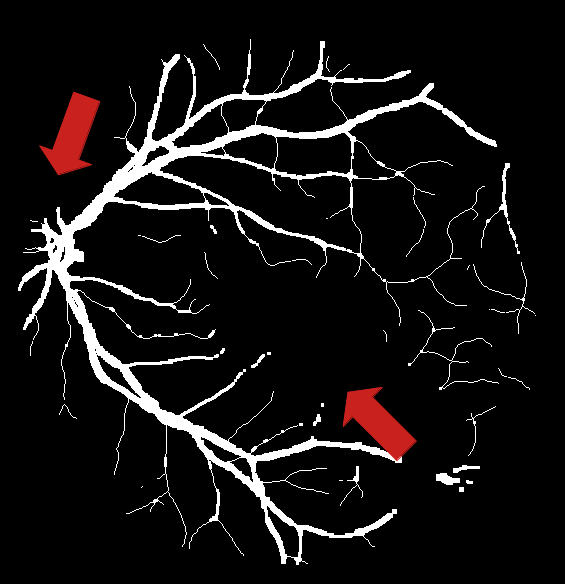}
  \end{subfigure}
  \begin{subfigure}{0.235\linewidth}
    \includegraphics[width=\linewidth]{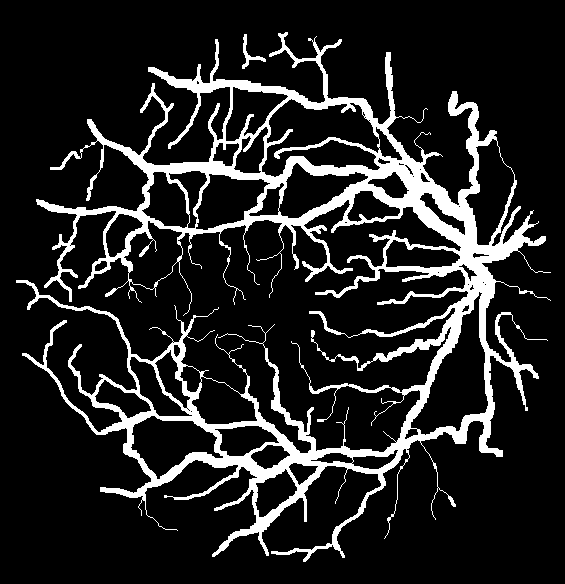}
  \end{subfigure}
  
  \caption{Deformed annotations on the DRIVE dataset. The first row shows the original annotation; following rows show deformation examples (ordered by column). The last column is a combination of the shift, random width, and branch crop deformations, with a probability of 0.5 for each deformation. Examples of visual changes are marked by red arrows. \textit{(The last column is unmarked as the changes are visible enough.)}}
  \label{fig:result-vessels-deformed}
\end{figure}

We created 3 deformations (Figure \ref{fig:result-vessels-deformed}): a simple \textbf{shift}, a \textbf{random width change} that changes the width of the vessels but does not create or remove any vessels, and a \textbf{random branch cutter} which removes vessels with a probability inversely proportional to the thickness. Below we explain the latter two.

To change the width of the vessels, given a binary image $I$, we first compute the skeleton $
\mathcal{S}_I$ as well as the distance to the closest pixel equal to 0 at each pixel, noted $D_I^0$ (Analog to $d$ in Section \ref{sec:metric-center}). Then we multiply $D_I^0$ by some Perlin noise \cite{perlin} $\mathcal{N}^P$ to finally apply a \textbf{decreasing dilation} which consists of iteratively applying some maxpool of size $3 \times 3$ and subtract 1 each time to the pixels that were equal to 0 before the pooling. This way, a pixel of value $x$ will generate a diamond of width $x$.
\begin{equation}
    \mathcal{D}_{\text{width}}(I) = \left( \textbf{decrease\_dilate}(D_I^0 * \mathcal{S}_I * \mathcal{N}^P)\right) > 0
\end{equation}

The parameters chosen for the Perlin noise were: amplitude between $0.1$ and $2.0$, resolution of $3$, $4$ octaves with a persistence of $0.5$, and a lacunarity of $3$.

The random branch cutter also relies on the same idea, however instead of multiplying the distance values $D_I^0$ by some Perlin noise, the values are changed to 0 based on a probability $\mathcal{P}$ computed as follows:
\begin{equation*}
    \mathcal{P}(I) = \mathcal{N}^P * \left(\frac{\sum \mathcal{S}_I * D_I^0}{\sum \mathcal{S}_I} * \frac{1}{\epsilon + D_I^0}\right)^{\alpha}
\end{equation*}
where $\epsilon = 1 \times 10^{-3}$ is a constant to avoid dividing by zero and $\alpha$ is the selectiveness, set to $0.2$ in our experiments. A pixel is then set to 0 if $\mathcal{P}(I) > 1 - p$ with $p=0.35$. All the parameters for the Perlin noise are identical.

\subsubsection{Quantitative results}
\label{sec:result-gen-quantitative}

Table \ref{tab:result-vessels-deformed} summarizes the results obtained with the four different deformations and shows the performance of each loss when the model is trained on the deformed data, and Table \ref{tab:result-vessels-deformed-relative} helps to interpret these results by reporting relative performance between the network trained on the clean data and deformed data. We see that SKIL substantially improves performance on \textit{shifted labels} deformations due to its soft dilation. The results for the \textit{random width} deformation show that even with the clean data the network does not perform well regarding the width prediction as the deformation does not deteriorate this metric at all. The \textit{cropped branches} deformation seems to indicate that SKIL slightly performs better when some annotations are missing but the difference is too small to generalize. With combined deformations, SKIL outperforms CL-Dice and Dice so we conclude that SKIL does adapt to poorly annotated data.

\begin{table}[H]
\centering
    \scalebox{0.92}{
\ra{0.8}
\setlength\tabcolsep{3pt}
\begin{tabular}{@{}lcccccccc@{}}\toprule
& \multicolumn{2}{c}{\textbf{Classical}} & & \multicolumn{5}{c}{\textbf{LineAcc}} \\ 
\cmidrule{2-3} \cmidrule{5-9} 
& IOU & Dice && Pos. & Width & Length && Combined\\ \midrule
\multicolumn{5}{l}{\textit{Shifted labels}}\\
\phantom{ab} Dice & 0.215 & 0.349 ± 0.002 && 0.485 & 0.490 & 0.504 && 0.444 ± 0.002\\
\phantom{ab} CL-Dice & \textbf{0.221} & \textbf{0.359 ± 0.002} && 0.420 & 0.441 & 0.481 && 0.407 ± 0.002\\
\phantom{ab} SKIL-Dice  & 0.210 & 0.340 ± 0.003 && \textbf{0.751} & \textbf{0.564} & \textbf{0.700} && \textbf{0.618 ± 0.002}\\
\multicolumn{5}{l}{\textit{Random width}}\\
\phantom{ab} Dice & \textbf{0.276} & \textbf{0.419 ± 0.001} && 0.778 & 0.574 & 0.678 && 0.638 ± 0.001\\
\phantom{ab} CL-Dice & 0.274 & 0.418 ± 0.001 && 0.776 & 0.541 & 0.688 && 0.635 ± 0.001\\
\phantom{ab} SKIL-Dice  & 0.268 & 0.406 ± 0.001 && \textbf{0.831} & \textbf{0.616} & \textbf{0.783} && \textbf{0.687 ± 0.001}\\
\multicolumn{5}{l}{\textit{Cropped branches}}\\
\phantom{ab} Dice & 0.267 & 0.407 ± 0.001 && 0.682 & 0.581 & 0.599 && 0.576 ± 0.001\\
\phantom{ab} CL-Dice & \textbf{0.274} & \textbf{0.417 ± 0.001} && 0.671 & 0.549 & 0.605 && 0.571 ± 0.001\\
\phantom{ab} SKIL-Dice  & 0.264 & 0.401 ± 0.001 && \textbf{0.795} & \textbf{0.637} & \textbf{0.689} && \textbf{0.651 ± 0.001}\\
\multicolumn{5}{l}{\textit{Combined deformations}}\\
\phantom{ab} Dice & \textbf{0.271} & \textbf{0.419 ± 0.001} && 0.653 & 0.507 & 0.577 && 0.551 ± 0.001\\
\phantom{ab} CL-Dice & 0.269 & 0.418 ± 0.001 && 0.623 & 0.475 & 0.568 && 0.532 ± 0.001\\
\phantom{ab} SKIL-Dice  & 0.266 & 0.410 ± 0.001 && \textbf{0.798} & \textbf{0.558} & \textbf{0.718} && \textbf{0.651 ± 0.001}\\
\bottomrule
\end{tabular}
}
\caption{Different metrics on the \textbf{vessels datasets (aggregation of DRIVE, HRF, STARE, CHASE-DB1)} \textit{(averaged over 10 runs)} when training for 200 epochs with 3 losses: Dice, CL-Dice and SKIL \textit{(This Work)}. Standard error is reported. The experiments were run on 4 types of deformations. Combined deformation is a combination of shift, random width, and branch crop with a probability of 0.75 for each deformation.}
\label{tab:result-vessels-deformed}
\end{table}
\vspace{-1em}

\begin{table}[H]
\centering
    \scalebox{0.88}{
\ra{0.8}
\setlength\tabcolsep{2pt}
\begin{tabular}{@{}lcccccccc@{}}\toprule
& \multicolumn{2}{c}{\textbf{Classical}} & & \multicolumn{5}{c}{\textbf{LineAcc}} \\ 
\cmidrule{2-3} \cmidrule{5-9} 
& IOU & Dice && Pos. & Width & Length && Combined\\ \midrule
\multicolumn{5}{l}{\textit{Shifted labels}}\\
\phantom{ab} Dice & 81.4\% & 87.3 ± 0.7\% && 65.0\% & 77.7\% & 78.4\% && 71.8 ± 0.3\%\\
\phantom{ab} CL-Dice & 81.9\% & \textbf{87.6 ± 0.4\%} && 56.4\% & 75.5\% & 73.2\% && 65.9 ± 0.3\%\\
\phantom{ab} SKIL-Dice  & \textbf{81.7\%} & 87.4 ± 0.7\% && \textbf{91.25\%} & \textbf{82.3\%} & \textbf{93.8\%} && \textbf{91.0 ± 0.2\%}\\
\multicolumn{5}{l}{\textit{Random width}}\\
\phantom{ab} Dice & \textbf{104.5\%} & \textbf{104.8 ± 0.2\%} && \textbf{104.3\%} & 91.0\% & \textbf{105.4\%} && \textbf{103.2 ± 0.1\%}\\
\phantom{ab} CL-Dice & 101.5\% & 102.0 ± 0.1\% && 104.2\% & \textbf{92.6\%} & 104.7\% && 102.8\% ± 0.1\%\\
\phantom{ab} SKIL-Dice  & 104.3\% & 104.4 ± 0.1\% && 101.0\% & 89.9\% & 105.0\% && 101.2 ± 0.1\%\\
\multicolumn{5}{l}{\textit{Cropped branches}}\\
\phantom{ab} Dice & 101.1\% & 101.8 ± 0.1\% && 91.4\% & 92.1\% & \textbf{93.2\%} && 93.2 ± 0.2\%\\
\phantom{ab} CL-Dice & 101.5\% & 101.7 ± 0.1\% && 90.1\% & \textbf{94.0\%} & 92.1\% && 92.4 ± 0.1\%\\
\phantom{ab} SKIL-Dice  & \textbf{102.7\%} & \textbf{103.1 ± 0.1\%} && \textbf{96.6\%} & 93.0\% & 92.4\% && \textbf{95.9 ± 0.1\%}\\     
\multicolumn{5}{l}{\textit{Combined derformations}}\\
\phantom{ab} Dice & 102.7\% & 104.8 ± 0.2\% && 87.5\% & 80.3\% & 89.7\% && 89.2 ± 0.2\%\\
\phantom{ab} CL-Dice & 99.3\% & 102.0 ± 0.2\% && 83.6\% & 81.3\% & 86.5\% && 86.1 ± 0.2\%\\
\phantom{ab} SKIL-Dice  & \textbf{103.5\%} & \textbf{105.4 ± 0.2\%} && \textbf{97.0\%} & \textbf{81.5\%} & \textbf{96.2\%} && \textbf{95.9 ± 0.1\%}\\
\bottomrule
\end{tabular}
}
\caption{Relative performance of Table \ref{tab:result-vessels-deformed}. For a given method and metric, the score trained on the deformed data is divided by the score trained on the clean data.}
\label{tab:result-vessels-deformed-relative}
\end{table}


\section{Conclusion}
\label{sec:conclusion}

In this work, we present SKIL, a skeleton-based segmentation loss to identify cracks on rock walls, and evaluate its performance on both rock cracks and a similar task, blood vessel detection. We also present three novel metrics (LineAcc) for evaluating this method. Finally, we publish a new dataset containing 236 images of rocks with graspable edges and cracks annotated. Experiments show that SKIL and the LineAcc metrics perform well on our dataset of rock cracks and on several existing datasets for blood vessel detection.

In this work, we take the first step to developing an accurate learning-based approach for visual rock crack detection to enable a climbing robot. Next steps include parameter finetuning, deployment on hardware for field testing with a partial robot system, and interfacing with motion planning systems to enable full autonomy. 


\section*{Acknowledgements}
\label{sec:acknowledgement}

Support for this work was provided by NASA under the Innovative Advanced Concepts (NIAC) program.


\bibliographystyle{IEEEtran}
\bibliography{IEEEabrv,egbib}

\end{document}